\documentclass[journal]{IEEEtran}

\usepackage{times}
\usepackage{epsfig}
\usepackage{graphicx}
\usepackage{amsmath}
\usepackage{amssymb}
\usepackage{array}
\usepackage{multirow}
\usepackage{xcolor}
\usepackage[normalem]{ulem}
\usepackage{soul}
\useunder{\uline}{\ul}{}
\usepackage{slashbox}
\usepackage{subcaption}
\usepackage{hyperref}

\usepackage{rotating}
\usepackage{adjustbox}
\usepackage[flushleft]{threeparttable}

\usepackage[switch]{lineno}
\usepackage{enumitem,amssymb}
\usepackage{pifont}

\newcolumntype{C}[1]{>{\centering}m{#1}}

\begin{document}

    \IEEEpeerreviewmaketitle
    \bibliographystyle{IEEEtran}

\title{A Gated Fusion Network for Dynamic Saliency Prediction}
\author{Aysun Kocak$^1$, Erkut Erdem$^1$ and~Aykut Erdem$^2$\\
$^1$Department of Computer Engineering, Hacettepe University, Ankara, Turkey\\
$^2$Department of Computer Engineering, Ko\c{c} University, Istanbul, Turkey}

\maketitle  
\IEEEpeerreviewmaketitle

\begin{abstract}
Predicting saliency in videos is a challenging problem due to complex modeling of interactions between spatial and temporal information, especially when ever-changing, dynamic nature of videos is considered. Recently, researchers have proposed large-scale datasets and models that take advantage of deep learning as a way to understand what's important for video saliency. These approaches, however, learn to combine spatial and temporal features in a static manner and do not adapt themselves much to the changes in the video content. In this paper, we introduce Gated Fusion Network for dynamic saliency (GFSalNet), the first deep saliency model capable of making predictions in a dynamic way via gated fusion mechanism. Moreover, our model also exploits spatial and channel-wise attention within a multi-scale architecture that further allows for highly accurate predictions. We evaluate the proposed approach on a number of datasets, and our experimental analysis demonstrates that it outperforms or is highly competitive with the state of the art. Importantly, we show that it has a good generalization ability, and moreover, exploits temporal information more effectively via its adaptive fusion scheme.
\end{abstract}


\section{Introduction}
Human visual system employs visual attention mechanisms to effectively deal with huge amount of information by focusing only on salient or attention grabbing parts of a scene, and thus filtering out irrelevant stimuli. Saliency estimation methods offer different computational models of attention to mimic this key component of our visual system. These methods generate a so-called saliency map within which a pixel value indicates the likelihood of that pixel being fixated by a human. Since the pioneering work of~\cite{itti1998model}, this research area has gained a lot of interest in the last few decades (please refer to~\cite{borji2013state, Filipe2015} for an overview), and it has found to have practical use in a variety of computer vision tasks such as visual quality assessment~\cite{7303954, 7444164}, image and video resizing~\cite{6199980,6529173}, video summarization~\cite{6527322}, to name a few. 
Early saliency prediction approaches use low-level (color, orientation, intensity) and/or high-level (pedestrians, faces, text, etc.) image features to estimate salient regions. While low-level cues are used to detect regions that are different from their surroundings, top-down cues are used to infer high-level semantics to guide the model. For example, humans tend to focus some object classes more than others. Recently, deep learning based models have started to dominate over the traditional approaches as they can directly learn both low and high-level features relevant for saliency prediction~\cite{Bruce_2016_CVPR, bylinskii2016should}. 

\begin{figure}[!t]
\centering
\begin{tabular}{p{0.48\columnwidth}@{$\;$}p{0.48\columnwidth}}
\includegraphics[width=0.24\columnwidth]{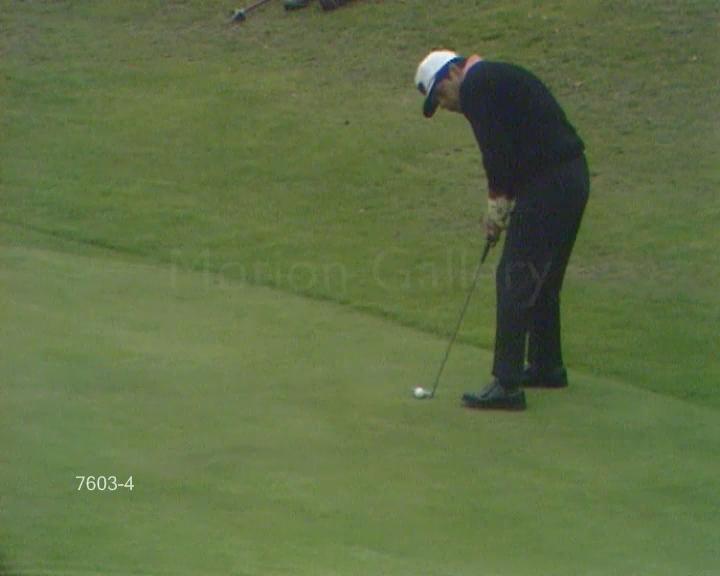}\includegraphics[width=0.24\columnwidth]{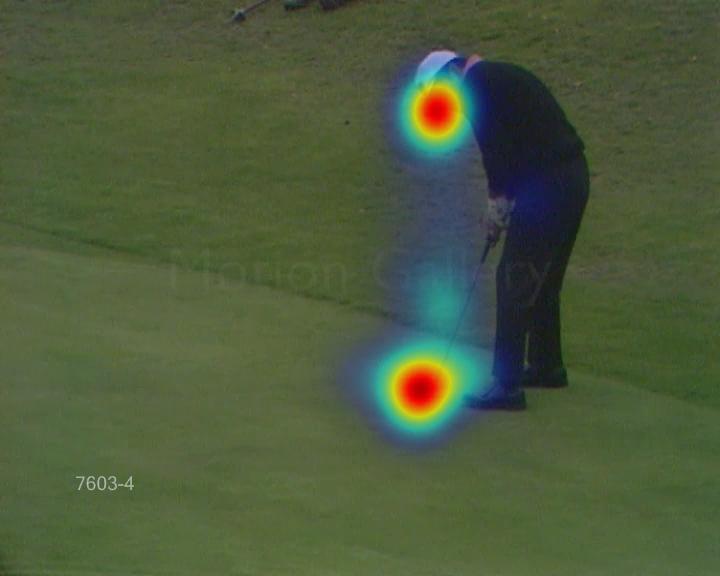}&
\includegraphics[width=0.24\columnwidth]{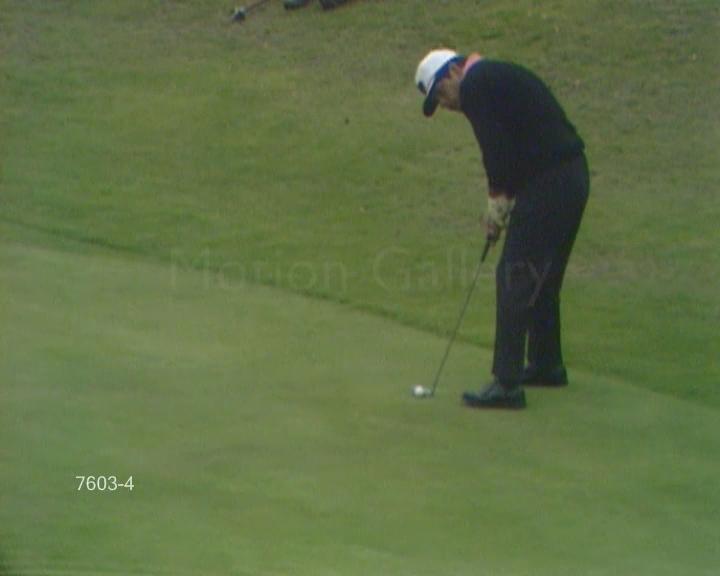}\includegraphics[width=0.24\columnwidth]{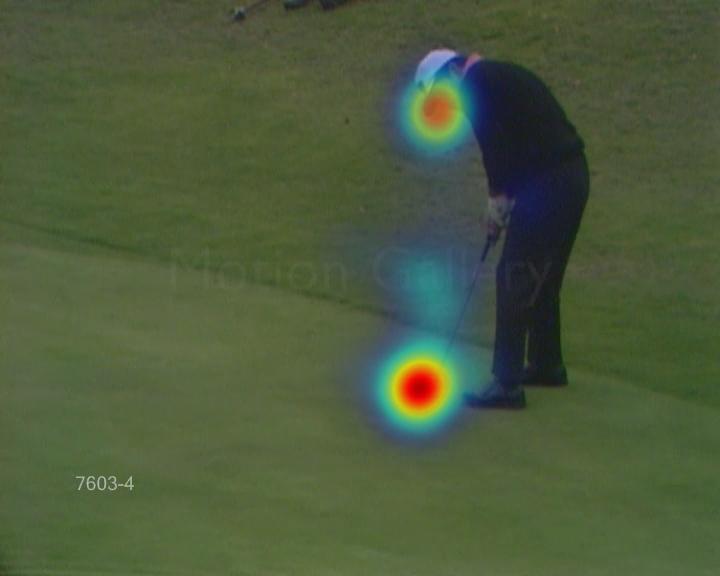} \\
\includegraphics[width=0.24\columnwidth]{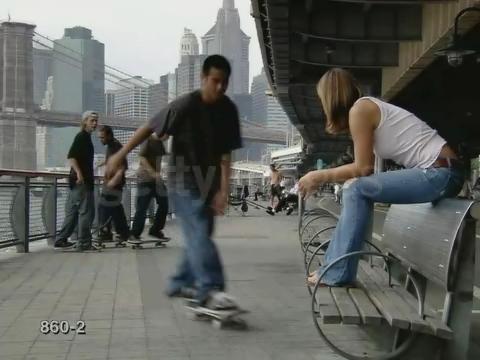}\includegraphics[width=0.24\columnwidth]{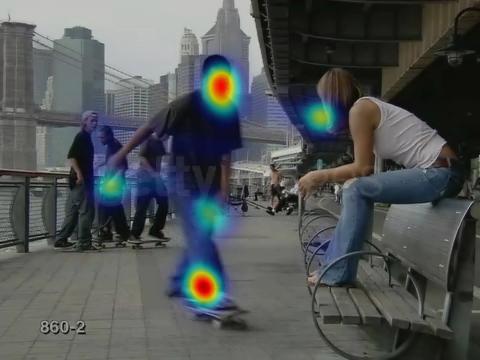}&
\includegraphics[width=0.24\columnwidth]{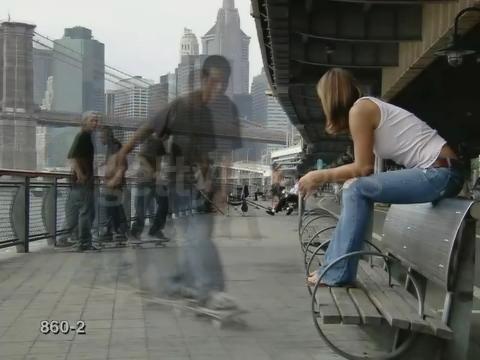}\includegraphics[width=0.24\columnwidth]{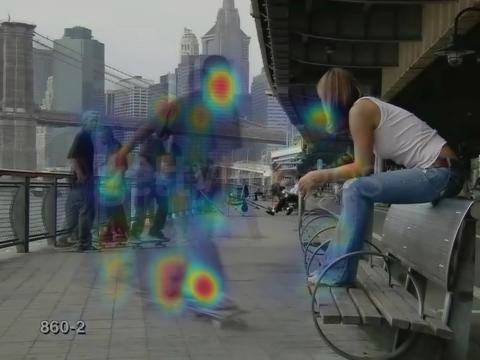} \\ 
{\footnotesize{ \vspace{-0.3cm} A single input frame and its corresponding fixation map}} & {\footnotesize{ \vspace{-0.3cm} Four consecutive overlaid frames and their overlaid fixation maps}}
\end{tabular}
\caption{Predicting video saliency requires finding a harmonious interaction between appearance and temporal information. For example, while the first row shows a case in which attention is guided more by visual appearance, in the second row, motion is the most determining factor for attention. Hence, we speculate that an adaptive scheme would be better suited for this task.}
\label{fig:contrib-adapt}
\end{figure}

Most of the literature on saliency estimation focuses on  static images. Lately, predicting saliency in videos has also gained some attraction, but it still remains a largely unexplored field of research. Video saliency models (also called dynamic saliency models) aim to predict attention grabbing regions in dynamically changing scenes. While static saliency estimation considers only low-level and high-level spatial cues, dynamic saliency needs to take into account temporal information too as there is evidence that moving objects or object parts can also guide our attention. Motion and appearance play complementary roles in human attention and their significance can change over time. As we illustrate in Fig.~\ref{fig:contrib-adapt}, in dynamic scenes, humans tend to focus more on moving parts of the scene and the eye fixations change over time, showing the importance of motion cues (bottom row). On the other hand, when there is practically no motion in the scene, low-level appearance cues dominantly guide our attention and we focus more on the regions showing different visual characteristics than their surroundings (top row). Motivated by these observations, in this work, we develop a deep dynamic saliency model which handles spatial and temporal changes in the visual stimuli in an adaptive manner. 

The first generation of dynamic saliency methods were simply extensions of the static saliency approaches, \textit{e.g.}~\cite{guo2008spatio, cui2009temporal, seo2009static, sultani2014human, mauthner2015encoding}. In other words, these methods adapted the strategies proposed for static scenes and mostly modified them to work on either 3D feature maps that are formed by stacking 2D spatial features over time or 2D feature maps encoding motion information like optical flow images. Several follow-up works, however, have approached the problem from a fresh perspective and developed specialized methods for dynamic saliency detection, \textit{e.g.} ~\cite{6239258, mathe2012dynamic, rudoy2013learning, inproceedingszhong, liu2014superpixel, zhou2014learning, zhao2015fixation, hossein2015many, awsd}. These models either utilize novel spatio-temporal features or employ data-driven techniques to learn relevant features from data. As with the case of state-of-the-art static saliency models, approaches based on deep learning have also shown promise for dynamic saliency. These studies basically explore different neural architectures used for processing temporal and spatial information in a joint manner, and they either use 3D convolutions~\cite{BazzaniLT16}, LSTMs~\cite{BazzaniLT16, wang2018revisiting} or multi-stream architectures that encode temporal information separately~\cite{PalazziACSC17, BakEE16, Jiang_2018_ECCV}.
 
In this work, we introduce Gated Fusion Network for video saliency (GFSalNet). Our proposed network model is radically different from the previously proposed deep models in that it includes a novel content-driven fusion scheme to combine spatial and temporal streams in a more dynamic manner.  In particular, our model is based on two-stream CNNs~\cite{karpathy2014large,simonyan14b}, which have been successfully applied to various video analysis tasks. To our interest, these architectures are inspired by the ventral and dorsal pathways in the human visual cortex~\cite{two-stream-hypothesis}. Although the use of two-stream CNNs in video saliency prediction has been investigated before~\cite{BakEE16}, the main novelty of our work lies in the ability to fuse appearance and motion information in a spatio-temporally coordinated manner by estimating the importance of each cue with respect based on the current video content.

The rest of the paper is organized as follows: In Section~2, we give a brief overview of the existing dynamic saliency approaches. In Section~3, we present the details of our proposed deep architecture for video saliency. In Section~4, we give the details of our experimental setup, including evaluation metrics, datasets and the competing dynamic saliency models, and discuss the results of our experiments. Finally, in the last section, we offer some concluding remarks.

Our codes and predefined models, along with the saliency maps extracted with our approach, will be publicly available at the project website\footnote{\url{https://hucvl.github.io/GFSalNet/}}.

\section{Related Work}
 Early visual saliency models can be dated back to 1980s with the Feature Integration Theory by~\cite{TREISMAN198097}. The first models of saliency, such as~\cite{koch85,itti1998model}, provide computational solutions to~\cite{TREISMAN198097}, and since then a notable number of saliency models are developed, most of which deal with static scenes. For a detailed list of pre-deep learning saliency estimation approaches, please refer to~\cite{borji2013state}. After the availability of large-scale datasets, researchers proposed various deep learning based models for static saliency that outperformed previous approaches by a large margin~\cite{7410395,7780989,7937829,7762165,shallowPan,Wang:2018:DVA:3195804.3195826,6909754,8400593}.

\textbf{Early models for dynamic saliency} generally depend on previously proposed static saliency models. Adaptation of these models to dynamic scenes is achieved by considering features related to motion such as the optical flow information. For example,~\cite{guo2008spatio} proposed a saliency prediction method called PQFT that predicts the salient regions via the phase spectrum of Fourier Transform of the given image. In particular, PQFT generates a quaternion image representation by using color, intensity, orientation and motion features and estimates the salient regions in the frequency domain by using this combined representation. \cite{cui2009temporal} extracted salient parts of video frames by similarly performing a spectral analysis of the frames considering both spatial and temporal domains.~\cite{seo2009static} employed local regression kernels as features to calculate self similarities between pixels or voxels for figure-ground segregation. ~\cite{sultani2014human} extended the previously proposed static saliency model by~\cite{harel2006graph}'s model by including motion cues to the graph-theoretic formulation. \cite{Fang:ICME2014} employ a two stream approach that generates spatial saliency map (using color and texture features) and temporal saliency map (using optical flow feature) separately and combines these maps with an entropy based adaptive method.~\cite{mauthner2015encoding} proposed a dynamic saliency model for activity recognition that works in an unsupervised manner. Their method is based on an encoding scheme that considers color along with motion cues.

Following these early approaches, the researchers started to develop novel video saliency models specifically designed for dynamic stimuli. For instance,~\cite{6239258} proposed a sparsity based framework that generates spatial saliency maps and temporal saliency maps separatelty based on entropy gain and temporal consistency, respectively, and then combines them. ~\cite{mathe2012dynamic} integrated several visual cues such as static and dynamic image features based on color, texture, edge distribution, motion boundary histograms, through learning-based fusion strategies and later employed this dynamic saliency model for action recognition. ~\cite{rudoy2013learning} suggested a learning-based model that generates a candidate set regions with the use of existing methods and then predicts gaze transitions over subsequent video frames conditionally on these regions.~\cite{inproceedingszhong} proposed a simple dynamic saliency model that combines spatial saliency maps with temporal saliency using pixel-wise maximum operation. In their work, while the spatial saliency maps are extracted using multi-scale analysis of low-level features, temporal saliency maps are obtained by examining dynamic consistency of motion through an optical flow model.~\cite{liu2014superpixel} suggested an approach that independently estimates superpixel-level and pixel-level temporal and spatial saliency maps and subsequently combines them using an adaptive fusion strategy.~\cite{zhou2014learning} proposed an approach that oversegments video frames by using both spatial and temporal information and estimates the saliency score for each region by computing the regional contrast values via low-level features extracted from these regions.~\cite{zhao2015fixation} suggested to learn a filter bank from low-level features for fixations. This filterbank encodes the association between local feature patterns and probabilities of human fixations, and is used to re-weight fixation candidates.~\cite{hossein2015many} formulated another dynamic saliency model by exploiting the  compressibility principle. More recently,~\cite{awsd} proposed a saliency model (called AWS-D) for dynamic scenes by considering the observation that high-order statistical structures carry most of the perceptually relevant information. AWS-D~\cite{awsd} removes the second-order information from input sequence via a whitening process. Then, it computes bottom-up spatial saliency maps using a filter bank at multiple scales, and temporal saliency maps with the use of a 3D filter bank. Finally, it combines all these maps by considering their relative significance.

\textbf{Deep learning based dynamic saliency models} have received attention only recently.~\cite{BazzaniLT16} proposed a recurrent mixture density network (RMDN) for spatio-temporal visual attention. The method uses a C3D architecture~\cite{7410867} as a backbone to integrate spatial and temporal information. This representation module is fed to a Long Short-Term Memory (LSTM) network, which is connected to Mixture Density Network (MDN) whose outputs are the parameters of a Gaussian mixture model expressing the saliency map of each frame.~\cite{BakEE16} suggested a two stream CNN model~\cite{karpathy2014large,simonyan14b} which considers the motion and appearance clues in videos. While, optical flow images are used to feed the temporal stream, raw RGB frames are used as input for the spatial stream.~\cite{PalazziACSC17} presented an attention network to predict where driver is focused. In this work, the authors also  proposed a dataset that consists of ego-centric and car-centric driving videos and eye tracking data belongs to the videos. Their network consists of three independent paths, namely spatial, temporal and semantic paths. While the spatial path uses raw RGB data as input, the temporal one uses optical flow data to integrate motion information and the last one processes the segmentation prediction on the scene given by the model by~\cite{YuK15}. In the final layer of the network, the three independent maps are summed and then normalized to obtain the final saliency map.~\cite{Jiang_2018_ECCV} proposed a deep model called OM-CNN which consists of two subnetworks, namely objectness subnet to highlight the regions that contain an object, motion subnet to encode temporal information, whose outputs are then combined to generate some spatio-temporal features.~\cite{wang2018revisiting} proposed a model called ACLNet which employs a CNN-LSTM architecture to predict human gaze in dynamic scenes. The proposed approach focuses static information with an attention module and allows an LSTM to focus on learning dynamic information. Recently,~\cite{Linardos2019} proposed an encoder-decoder based deep neural network called SalEMA, which employs a convolutional recurrent neural network method to include temporal information. In particular, it processes a sequence of RGB video frames as input to employ spatial and temporal information with the temporal information being inferred by the weighted average of the convolution state of the current frame and all the previous frames. \cite{min2019tased} suggested a different model called TASED-Net, which utilizes a 3D fully-convolutional encoder-decoder network architecture where the encoded features are spatially upsampled while aggregating the temporal information. \cite{lai2019video} recently developed another two-stream spatiotemporal salieny model called STRA-Net that considers dense residual cross connections and a composite attention module. %

\begin{figure*}[!t]
\centering
\includegraphics[width=0.9\textwidth,height=0.35\textheight]{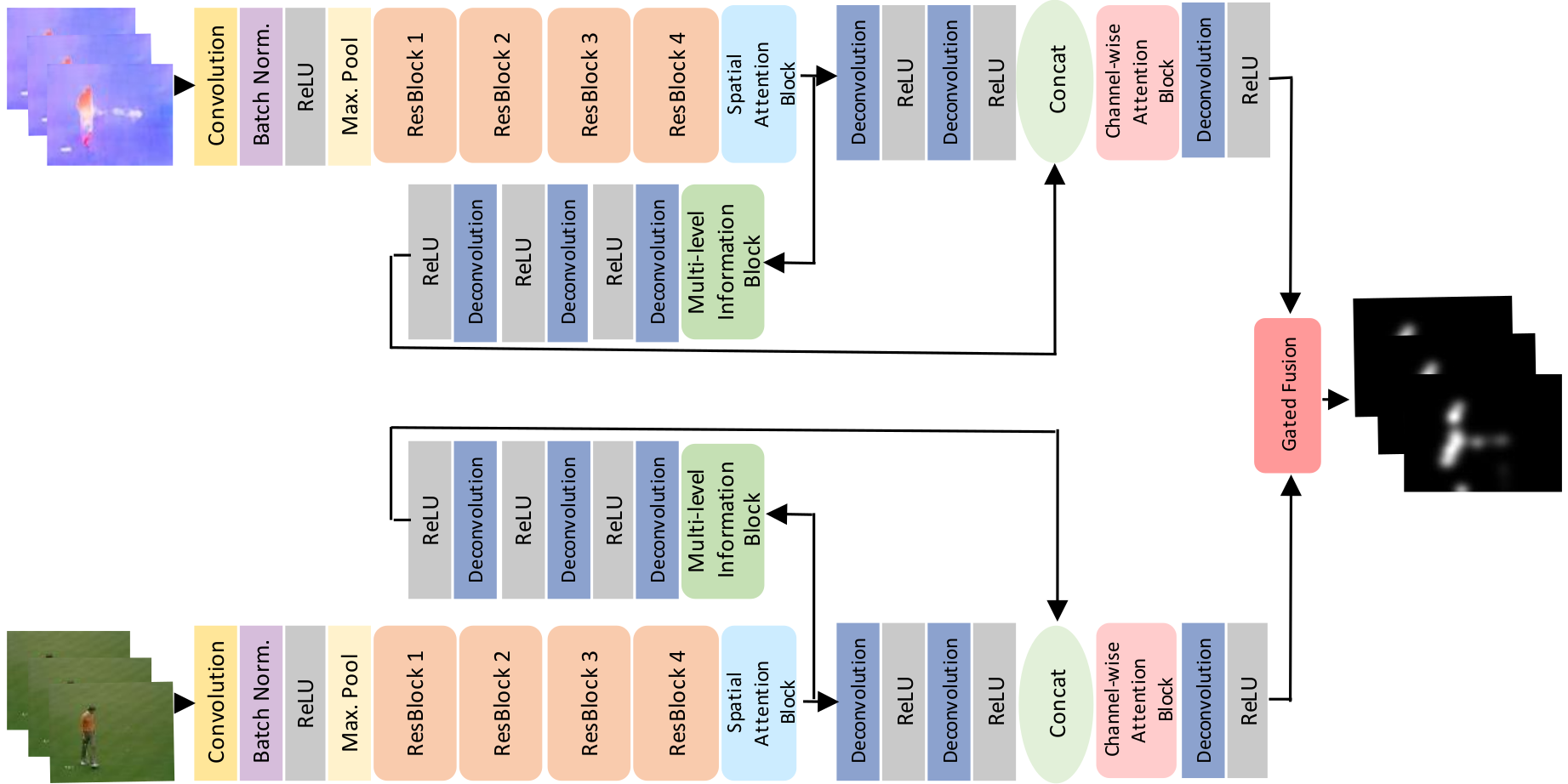}
\caption{Our two-stream dynamic saliency model uses RGB frames for spatial stream and optical flow images for temporal stream. These streams are integrated with a dynamic fusion strategy that we referred to as gated fusion. Our architecture also employs multi-level information block to fuse multi-scale features and attention blocks for feature selection.}
\label{fig:fullnet}
\end{figure*}

The aforementioned dynamic saliency models suffer from different drawbacks. The early methods employ (hand-crafted) low-level features that do not provide a high-level understanding of the video frames. Deep models eliminate this pitfall by utilizing an end-to-end learning strategy and, hence, provide better saliency predictions. They differ from each other by how they include motion information within their respective architectures. As we reviewed, the two main alternative approaches include using recurrent connections or processing data in multiple streams. Although RNN-based models help to encode temporal information with less amount of parameters, the encoding procedure compresses all the relevant information into a single vector representation, which affects the robustness especially for longer sequences. In that respect, the accuracy of the two-stream models do not, in general, degrade as the length of a sequence increases. Moreover, they are more interpretable as they need to perform fusion of spatial and temporal features in an explicit manner. On the other hand, their performance depends on accurate estimation of the optical flow maps used as input to the temporal stream. Hence, most of these two-stream models employ recent deep-learning based optical flow estimation models and even some of them uses some additional post-processing steps such as confining the absolute values of the magnitudes within a certain interval to avoid noise, as in STRA-Net~\cite{lai2019video}. Our proposed model also uses a two-stream approach, but as we will show, it exploits a novel and more dynamic fusion strategy, which boosts the performance and further improves the interpretability.

\section{Our Model}
\label{sec:ourmethod}
A general overview of our proposed spatio-temporal network architecture is given in Fig.~\ref{fig:fullnet}. We use a two-stream architecture that processes temporal and spatial information in separate streams, similar to the one in~\cite{BakEE16}. That is, we respectively feed the spatial stream and temporal stream with RGB video frames and the corresponding optical flow images as inputs. Different than~\cite{BakEE16}, however, our network combines information coming from several levels~(Section~\ref{sec:multi}) and fuses both streams via a novel dynamic fusion strategy~(Section~\ref{sec:gated}). We additionally utilize attention blocks (Section~\ref{sec:att}) to select more relevant features to further boost the performance of our model. Here, we use a pre-trained ResNet-50 model~\cite{he2015deep} as the backbone of our saliency network  as commonly explored by the previous saliency studies. In particular, we remove the average pooling and fully connected layers after the last residual block ({\sffamily {\small ResBlock4}}) and then adapt it for saliency prediction by adding extra blocks. Using ResNet-50 model allows us to  encode both low-, mid- and high-level cues in the visual stimuli in an efficient manner. Moreover, the number of network parameters is much smaller as compared to other alternative backbone networks.
 
\subsection{Multi-level Information Block}
\label{sec:multi}
As its name implies, the purpose of multi-level information block is to let the information extracted at different levels guide the saliency prediction process. It has proven to be useful that employing a multi-level/multiscale structure almost always improves the performance for many different vision tasks such as object detection~\cite{8099589}, segmentation~\cite{unet,7298965,Pinheiro}, and static saliency detection~\cite{zhao2019pyramid,Dong2018HolisticAD}. In our work, we also employ a multi-level information block to enhance feature learning capability of our model. Specifically, it allows low-, mid-, and high-level information to be fused together and to be taken into account simultaneously while making predictions.

\begin{figure}[!t]
\centering
\includegraphics[width=0.45\textwidth,height=0.12\textheight]{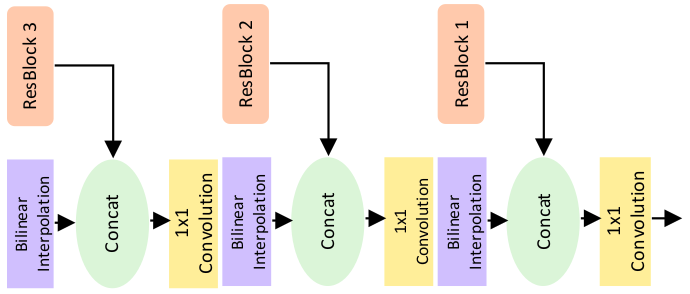}
\caption{Multi-level information block. It is used to integrate multiscale features extracted at different levels of the deep network for predicting salient parts of the given input video frame.}
\label{fig:ml_last}
\end{figure}

Fig.~\ref{fig:ml_last} shows the proposed multi-level information block that we employ in our model. This block considers low-level and high-level representations of frames by processing features maps which are extracted at each residual block. The aim is to combine primitive image features (\textit{e.g.} edges, shared common patterns) obtained at lower levels with rich semantic information (\textit{e.g.} object parts, faces, text) extracted at higher levels of the network. Here, we prefer to utilize $1\times 1$ convolution and bilinear interpolation layers to combine cues from higher and lower levels. That is, after each residual block, we expand the feature map with bilinear interpolation to make equal size of the feature map with the size of the output of the previous residual block. Then, we concatenate the expanded feature map with the previous residual block's output and fuse them via $1\times 1$ convolution layers.

 \subsection{Attention Blocks}
 \label{sec:att}
 Neural attention mechanisms allow for learning to pay attention to features more useful for a given task, and hence, it has been demonstrated many times that they can boost the performance of a neural network architecture proposed for any computer vision problem, such as object detection~\cite{att1}), visual question answering~\cite{sat}, pose estimation~\cite{pose}, image captioning~\cite{chen2016sca} and salient object detection~\cite{zhao2019pyramid}. Motivated with these observations, in our work, we integrate several attention blocks to our proposed deep architecture to let the model choose the most relevant features for the dynamic saliency estimation problem. Resembling the structures in~\cite{chen2016sca, zhao2019pyramid}, we exploit two separate attention mechanisms: 
 \emph{spatial} and \emph{channel-wise} attention, as explained below. 
 
\begin{figure}[!t]
    \centering
        \begin{subfigure}{0.18\textwidth}
        \centering
        \includegraphics[width=0.95\textwidth,height=0.09\textheight]{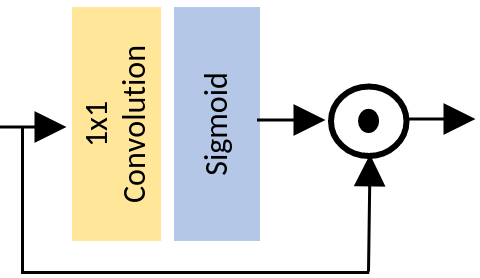}
        \label{fig:spatial_att}
        \caption{Spatial attention}
    \end{subfigure}%
    \begin{subfigure}{0.28\textwidth}
        \centering
        \includegraphics[width=0.95\textwidth,height=0.09\textheight]{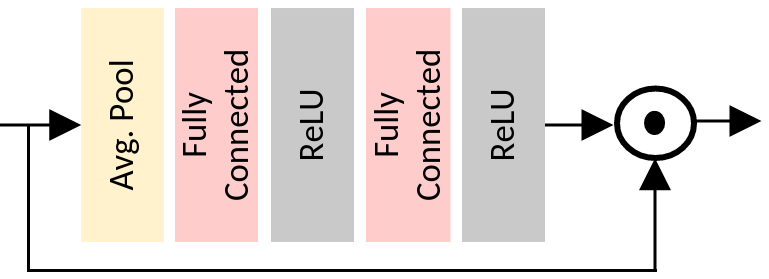}
        \label{fig:chw_att}
        \caption{Channel-wise attention}
    \end{subfigure}%
    \caption{Attention blocks: (a) spatial attention block, (b) channel-wise attention block.  While the spatial attention block defines spatial importance weights for individual feature maps, the channel-wise attention block introduces feature-level weighting which allows for a better use of context information.}
    \label{fig:attent}
\end{figure}

 Fig.~\ref{fig:attent}(a) shows our spatial attention block, which we introduce at the lower levels of our network model (see Fig.~\ref{fig:fullnet}) that helps to filter out the irrelevant information. The block takes the output of {\sffamily {\small ResBlock4}}, shaped $[B\times C\times H\times W]$ with $C=2048$, as input and it determines the important locations by calculating a weight tensor, which is shaped $[B\times 1\times H\times W]$. To estimate this tensor, input channels are fused via $1\times 1$ convolution layer following by a sigmoid layer. The output (shaped $[B\times C\times H\times W]$) of this block is a result of Hadamard product between input and spatial weight tensor. 
 
 The second type of our attention block, the channel-wise attention block, is shown in Fig.~\ref{fig:attent}(b), whose main purpose is to utilize the context information in a more efficient way. The block consists of average pooling, full connected and ReLU layers. In particular, it takes the concatenation of the feature maps from the main stream and multi-level information block as input which is shaped $[B\times 96\times H\times W]$, then downsamples it with average pooling (output shape is $[B\times 96]$). The weight of each channel is determined after two fully connected layers followed by ReLUs. The shape of the matrices are $[B\times 24]$ and $[B\times 96]$ respectively. The output of last ReLU which is shaped $[B\times 96\times 1\times 1]$, contains a scalar value to weight each channel. At the end of the block, the input feature map is weighted via Hadamard product.  

\subsection{Gated Fusion Block}
\label{sec:gated}
One of the main contributions of our framework is to employ a dynamic fusion strategy to combine temporal and spatial information. Gated fusion has been exploited before for different problems such as image dehazing~\cite{Ren-CVPR-2018}, image deblurring~\cite{Zhang2018}, semantic segmentation~\cite{8099644}. The main purpose to use a gated fusion block is to combine different kind of information with a dynamic structure which considers the current inputs' characteristics. For example, in~\cite{8099644} feature maps that are generated via RGB information and depth information is combined for solving semantic segmentation. In our case, our aim is to come up with a fusion module that considers the content of the video at inference time. To our knowledge, we are the first to provide a truly dynamic approach for dynamic saliency. As opposed to the classical learning based approaches that learn the contributions of temporal and spatial streams in a static manner from the training data, our gated fusion block performs the fusion process in an adaptive way. That is, it decides the contribution of each stream on a location- and time-aware manner according to the content of the video.

\begin{figure}[!t]
\centering
\includegraphics[width=0.42\textwidth]{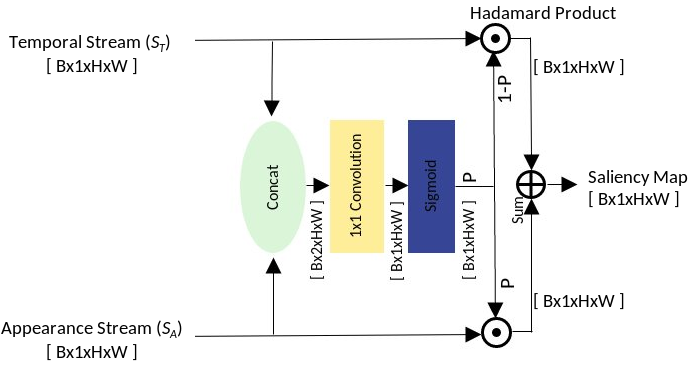}
\caption{Gated fusion block. It integrates the spatial and temporal streams to learn a weighted gating scheme to determine their contributions in predicting dynamic saliency of the current input video frame.}
\label{fig:gatedmod}
\end{figure}

\begin{figure*}[!t]
\centering
\resizebox{\linewidth}{!}{
\begin{tabular}{cc@{}c@{}c@{}c@{}|c@{}c@{}c@{}c}
\rotatebox{90}{\huge{Appearance}}&\includegraphics[width=0.25\linewidth]{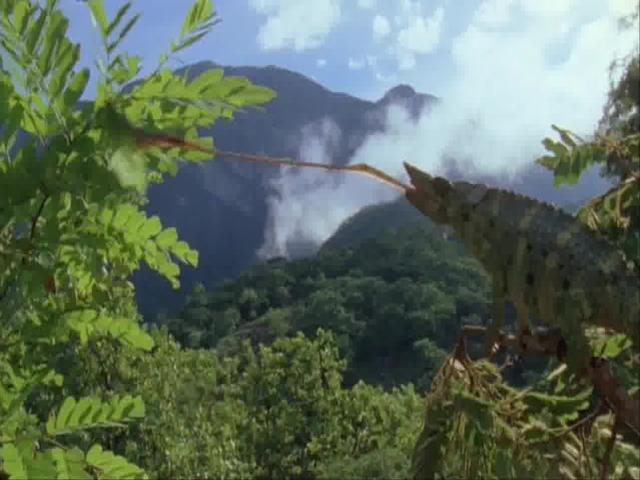} & \includegraphics[width=0.25\linewidth]{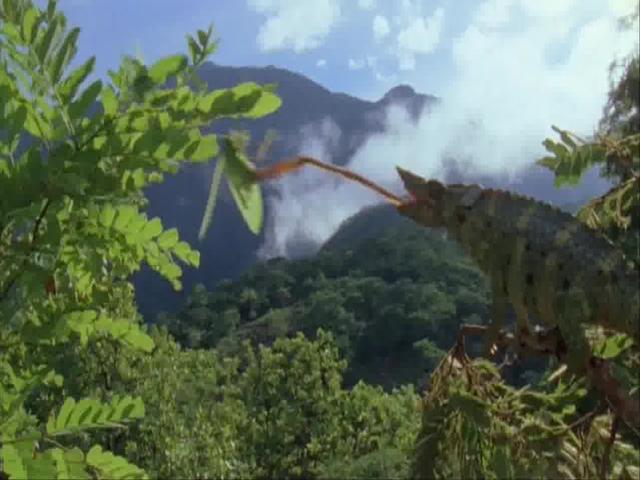} &\includegraphics[width=0.25\linewidth]{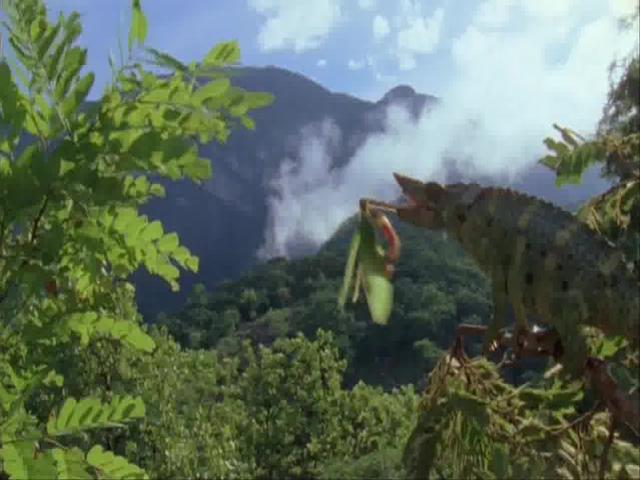} &\includegraphics[width=0.25\linewidth]{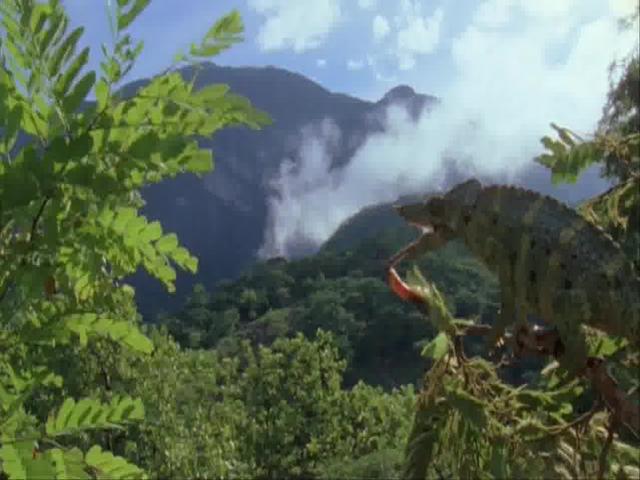} &\includegraphics[width=0.25\linewidth]{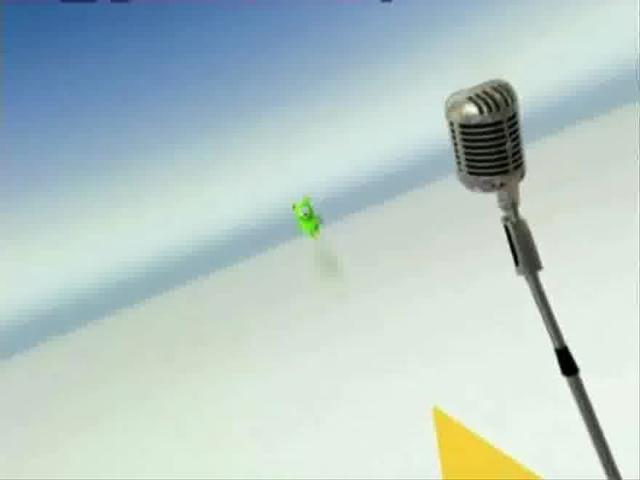} &\includegraphics[width=0.25\linewidth]{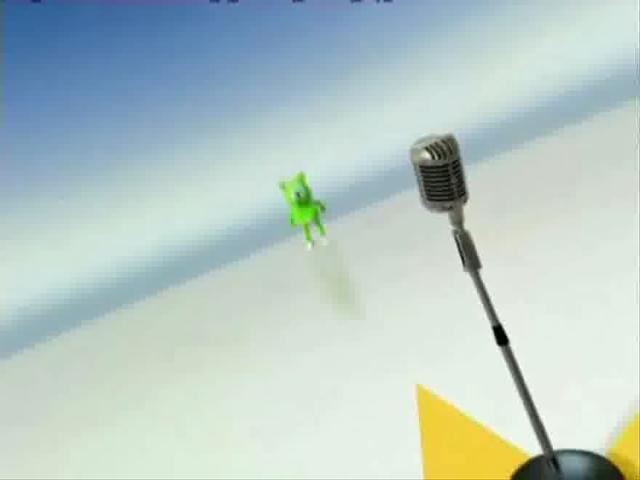} &\includegraphics[width=0.25\linewidth]{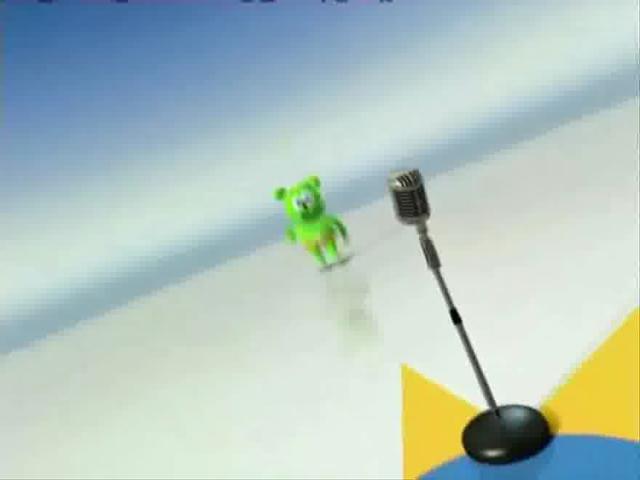} &\includegraphics[width=0.25\linewidth]{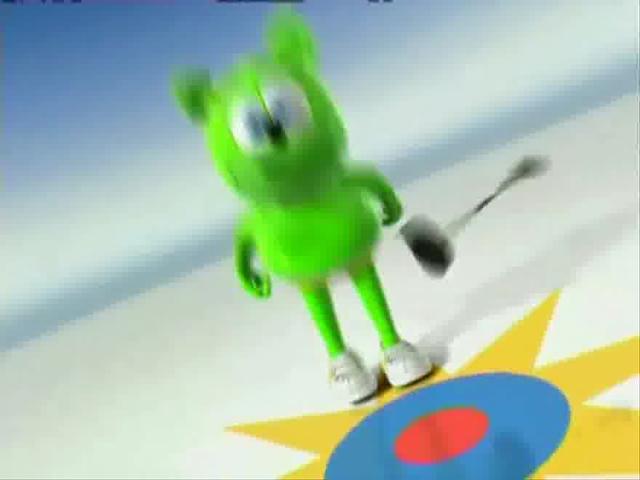}\\
\rotatebox{90}{\huge{$\;\;\,$Motion}}&\includegraphics[width=0.25\linewidth]{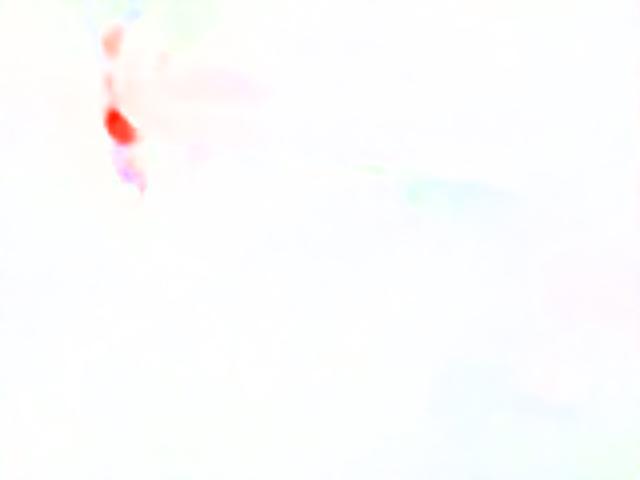} &\includegraphics[width=0.25\linewidth]{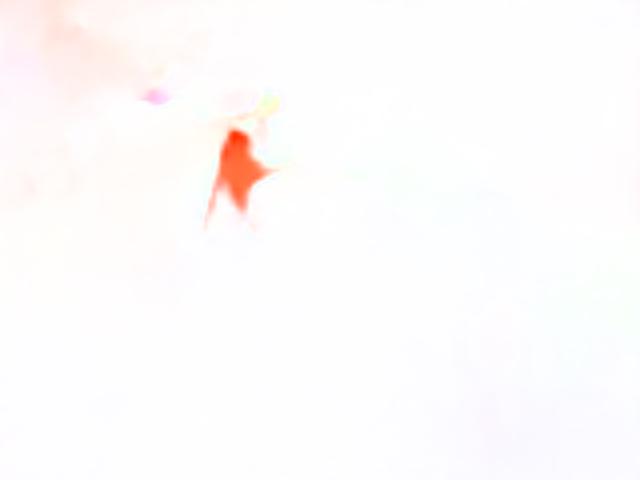} &\includegraphics[width=0.25\linewidth]{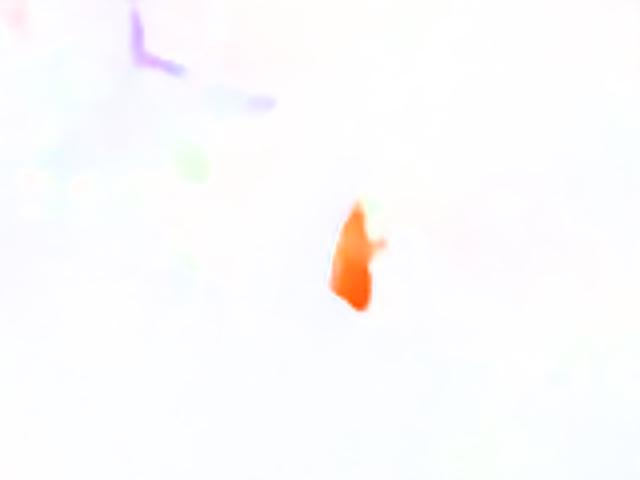} &\includegraphics[width=0.25\linewidth]{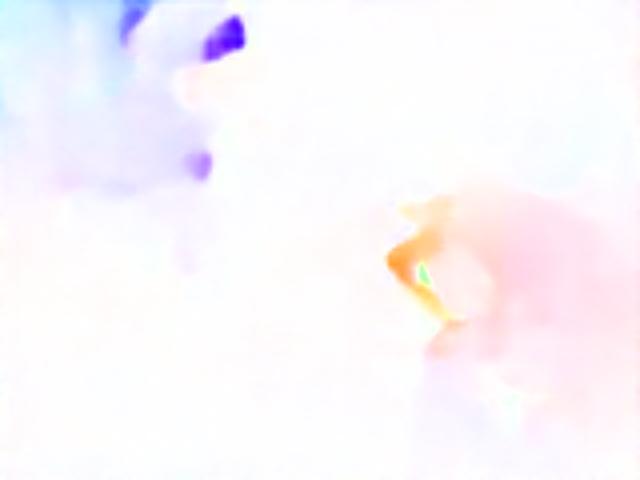} &\includegraphics[width=0.25\linewidth]{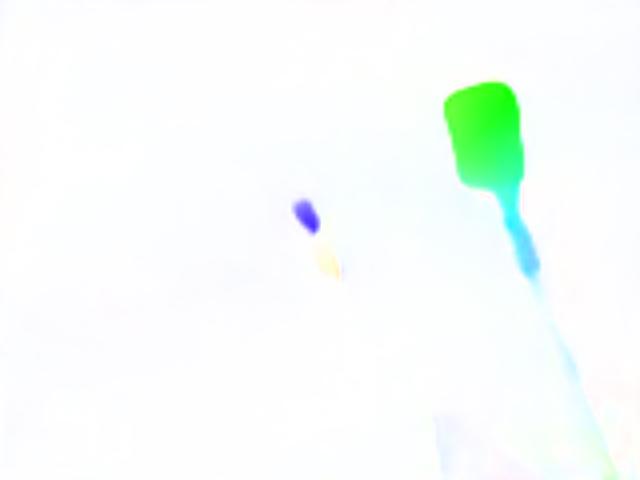} &\includegraphics[width=0.25\linewidth]{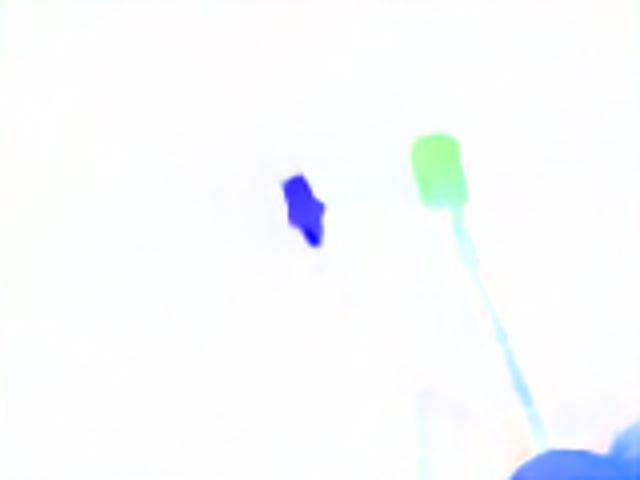} &\includegraphics[width=0.25\linewidth]{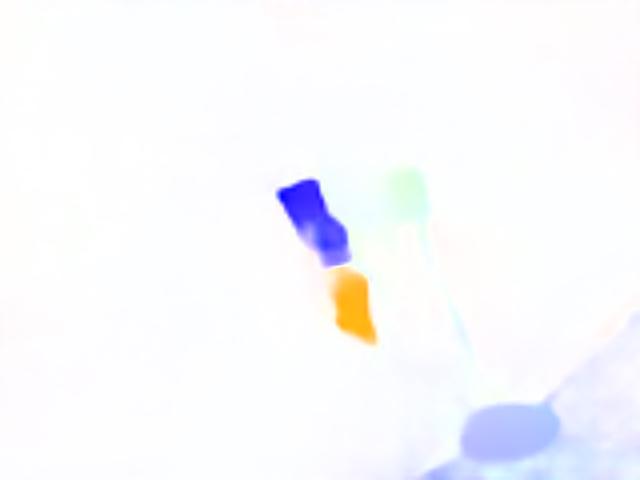} &\includegraphics[width=0.25\linewidth]{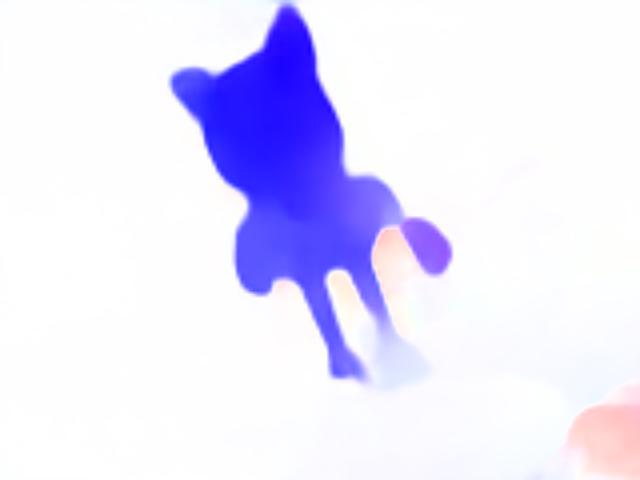} \\
\rotatebox{90}{\huge{$\quad\;\;\; S_A$}}&\includegraphics[width=0.25\linewidth]{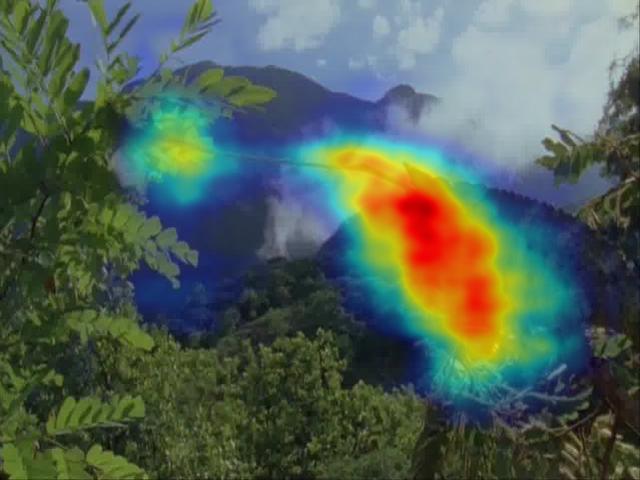} &
\includegraphics[width=0.25\linewidth]{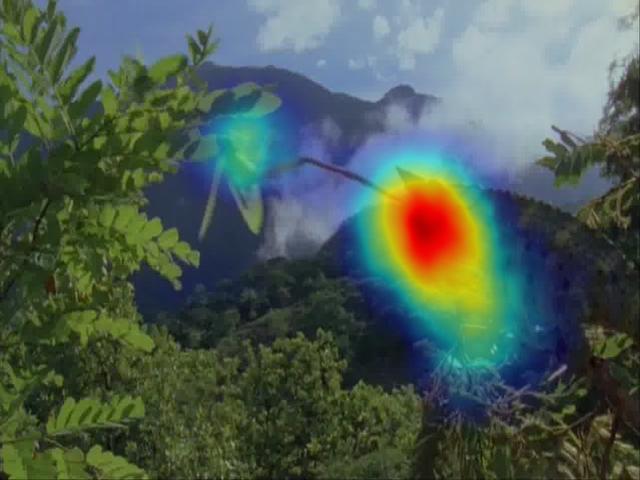} &
\includegraphics[width=0.25\linewidth]{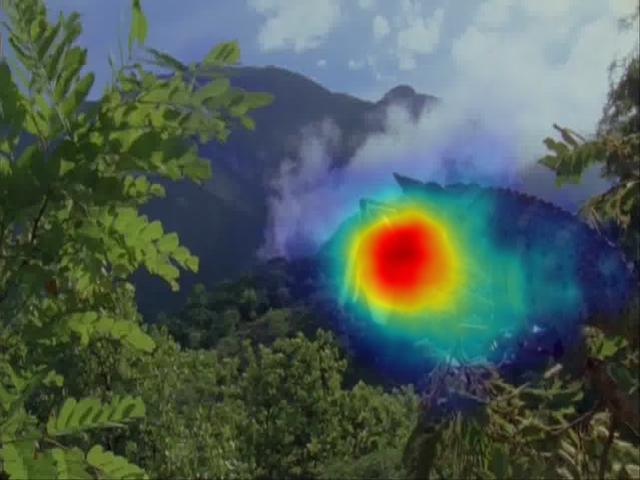} &
\includegraphics[width=0.25\linewidth]{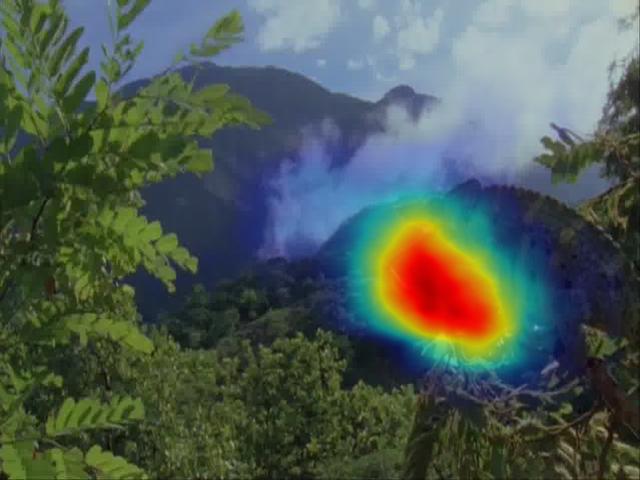} &
\includegraphics[width=0.25\linewidth]{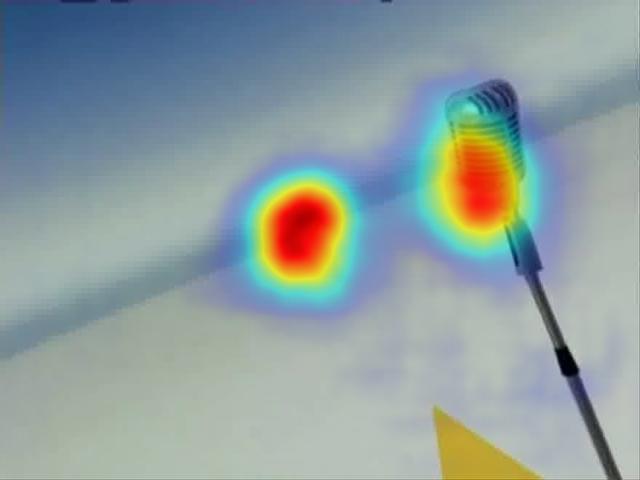} &
\includegraphics[width=0.25\linewidth]{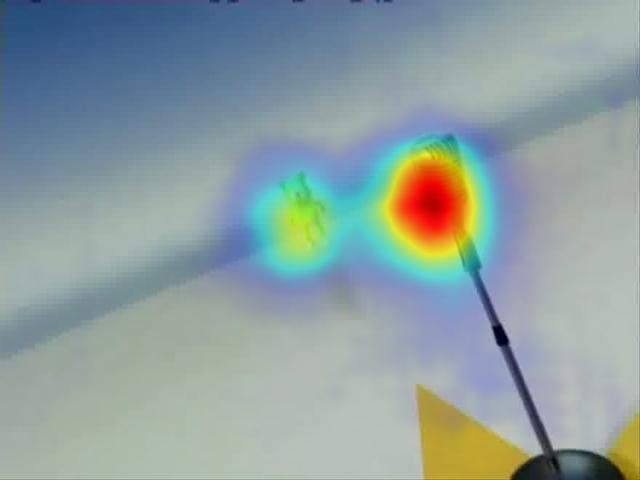} &
\includegraphics[width=0.25\linewidth]{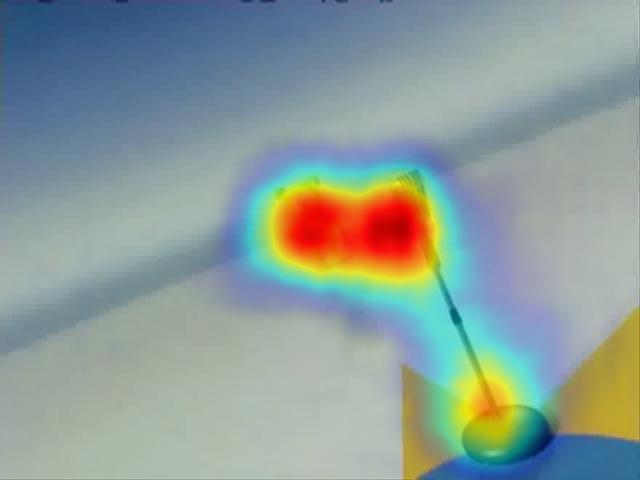} &
\includegraphics[width=0.25\linewidth]{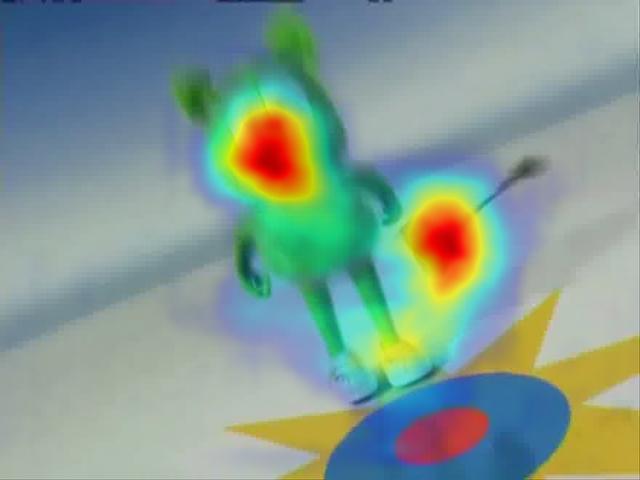} \\
\rotatebox{90}{\huge{$\quad\;\;\; G_A$}}&\includegraphics[width=0.25\linewidth]{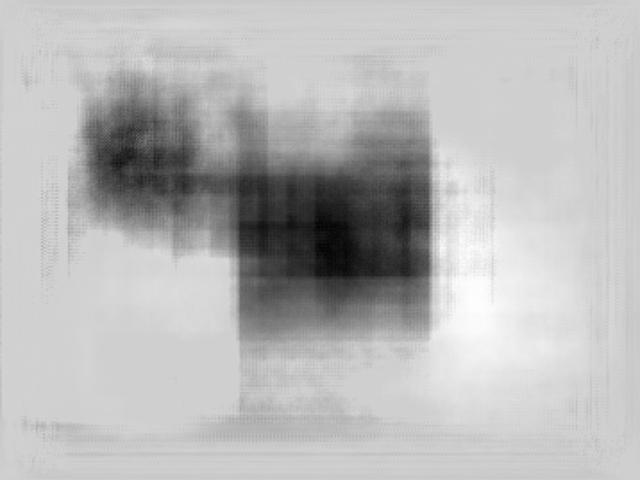} &
\includegraphics[width=0.25\linewidth]{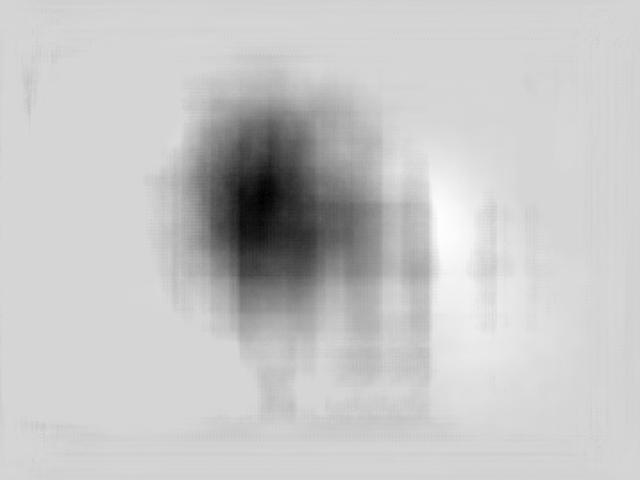} &
\includegraphics[width=0.25\linewidth]{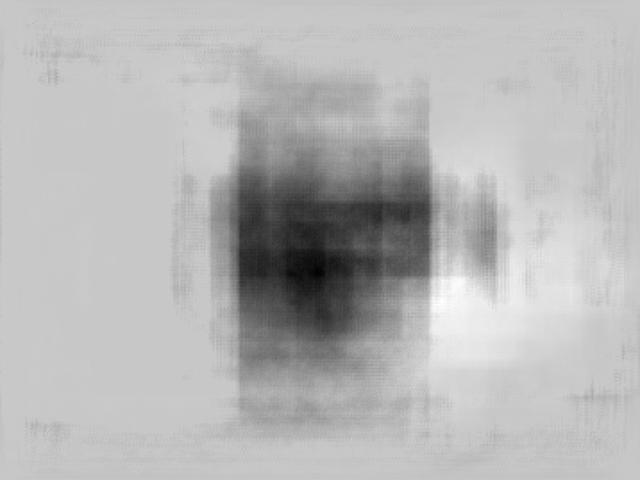} &
\includegraphics[width=0.25\linewidth]{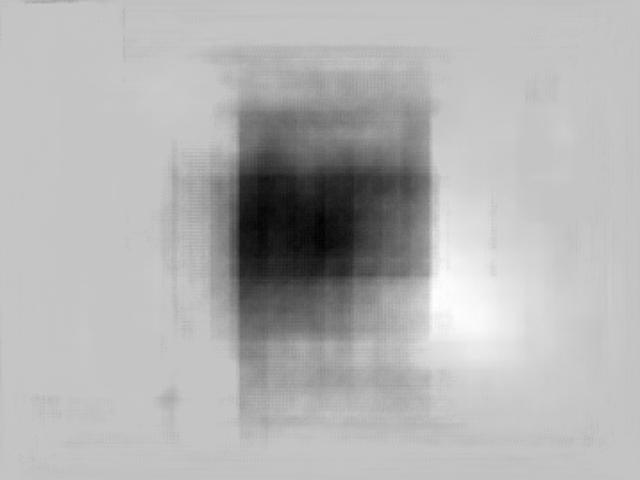} &
\includegraphics[width=0.25\linewidth]{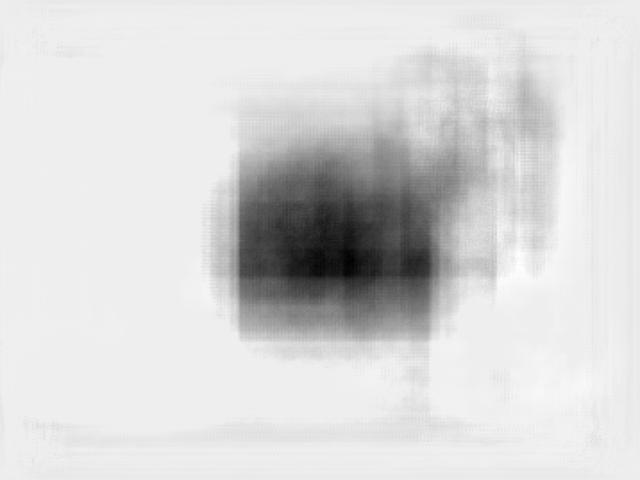} &
\includegraphics[width=0.25\linewidth]{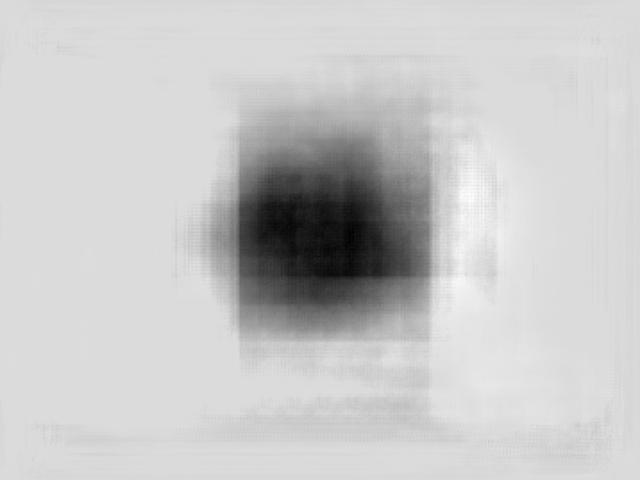} &
\includegraphics[width=0.25\linewidth]{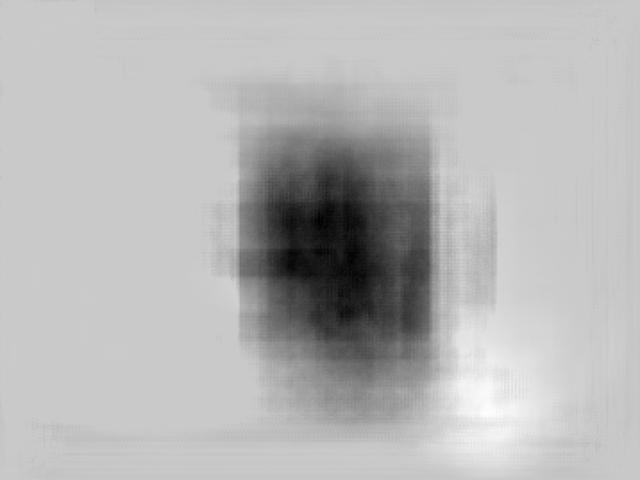} &
\includegraphics[width=0.25\linewidth]{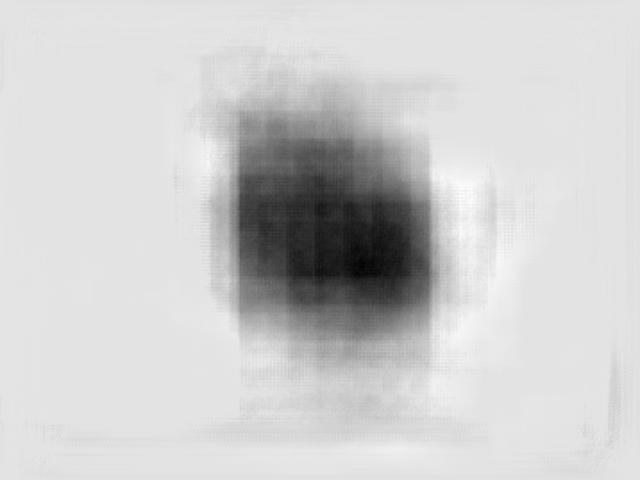} \\

\rotatebox{90}{\huge{$\quad\;\;\; S_T$}}&\includegraphics[width=0.25\linewidth]{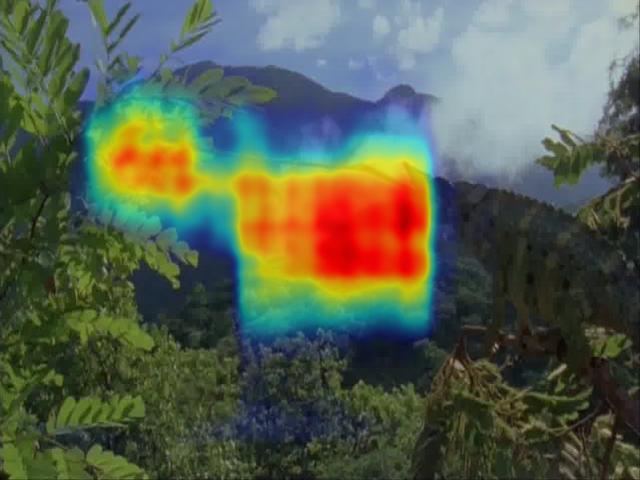} &
\includegraphics[width=0.25\linewidth]{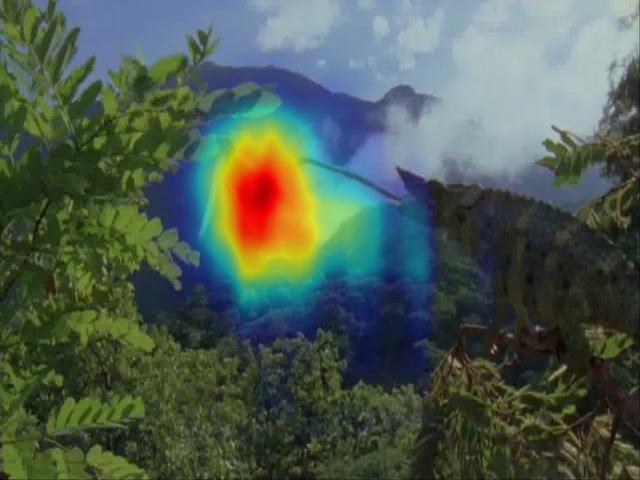} &
\includegraphics[width=0.25\linewidth]{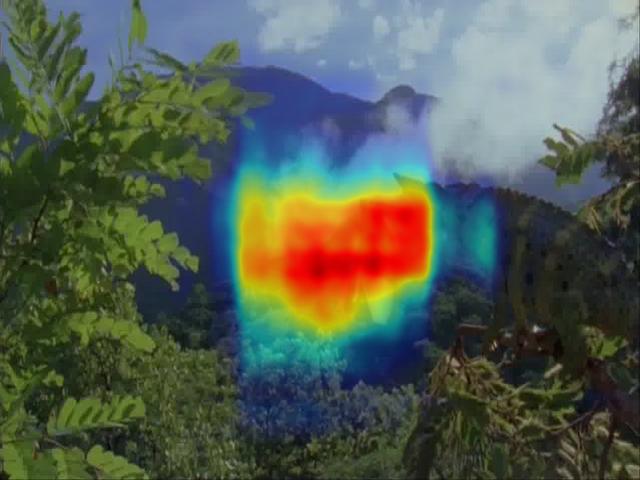} &
\includegraphics[width=0.25\linewidth]{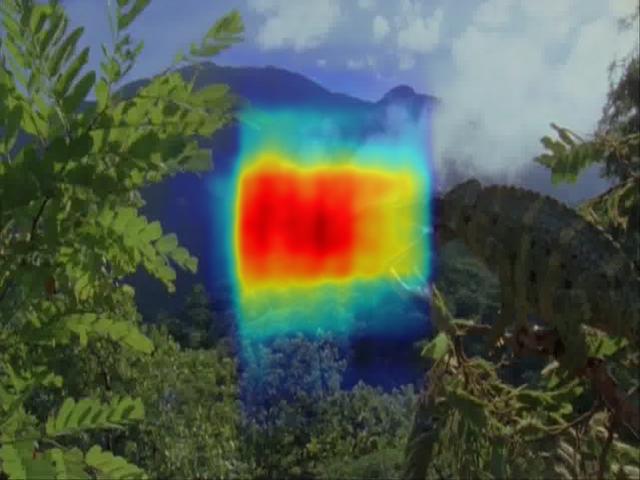} &
\includegraphics[width=0.25\linewidth]{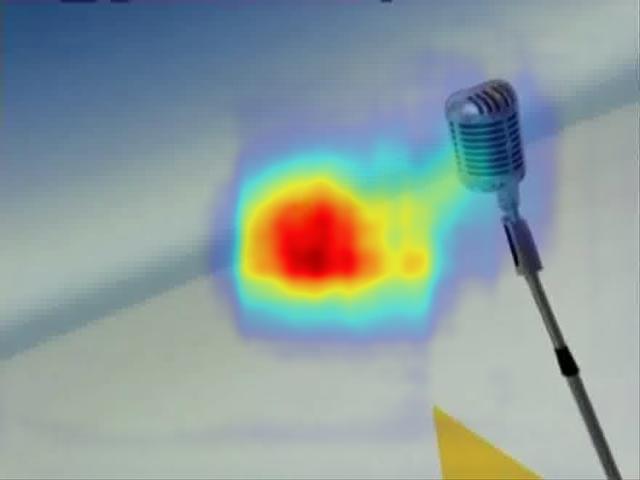} &
\includegraphics[width=0.25\linewidth]{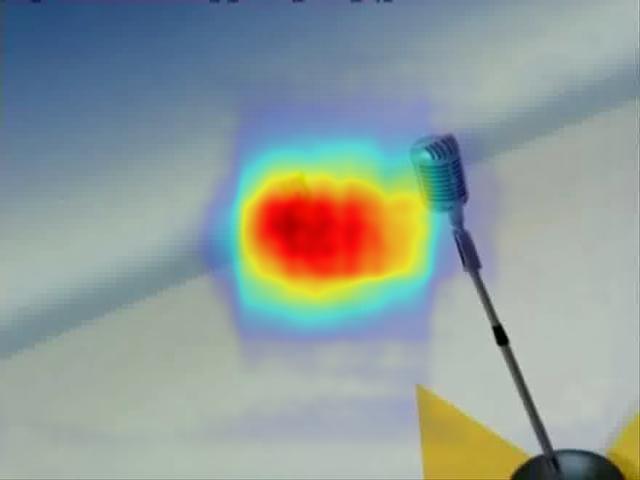} &
\includegraphics[width=0.25\linewidth]{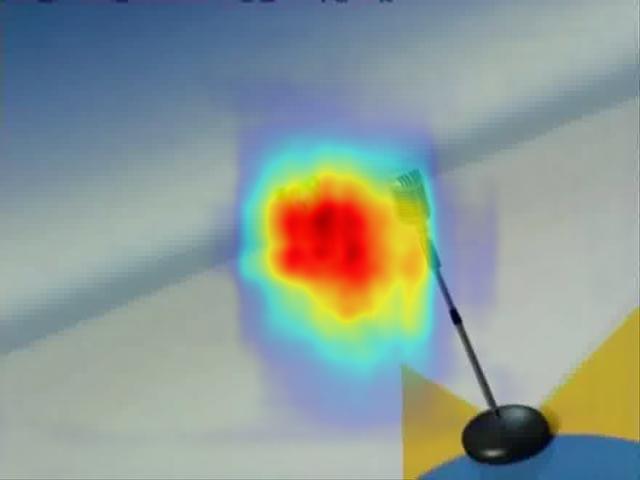} &
\includegraphics[width=0.25\linewidth]{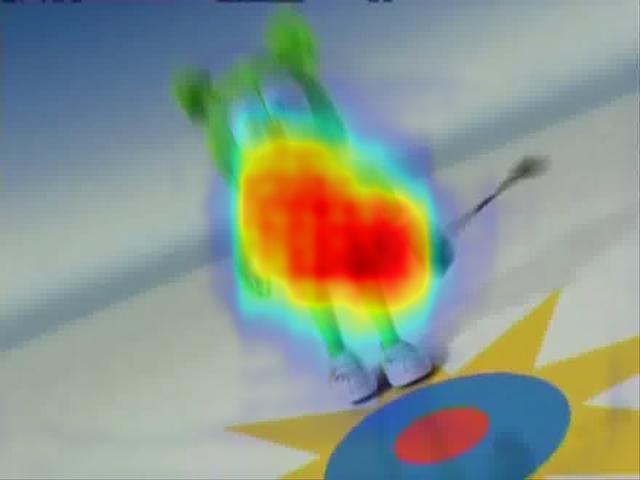} \\

\rotatebox{90}{\huge{$\quad\;\;\; G_T$}}&\includegraphics[width=0.25\linewidth]{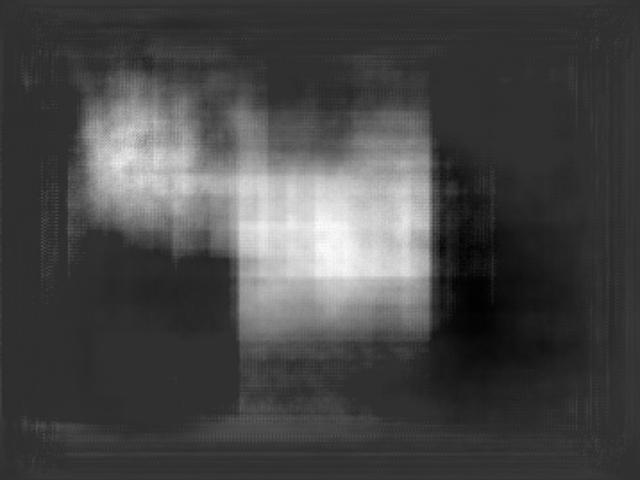} &
\includegraphics[width=0.25\linewidth]{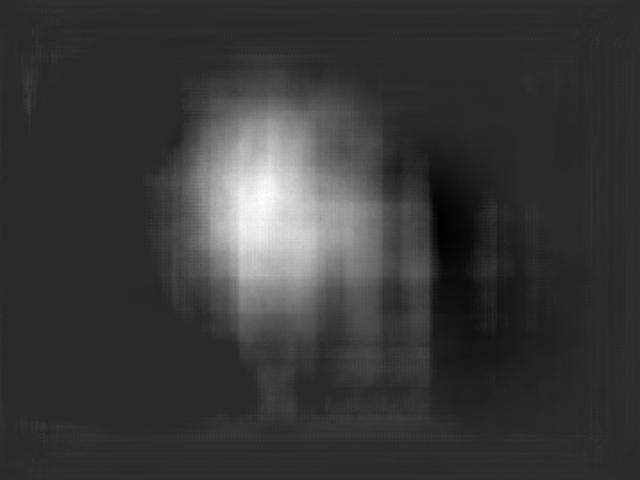} & 
\includegraphics[width=0.25\linewidth]{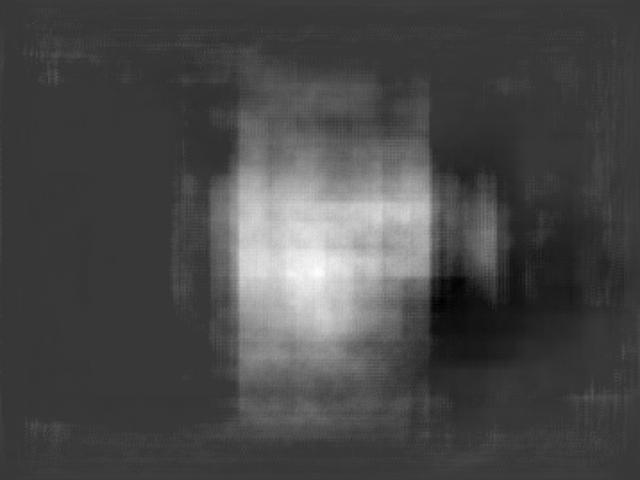} & 
\includegraphics[width=0.25\linewidth]{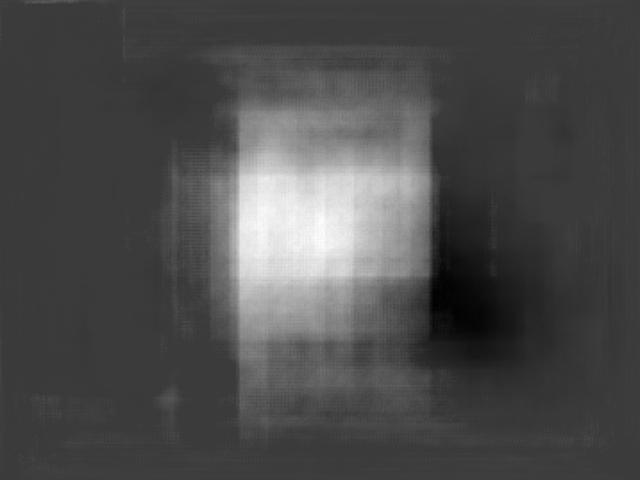} & 
\includegraphics[width=0.25\linewidth]{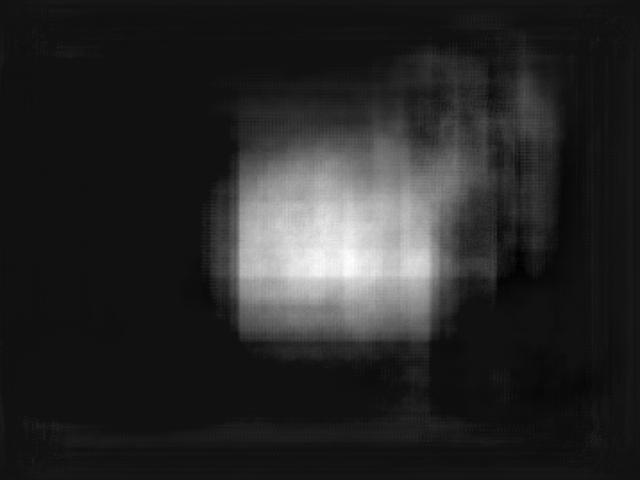} & 
\includegraphics[width=0.25\linewidth]{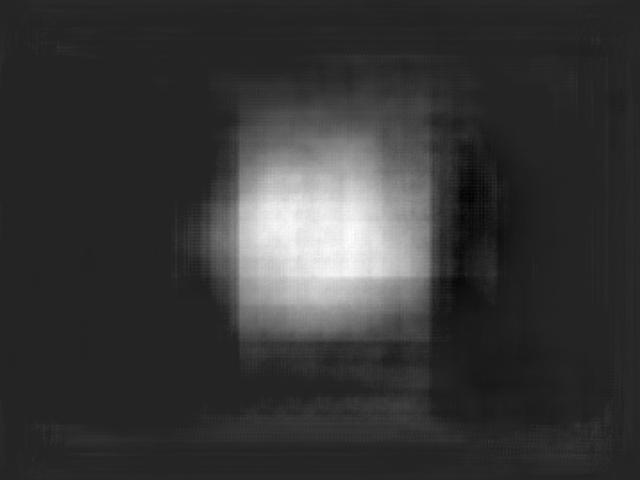} & 
\includegraphics[width=0.25\linewidth]{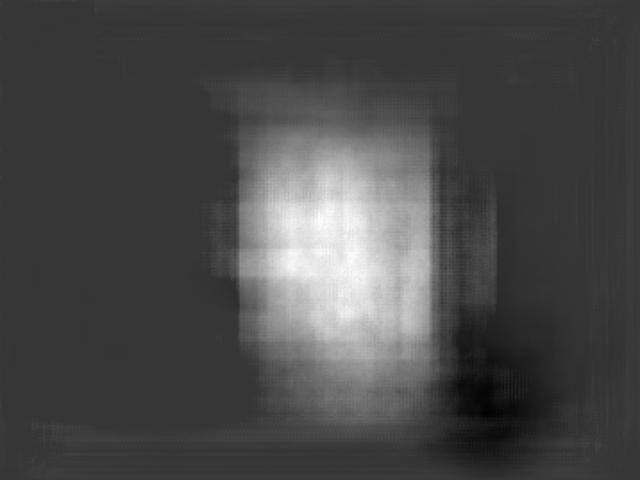} & 
\includegraphics[width=0.25\linewidth]{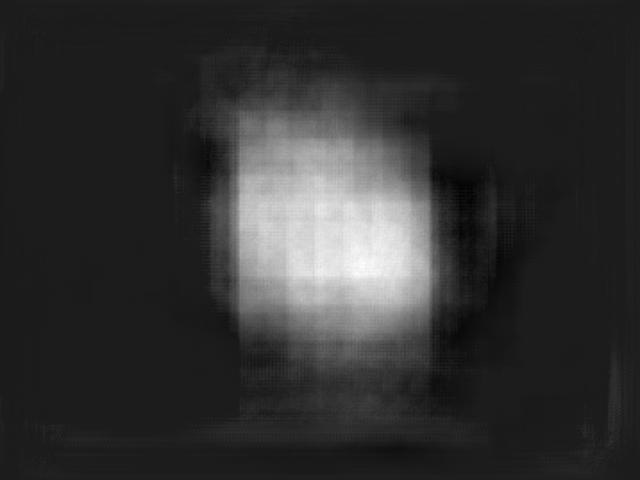} \\

\rotatebox{90}{\huge{Prediction}}&\includegraphics[width=0.25\linewidth]{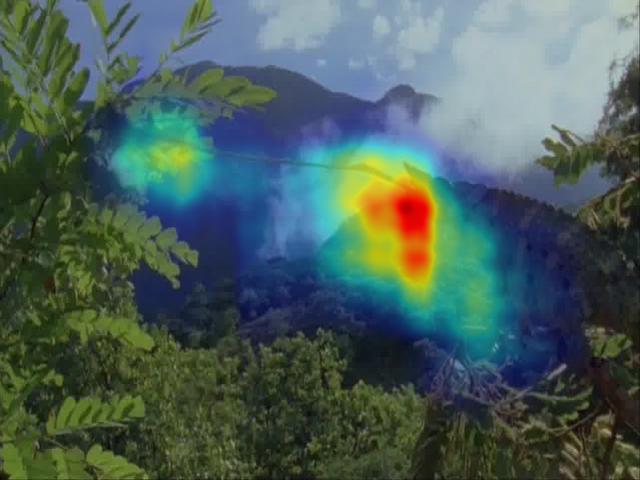} &
\includegraphics[width=0.25\linewidth]{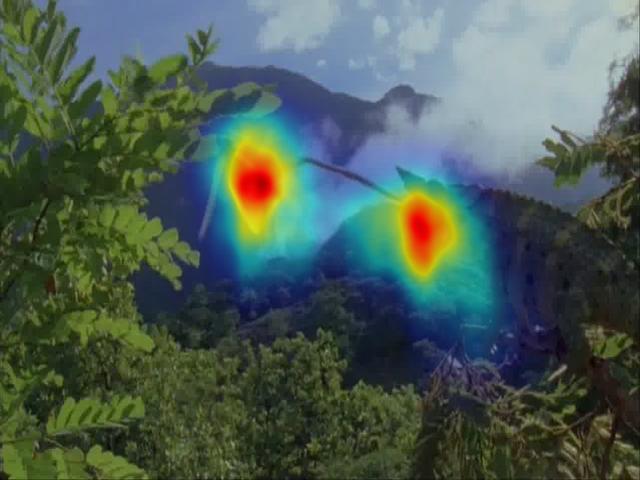} &
\includegraphics[width=0.25\linewidth]{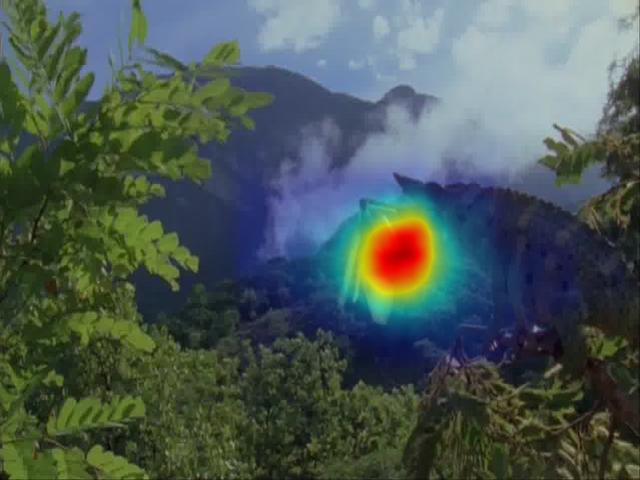} &
\includegraphics[width=0.25\linewidth]{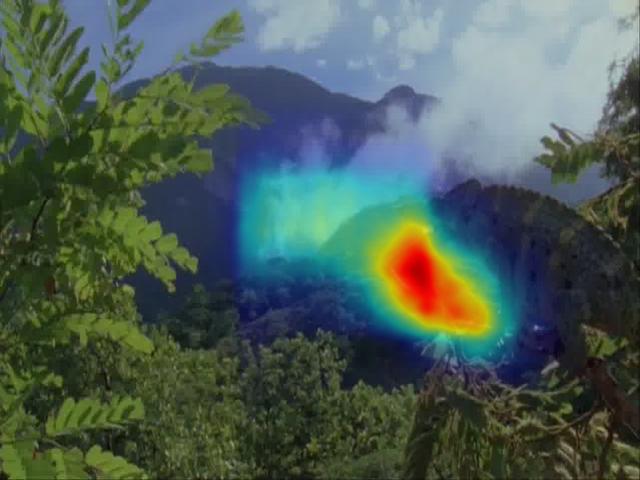} &
\includegraphics[width=0.25\linewidth]{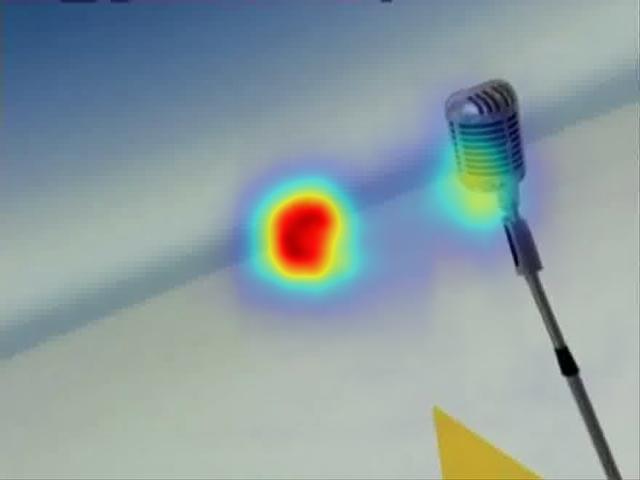} &
\includegraphics[width=0.25\linewidth]{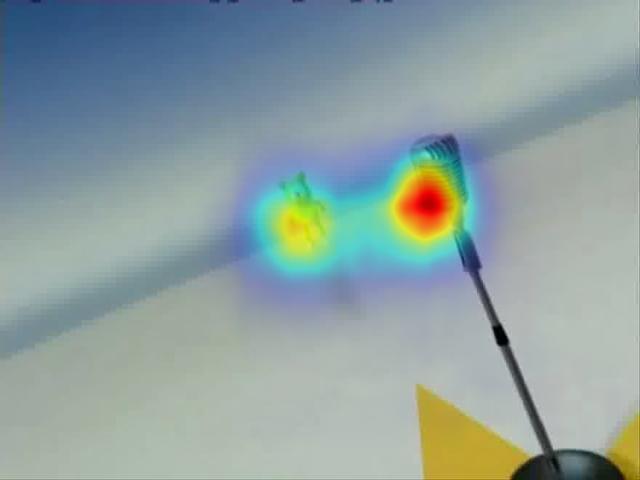} &
\includegraphics[width=0.25\linewidth]{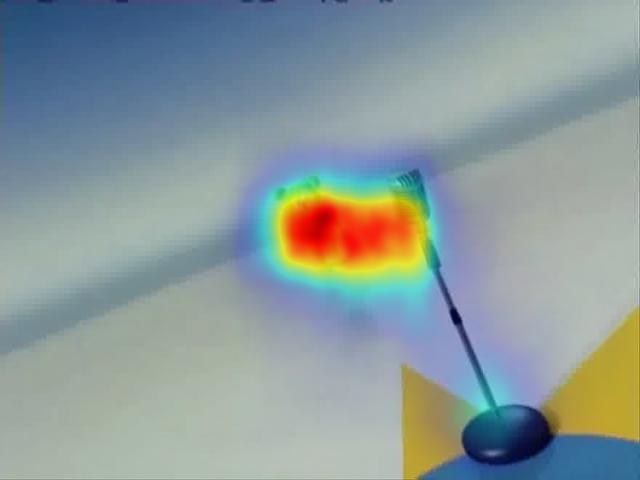} &
\includegraphics[width=0.25\linewidth]{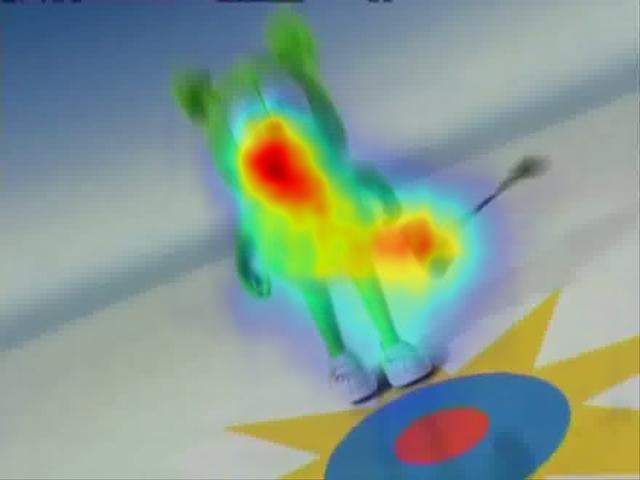} \\

\rotatebox{90}{\huge{$\qquad$GT}}&\includegraphics[width=0.25\linewidth]{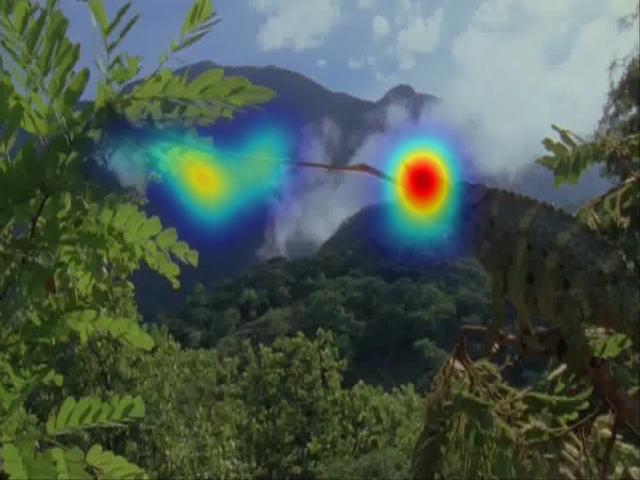} &
\includegraphics[width=0.25\linewidth]{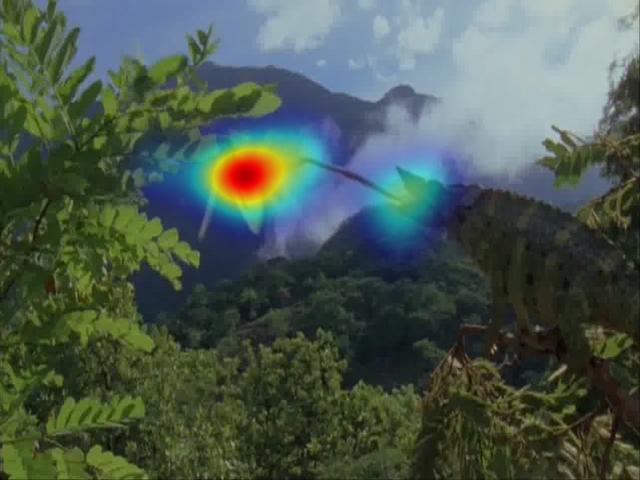} &
\includegraphics[width=0.25\linewidth]{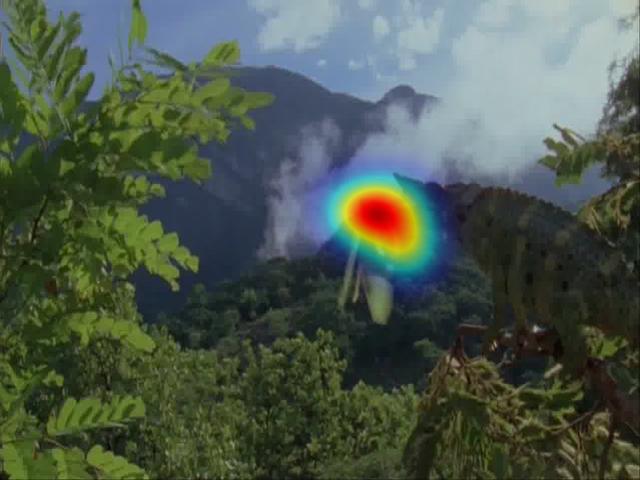} &
\includegraphics[width=0.25\linewidth]{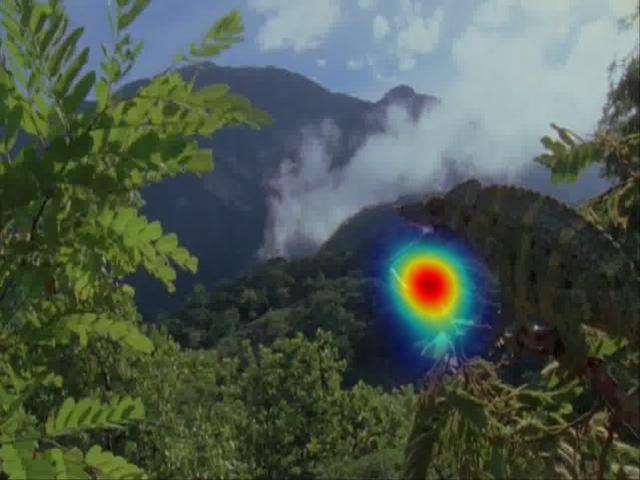} &
\includegraphics[width=0.25\linewidth]{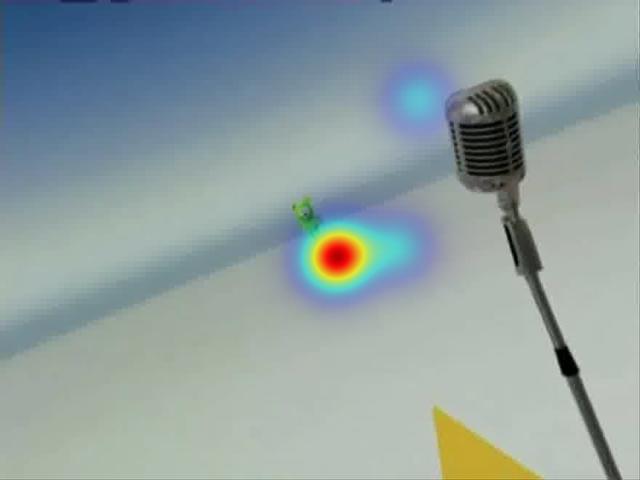} &
\includegraphics[width=0.25\linewidth]{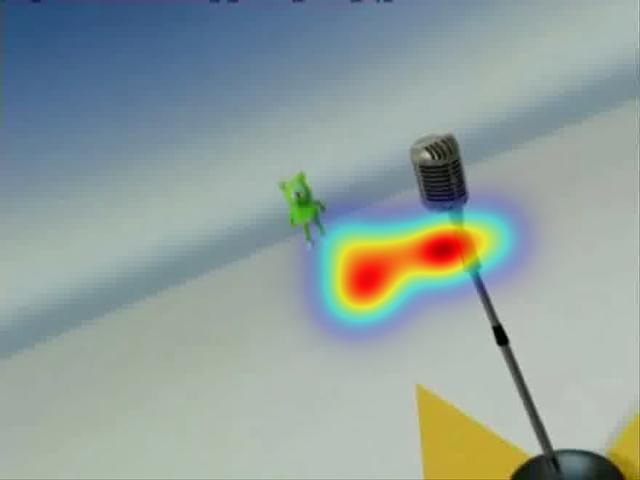} &
\includegraphics[width=0.25\linewidth]{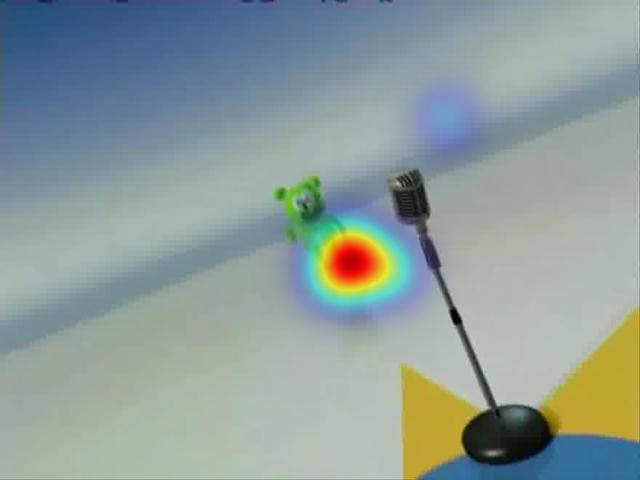} &
\includegraphics[width=0.25\linewidth]{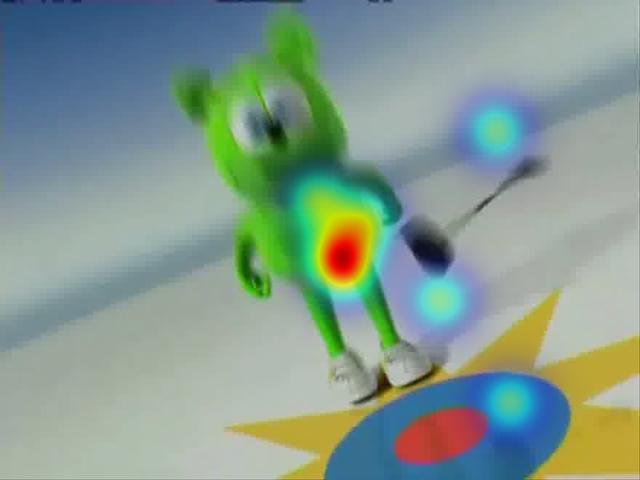}
\end{tabular}}
\caption{Gated fusion block estimates the final saliency map by combining the appearance and the temporal maps $S_A$ and $S_T$ with the spatially varying weights $G_A$ and $G_T$.}
\label{fig:gated_detail}
\end{figure*}

The structure of the proposed gated fusion block is shown in Fig.~\ref{fig:gatedmod}. It takes the feature maps of the spatial and temporal streams as inputs and produces a probability map which is used to designate contribution of each stream with regard to their current characteristics. Let $S_{A}$, $S_{T}$ denote the feature maps from spatial and temporal streams, respectively. Gated fusion module first concatenates these features and then learns their correlations by applying a $1\times 1$ convolution layer. After that, it uses a sigmoid layer to regularize the feature map which is used to estimate weights of the gate. Let $G_{A}$ and $G_{T}$ denote how confidently we can rely on appearance and motion, respectively, as follows:
\begin{equation} 
G_{A} = P \;, \quad G_{T} =1-P \;,
\end{equation}
where $P$ is the output of the sigmoid layer. Then, gated fusion module estimates the weights denoting the contributions of the spatial and temporal streams, as given below:
\begin{equation} 
 S'_{A} = S_{A} \odot G_{A} \;, \quad
  S'_{T} = S_{T} \odot G_{T} \;, 
\end{equation}
where $\odot$ represents the Hadamard product operation. Finally, it generates the final saliency map, $ S_{final}$, via weighting the appearance and temporal streams' feature maps with the estimated probability map:
\begin{equation}
    S_{final}= S'_{A} +  S'_{T} \;.
    \label{eq:eq5}
\end{equation}

Fig.~\ref{fig:gated_detail} visualizes how gated fusion block works. While the appearance stream computes a saliency map from the RGB frame, the temporal stream extracts a saliency map from the optical image obtained from successive frames. As can be seen, these intermediate maps encode different characteristic of the input dynamic stimuli. The appearance based saliency map mostly focuses on the regions that have distinct visual properties than theirs surroundings, whereas the motion based saliency map mainly pay attention to motion. Gated fusion scheme estimates spatially varying probability maps and employs them to integrate the appearance and temporal streams, which results in more confident predictions. The spatial stream generally gives more accurate predictions than the temporal stream, as will be presented in the Experiments section. On the other hand, as can be seen from the estimated weight maps, the gated fusion scheme in the proposed model has a tendency to pay more attention to the temporal stream. We suspect that this is because the model considers that it may carry auxiliary information.

\section{Experiments}
\label{sec:experiments}
In the following, we first provide a brief review of the benchmark datasets we used in our experimental analysis. Then, we give the details of our training procedure including the loss functions and settings we use to train our proposed model. Next, we summarize the evaluation metrics and the dynamic saliency models used in our experiments. We then discuss our findings and present some qualitative and quantitative results. Finally, we present an ablation study to evaluate the effectiveness of the blocks of the proposed dynamic saliency model.

\subsection{Datasets}
\label{ssec:datasets}
In our experiments, we employ six different datasets to evaluate the effectiveness of the proposed saliency model. The first four, namely UCF-Sports~\cite{RodriguezAS08}, \mbox{Holywood-2}~\cite{MatheSminchisescuPAMI2015}, DHF1K~\cite{wang2018revisiting}, and DIEM~\cite{Mital}, are the most commonly used benchmarks. Among them, we specifically utilize DIEM to test the generalization ability of our model. The last two datasets considered in our analysis, \mbox{DIEM-Meta}~\cite{temporalsal} and LEDOV-Meta~\cite{temporalsal},  are two recently proposed datasets, which are particularly designed to explore the performance of a dynamic saliency model under situations where understanding temporal effects is critical to give results more compatible with humans.\\

\noindent\textbf{UCF-Sports} dataset~\cite{RodriguezAS08} is the smallest dataset in terms of its size, consisting of 150 videos obtained from 13 different action
classes. It is originally collected for action recognition, but then enriched by~\cite{MatheSminchisescuPAMI2015} to include eye fixation data. The videos are annotated by 4 subjects under free-viewing condition. In the experiments, we used the same train/test splits given in~\cite{ucfsportsdataset}.

\noindent\textbf{Holywood-2} dataset~\cite{MatheSminchisescuPAMI2015} contains 1,707 videos from Hollywood-2 action recognition dataset~\cite{5206557}, among which 823 are used for training and the remaining 884 are left for testing. Since the videos are collected from 69 Hollywood movies with 12 action categories, its content is limited to human actions. In~\cite{MatheSminchisescuPAMI2015}, the authors collected human fixation data for each sequence from 3 subjects under free-viewing condition. In our experiments, we use all train and test frames.

\noindent\textbf{DHF1K}~\cite{wang2018revisiting} is the most recent and the largest video saliency dataset, which contains a total of 1000 videos with eye tracking data collected from 17 different human subjects. The authors split the dataset into 600 training, 100 validation videos and 300 test videos. The ground truth fixation data for the test split is intentionally kept hidden and the evalution of a model on the test data is carried out by the authors themselves.

\noindent\textbf{DIEM}~\cite{Mital} includes 84 natural videos. Each video sequence has eye fixation data collected from approximately 50 different human subjects. Following the common experimental setup first considered in~\cite{rudoy2013learning}, we used all frames from 64 videos for training and the first 300 frames from the remaining 20 videos as test set.

\noindent\textbf{DIEM-Meta}~\cite{temporalsal} and \noindent\textbf{LEDOV-Meta}~\cite{temporalsal} are two recently proposed datasets collected from the existing video saliency datasets DIEM~\cite{Mital} and LEDOV~\cite{Jiang_2018_ECCV}, respectively. The main difference between these and the aforementioned datasets lies in the characteristics of the video frames they consider. \cite{temporalsal} constructed these so-called meta-datasets by eliminating the video frames from their original counterparts where spatial patterns are generally enough to predict where people look. To detect them, they employ a deep static saliency model that they developed. DIEM-Meta and DIEM-Meta are thus better testbeds for evaluating whether or not a dynamic saliency model learns to use the temporal domain effectively. DIEM-Meta contains only 35\% of the video frames from DIEM, LEDOV-Meta includes just 20\% of the original LEDOV frames.

\subsection{Training Procedure}
\label{ssec:training_procedure}
As we mentioned previously, our network takes RGB video frames and optical flow images as inputs. We extract the frames from the videos by considering their original frame rate. We employ these RGB frames to feed our appearance stream. For the temporal stream, we generate the optical flow images between two consecutive frames by using PWC-Net~\cite{Sun2018PWC-Net}. We resize all the input images to $640\times480$ pixels and map the ground truth fixation points accordingly.

Instead of training our dynamic saliency network from scratch, we first train the subnet for the appearance stream on SALICON dataset~\cite{7298710}. Then, we initialize the weights of both of our subnets for spatial and temporal streams with this pre-trained static saliency model and finetune our whole two-stream network model using the dynamic saliency datasets described above. Pre-training on static data allows our dynamic saliency model to converge in fewer epochs when trained on dynamic stimuli. We use Kullback-Leibler (KL) divergence and Normalized Scanpath Saliency (NSS) loss functions (which we will explain in detail later) with Adam optimizer during the training process. We set the initial learning rate to 10e-5 and reduce it to one tenth in every 3000th iteration. The batch size is set to 8 for UCF-Sports and 16 for the other video datasets. We train our model on NVIDIA V100 GPUs (3$\times$GPUs) and while one epoch takes approximately 2 days for the larger datasets of DHF1K, DIEM and Hollywood-2, it takes approximately 2 hours for UCF-Sports. We train our models for 2-3 epochs. Our (unoptimized) Pytorch implementation achieves a near real-time performance of 8.2 fps for frames of size $640 \times 480$ on a NVidia Tesla K40c GPU.

For our experiments on standard benchmark datasets, we consider two different training settings for dynamic stimuli. In our first setting, we use the training split of the dataset under consideration to train our proposed model. On the other hand, in our second setting, we utilize a combined training set containing training sequences from both UCF-Sports, Hollywood-2 and DHF1K datasets. The second setting further allows us to test the generalization ability of our model on DIEM, DIEM-Meta and LEDOV-Meta datasets.\\

\noindent \textbf{Loss functions}. In our work, we employ the combination of KL-divergence and NSS loss functions to train our proposed dynamic saliency model. As explored in previous studies,~\cite{huang2015salicon, wang2018revisiting}, considering more than one loss function during training, in general, improves the model performance.  Moreover, empirical experiments on the analysis of the existing automatic evaluation metrics in~\cite{8315047} have shown that KL-divergence and NSS are good choices for evaluating saliency models.

Let $P$ denote the predicted saliency map, $F$ represent ground truth (binary) fixation map collected from human subjects and $S$ be the ground truth (continuous) fixation density map which is generated by blurring fixation maps with a small Gaussian kernel. 

KL-divergence is a widely used metric to compare two probability distributions. It has been proven to be effective for evaluating and trainig the performance of saliency models where the ground truth fixation map $S$ and the predicted saliency map $P$ are interpreted as probability distributions. Formally, KL-divergence loss function is defined as:
\begin{equation}
   \mathcal{L}_{KL}(P,S) = \sum_{i}S(i)log\left(\frac{S(i)}{P(i)}\right)\;.
\end{equation}

NSS is a location based metric which is computed as the average of the normalized predicted saliency values at fixated locations that is provided with the ground truth. By using this metric as a loss function, we force the saliency model to better detect the fixation locations and assign high likelihood scores to those pixel locations. This loss function is defined as below: 
\begin{equation}
    \mathcal{L}_{NSS}(P,F) = -\frac{1}{N}\sum_{i}\bar{P}(i)\times F(i)\;,
\end{equation}
where $N$ is the total number of fixated pixels $\sum_{i} F(i)$ and $\bar{P}$ is the normalized saliency map $\frac{P-\mu(P)}{\sigma(P)}$.\\

Our final loss function is then defined as:
\begin{equation}
    \mathcal{L}(P,F,S) = \alpha \mathcal{L}_{KL}(P,S) \; +\; \beta \mathcal{L}_{NSS}(P,F)\;,
\end{equation}
\noindent
where $\mathcal{L}_{KL}$ is the KL loss function, $\mathcal{L}_{NSS}$ is the NSS loss function, and $\alpha$ and $\beta$ are the weights for these loss functions. We first perform a set of experiments on SALICON dataset to empirically determine the optimal values of $\alpha$ and $\beta$, and then set $\alpha=1$ and $\beta=0.1$ for all the experiments.  

\subsection{Evaluation Metrics and Compared Saliency Models}
In our evaluation, we employ the following five commonly reported saliency metrics:  Area Under Curve (AUC-Judd), Pearson’s Correlation Coefficient (CC), Normalized Scanpath Saliency (NSS), Similarity Metric (SIM) and KL-divergence (KLDiv). For a detailed analysis of these metrics and their definitions, please refer to~\cite{8315047}. Each metric measures a different aspect of visual saliency and none of them is superior to the others. AUC metric considers the saliency map as classification map. A ROC curve is constituted by measuring the true and false positive rates under different binary classifier thresholds. While a score of 1 indicates a perfect match, a score close to 0.5 indicates the performance of chance. NSS is another commonly used metric, which we formally defined before while describing our loss functions. CC metric is a distribution based metric which is used to measure the linear relationship between saliency and fixation maps using the following formula:
\begin{equation}
    {CC}(P,S) = \frac{\sigma(P,S)}{\sigma(P)\times\sigma(S)}\;
\end{equation}
where $\sigma$ corresponds to covariance. A CC value close to \mbox{+1/-1} demonstrates a perfect linear relationship. SIM is another popular metric that measures the similarity between the predicted and human saliency maps, as defined below:
\begin{eqnarray} 
    {\text{SIM}}(P,S) = \sum_{i}\min(P_i,S_i)\; \nonumber \\
    \text{where} \; \sum_{i}P_i = 1 \; \text{and} \; \sum_{i}S_i =1
\end{eqnarray}
 KLDiv metric evaluates the dissimilatrity between two distributions. Since KLDiv represents the difference between the saliency map and the density map, a small value indicates a good result.
However, we note that, according to the aforementioned study, NSS and CC seem to provide more fair results. In our experiments, we report the scores obtained with the implementations provided by MIT benchmark website\footnote{\url{https://github.com/cvzoya/saliency/tree/master/code_forMetrics}}.

We compare our method with ten different models: SalGAN~\cite{Pan_2017_SalGAN}, PQFT~\cite{guo2008spatio},~\cite{Fang:ICME2014}, AWS-D~\cite{awsd},~\cite{BakEE16},  OM-CNN~\cite{Jiang_2018_ECCV}, ACLNet~\cite{wang2018revisiting}, SalEMA~\cite{Linardos2019}, STRA-Net~\cite{lai2019video}, and TASED-Net~\cite{min2019tased}. Among these, SalGAN~\cite{Pan_2017_SalGAN} is the only static saliency model that gives the state-of-the-art results in the image datasets. We evaluate this method on video datasets considering each frame as a static image. PQFT~\cite{guo2008spatio},~\cite{Fang:ICME2014}, and \mbox{AWS-D}~\cite{awsd} are non-deep learning models whereas all the other models employs deep learning techniques to predict where people look in videos. We note that in~\cite{BakEE16}, the authors tested different fusion strategies with static weighting schemes and here we only report the results obtained with convolutional fusion strategy, which was shown to perform better than the others.
In our experiments, we use the implementations and the trained models provided by the authors 
and test our approach against them with the settings explained in Sec.~\ref{ssec:datasets} for fair comparison. In particular, after a careful analysis, we notice that some methods do not report results on whole test set of Hollywood-2 and/or they mistakenly consider task-specific gaze data collected for UCF-Sports while generating the groundtruth fixation density maps. Hence, some of the results are different than those reported in the papers but they give a better picture of their performances. Moreover, in our experiments, we also provide the results of single-stream versions of our model that respectively consider either spatial or temporal information.

\subsection{Qualitative and Quantitative Results}
\noindent\textbf{Performance on UCF-Sports.} Table~\ref{tab:results_ucf} reports the comparative results on UCF-Sports test set, which contains 43 sequences. As can be seen, the single-stream versions of our proposed model gives worse scores than our full model. Moreover, spatial stream generally predicts saliency much better than the temporal stream, which is a trend that we observe on the other standard benchmark datasets too. Our model trained only on UCF-Sports outperforms all the competing models in most of the metrics. It results in a performance very close to those of SalEMA and STRA-Net in terms of SIM. We believe that weighting the predictions by the spatial and temporal streams using a gating mechanism allows the model to better handle the variations throughout video sequence, thus resulting in more accurate saliency maps on this action-specific relatively small dataset.

\begin{table}[!t]
		\caption{Performance comparison on UCF-Sports dataset. The best and the second best performing models are shown in bold typeface and underlined, respectively.}
	\centering
	\resizebox{\columnwidth}{!}{
	\begin{threeparttable}	
	\begin{tabular}{l@{$\;\;$}l||ccccc}
		\hline
		 \multicolumn{2}{c||}{\backslashbox{\small{Method}}{\small{Metric}}}&\small{AUC-J$\uparrow$}& \small{CC$\uparrow$} & \small{NSS$\uparrow$} & \small{SIM$\uparrow$}& \small{KLDiv$\downarrow$}\\ 
		\hline 
		
		{\small{Static}} & \small{SalGAN} &\small{0.869} &\small{0.389} &\small{2.074}&\small{0.258}&\small{2.169}\\
		\hline
		& \small{PQFT*}&\small{0.776}& \small{0.211}& \small{1.189}& \small{0.157}& \small{2.458}\\
		& \small{Fang et al.*}& \small{0.879}& \small{0.387} & \small{2.319} & \small{0.247}& \small{2.012}\\ 
		& \small{AWS-D*} & \small{0.845} & \small{0.313} & \small{1.870} & \small{0.195}& \small{2.202}\\
		& \small{Bak et al.} & \small{0.864} & \small{0.387} & \small{2.231} & \small{0.130}& \small{2.575}\\
	{\small{Dynamic}}	& \small{OM-CNN} &\small{0.880} & \small{0.398} & \small{2.443} &\small{0.294}& \small{1.902}\\
		& \small{ACLNet} &\small{0.876}  &\small{0.367}  & \small{2.045}  &  \small{0.292} & \small{2.135} \\
		& \small{SalEMA} &\small{0.895}&\small{0.470}&\small{2.979}&\textbf{\small{0.384}}& \small{1.728}\\
		& {\small{STRA-Net}} & {\small{0.902}} & {\small{0.479}} & {\small{2.916}}& {\textbf{\small{0.384}}}& \small{2.483}\\
		& {\small{TASED-Net}}& {\small{0.887}} & {\small{0.453}} & {\small{2.680}}& {\small{0.369}}& \small{1.876} \\ \hline
	{{\small{Ours}}}	& {\small{Spatial}} &{\small{0.870}} & {\small{0.461}} & {\underline{\small{3.029}}} &{\small{0.377}}&{\small{2.504}}\\
	{\small{(Single)}}	& {\small{Temporal}}  &{\small{0.851}} & {\small{0.418}} & {\small{2.535}} & {\small{0.345}}&{\small{2.721}}\\\hline
	{\small{Ours}} & 	{\small{Setting 1}} &  {\small{\textbf{0.914}}} & {\small{\textbf{0.526}}} & {\small{\textbf{3.333}}} & {\small{\underline{0.382}}}& {\small{\textbf{1.516}}}\\
		{\small{(Gated)}}& 	\small{Setting 2} &  {\small{\underline{0.911}}} & {\small{\underline{0.499}}} & {\small{{2.980}}} & {\small{{0.353}}}&\small{\underline{1.568}}\\
		\hline
	\end{tabular}
	\begin{tablenotes}  
	\small
      \item  * Non-deep learning model
    \end{tablenotes}
  \end{threeparttable}}
	\label{tab:results_ucf}
\end{table}

\noindent
\textbf{Performance on Hollywood-2.} In our experiments on Hollywood-2 dataset, we use all the frames from the test set that contains 884 video sequences. In that regard, it is the largest test set that we considered in our experimental evaluation. In Table~\ref{tab:results_holly}, we provide comparison against the competing saliency models. Our results show that our model gives better saliency predictions than all the other methods in terms of the AUC-J and KLDiv metrics. The performance of the model trained considering our second training setting that includes a larger and more diverse training set provides much better results than the one trained with the first setting. In terms of the remaining evaluation metrics, our results are highly competitive as compared to the recent state-of-the-art models, namely STRA-Net and TASED-Net, as well.

\begin{table}[!t]
		\caption{Performance comparison on Hollywood-2 dataset. The best and the second best performing models are shown in bold typeface and underlined, respectively.}
	\centering
	\resizebox{\columnwidth}{!}{%
	\begin{threeparttable}
	\begin{tabular}{l@{$\;\;$}l||ccccc}
		\hline
		\multicolumn{2}{c||}{\backslashbox{\small{Method}}{\small{Metric}}}&\small{AUC-J$\uparrow$} & \small{CC$\uparrow$} & \small{NSS$\uparrow$} & \small{SIM$\uparrow$}& \small{KLDiv$\downarrow$}\\ 
		\hline 
		{\small{Static}} & \small{SalGAN} & \small{0.892} & \small{0.428} & \small{2.383} &\small{0.298}&\small{{1.760}}\\
		\hline
		& \small{PQFT*}&\small{0.689}& \small{0.150}& \small{0.610}& \small{0.139}& \small{{2.387}}\\
		& \small{Fang et al.*}&\small{0.862}& \small{0.312}& \small{1.614}&\small{0.221}& \small{{1.781}}\\
		& \small{AWS-D*}&\small{0.747}&\small{0.227}&\small{0.994}&\small{0.193}&\small{{2.256}}\\ 
		& \small{Bak et al.} &\small{0.840}&\small{0.310}&\small{1.439}&\small{0.158}&\small{{2.339}}\\
		{\small{Dynamic}}& \small{OM-CNN} &\small{0.893}&\small{0.430}&\small{2.625}&\small{0.330}& \small{{1.896}}\\
		& \small{ACLNet}&\small{0.899} &\small{0.459}  & \small{2.463}  &  \small{0.342} &\small{{1.701}}\\
		& \small{SalEMA}&\small{0.873}&\small{0.383}&\small{2.226}&\small{0.330}&\small{{3.157}}\\
		& \small{STRA-Net}&\small{0.913}&\small{0.558}&\small{\underline{3.226}}&\small{\underline{0.459}}&\small{{2.251}}\\
		& \small{TASED-Net}&\small{0.916}&\textbf{\small{ 0.570}}&\textbf{\small{3.324}}&\textbf{\small{0.471}}&\small{{2.740}}\\ \hline
		{\small{Ours}} & \small{Spatial} &\small{0.904} & \small{0.501} & \small{3.051} &\small{0.378}&\small{1.473}\\
		{\small{(Single)}} & \small{Temporal}  &\small{0.898} & \small{0.489} & \small{2.581} &\small{0.362}&\small{1.468}\\ \hline
		{\small{Ours}}& \small{Setting 1} &   \small{\underline{0.914}}& \small{{0.549}}& \small{{3.114}} & \small{{0.413}}& \small{{\underline{1.277}}} \\
		{\small{(Gated)}} & 	\small{Setting 2} &  \small{\textbf{0.919}} & \small{\underline{0.563}} &\small{{3.201}} & \small{{0.424}}& \small{{\textbf{1.242}}}\\
		\hline
	\end{tabular}
	\begin{tablenotes}
      \small
      \item * Non-deep learning model
    \end{tablenotes}
  \end{threeparttable}}
	\label{tab:results_holly}
\end{table}

\clearpage
\noindent\textbf{Performance on DHF1K.} We test the performance of our model on the recently proposed DHF1K video saliency dataset, which includes 300 test videos. As mentioned before, the annotations for the test split are not publicly available and all the evaluations are carried out externally by the authors of the dataset. As Table~\ref{tab:results_dhf1k} shows, our proposed model achieves performance on par with the state-of-the-art models. In terms of AUC-J, along with the recent STRA-Net and TASED-Net models, it outperforms all the other saliency models. In terms of CC, our model gives roughly the second best result.

\noindent
\textbf{Performance on DIEM.} We also evaluate our model on DIEM test set consisting of 20 videos. Table~\ref{tab:results_diem} summarizes these quantitative results. As can be seen, our model achieves the highest scores in NSS and KLDiv metrics and very competitive in others. The second setting demonstrates the generalization capability of our proposed approach as compared to the recent models like SalEMA, STRA-Net and TASED-Net.

\begin{table}[!t]
		\caption{Performance comparison on DHF1K dataset. The best and the second best performing models are shown in bold typeface and underlined, respectively.}
	\resizebox{0.85\columnwidth}{!}{%
	\begin{threeparttable}
	\centering
	\begin{tabular}{l@{$\;\;$}l||cccc}
		\hline
		\multicolumn{2}{c||}{\backslashbox{\small{Method}}{\small{Metric}}}&\small{AUC-J$\uparrow$} & \small{CC$\uparrow$} & \small{NSS$\uparrow$} & \small{SIM$\uparrow$}\\ 
		\hline 
		{\small{Static}} & \small{SalGAN}&\small{0.866}&\small{0.370}&\small{2.043}&\small{0.262}\\
		\hline
		& \small{PQFT*}&\small{0.699}& \small{0.137}& \small{0.749}& \small{0.139}\\
		& \small{Fang et al.*}&\small{0.819}& \small{0.273}& \small{1.539}&\small{0.198}\\
		& \small{AWS-D*}&\small{0.703}&\small{0.174}&\small{0.940}&\small{0.157}\\ 
		& \small{Bak et al.}&\small{0.834}&\small{0.325}&\small{1.632}&\small{0.197}\\
		{\small{Dynamic}} & \small{OM-CNN} &\small{0.856}&\small{0.344}&\small{1.911}&\small{0.256}\\
		& \small{ACLNet}&\small{0.890} &\small{0.434}  & \small{2.354} &  \small{0.315} \\
		& \small{SalEMA} &\small{0.890}&\small{0.449}&\small{\underline{2.574}}&\small{\textbf{0.466}}\\
		& \small{STRA-Net}&\textbf{\small{0.895}}&\small{\underline{0.458}}&\small{2.558}&\small{0.355}\\
		& \small{TASED-Net}&\textbf{\small{0.895}}&\textbf{\small{0.470}}&\textbf{\small{2.667}}&\small{\underline{0.361}}\\ \hline
		{\small{Ours}} & {{\small{Setting 1}}} & {\underline{\small{0.891}}} & {\small{0.448}} & {\small{2.505}} & {\small{0.326}}\\
		{\small{(Gated)}} & {\small{Setting 2}}& {\small{\textbf{0.895}}}&{\small{0.457}} & {\small{2.528}}&{\small{0.321}} \\
		\hline
	\end{tabular}
	\begin{tablenotes}
      \small
      \item * Non-deep learning model
    \end{tablenotes}
  \end{threeparttable}}
\label{tab:results_dhf1k}
\end{table}

\begin{table}[!t]
		\caption{Performance comparison on DIEM dataset. The best and the second best performing models are shown in bold typeface and underlined, respectively.}
	\centering
	\resizebox{\columnwidth}{!}{
	\begin{threeparttable}
	\begin{tabular}{l@{$\;\;$}l||ccccc}
		\hline
		\multicolumn{2}{c||}{\backslashbox{\small{Method}}{\small{Metric}}}&\small{AUC-J$\uparrow$} & \small{CC$\uparrow$} & \small{NSS$\uparrow$} & \small{SIM$\uparrow$}&\small{KLDiv$\downarrow$}\\ 
		\hline 
		
		{\small{Static}} & \small{SalGAN}&\small{0.860}&\small{0.492}&\small{2.068}&\small{0.392}&\small{1.431}\\
		\hline
		& \small{PQFT*}&\small{0.680}& \small{0.190}& \small{0.656}& \small{0.220}&\small{2.140}\\
		& \small{Fang et al.*}&\small{0.825}& \small{0.360}& \small{1.407}&\small{0.313}&\small{1.688}\\
		& \small{AWS-D*}&\small{0.768}&\small{0.313}&\small{1.228}&\small{0.272}&\small{1.825}\\ 
        & \small{Bak et al.}&\small{0.810}&\small{0.313}&\small{1.212}&\small{0.206}&\small{2.050}\\
		{\small{Dynamic}} & \small{OM-CNN} &\small{0.847}&\small{0.464}&\small{2.037}&\small{0.381}&\small{1.599}\\

		& \small{ACLNet}&\small{\textbf{0.878}} &\small{\textbf{0.554}} & \small{{2.283}}  &  \small{0.444} &\small{\underline{1.331}}\\
		& \small{SalEMA}&\small{0.863}&\small{0.513}&\small{2.249}&\small{0.452}&\small{2.393}\\
		& \small{STRA-Net}&\small{0.864}&\small{0.527}&\small{2.277}&\small{0.456}&\small{2.461}\\
		& \small{TASED-Net} & \small{0.872} & \small{0.535} & \small{2.259} & \small{\textbf{0.470}}&\small{2.635}\\ \hline
		
        {\small{Ours}} & \small{Spatial} &\small{0.868} & \small{0.512} & \small{2.202} &\small{0.439}&\small{1.387}\\
		{\small{(Single)}} & \small{Temporal}  &\small{0.846} & \small{0.446} & \small{1.785} &\small{0.391}&\small{1.513}\\ \hline
		
        {\small{Ours}} & {{\small{Setting 1}}} &{\small{0.870}} & \underline{\small{0.543}} & \textbf{\small{2.313}} & \underline{\small{0.454}}& {\small{1.401}}\\
		{\small{(Gated)}} & {\small{Setting 2}}&  \underline{\small{0.874}} & \small{0.525} &\small{2.228} & \small{0.421}&\small{\textbf{1.176}} \\
		\hline
	\end{tabular}
	\begin{tablenotes}
      \small
      \item * Non-deep learning model
    \end{tablenotes}
  \end{threeparttable}}
	\label{tab:results_diem}
\end{table}

\begin{figure*}[!t]
\centering
\resizebox{\linewidth}{!}{
\begin{tabular}{cc@{}c@{}c@{}c@{}|c@{}c@{}c@{}c}
\rotatebox{90}{\huge{$\quad\;\;\;$GT}} & \includegraphics[width=0.25\linewidth]{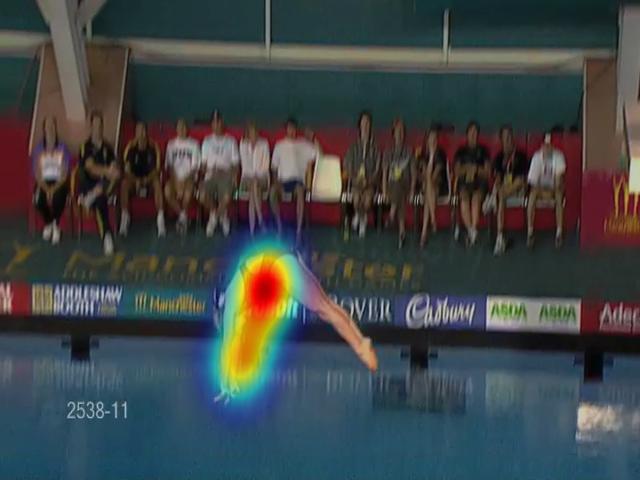} &\includegraphics[width=0.25\linewidth]{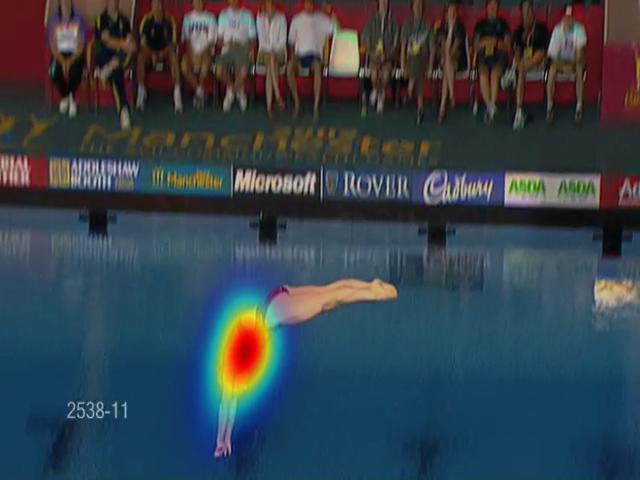} &\includegraphics[width=0.25\linewidth]{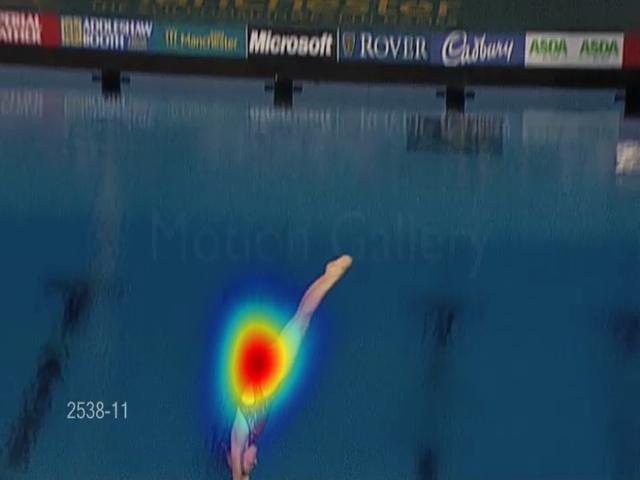}
&\includegraphics[width=0.25\linewidth]{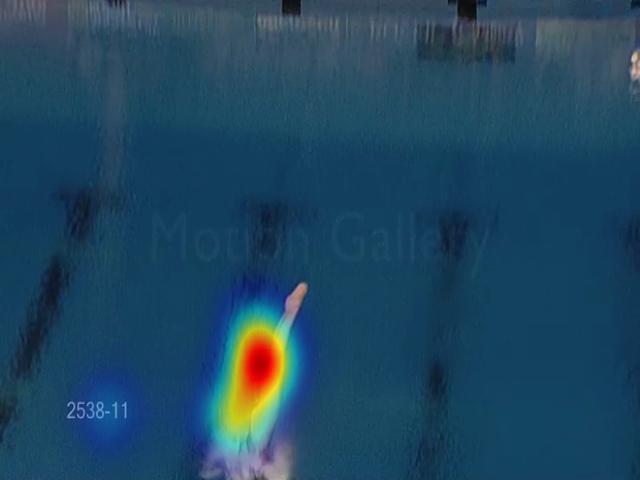}
&\includegraphics[width=0.25\linewidth]{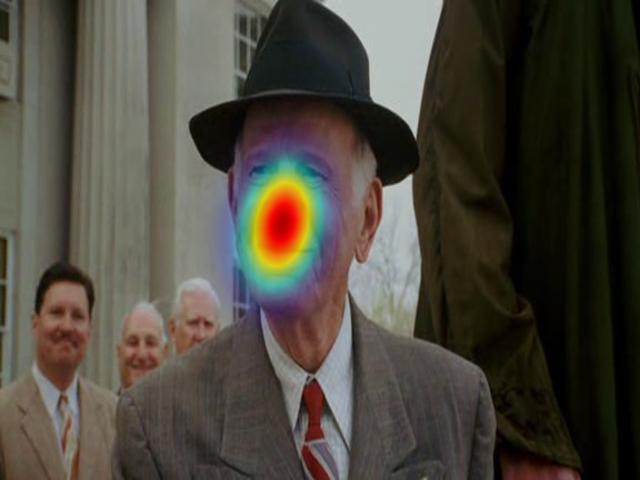} &\includegraphics[width=0.25\linewidth]{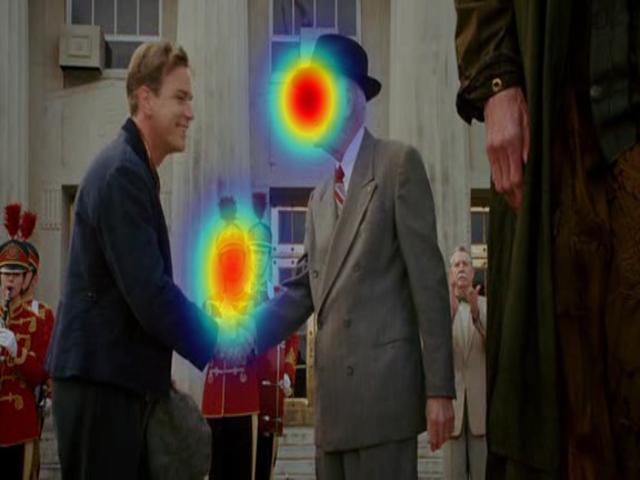} &\includegraphics[width=0.25\linewidth]{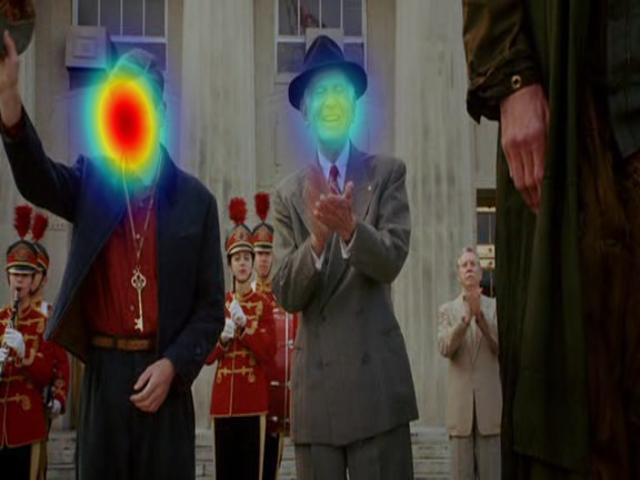} &\includegraphics[width=0.25\linewidth]{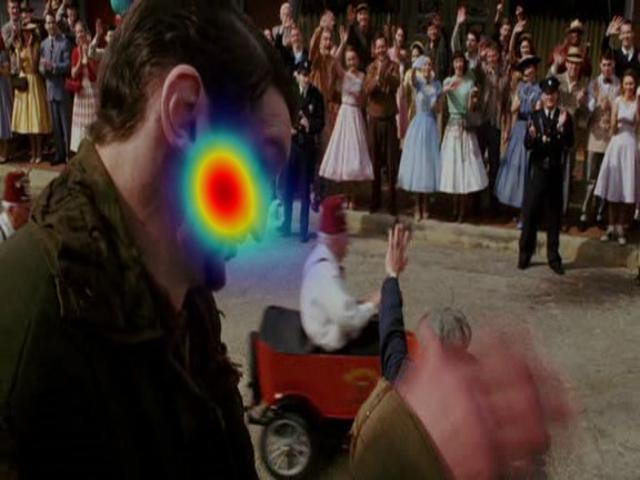}\\
\rotatebox{90}{\huge{$\quad\;$Ours}} & \includegraphics[width=0.25\linewidth]{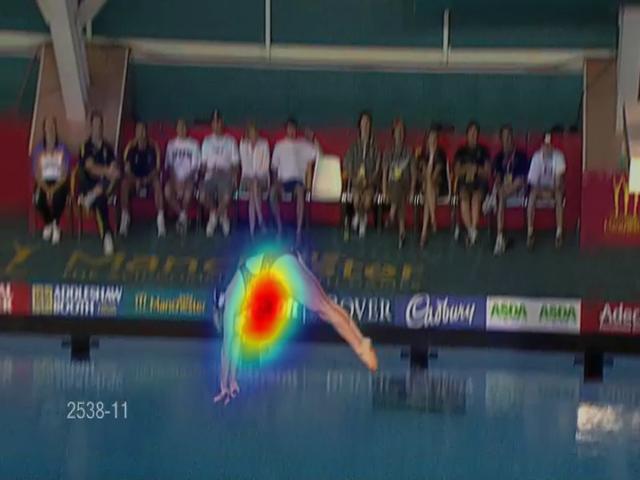} &\includegraphics[width=0.25\linewidth]{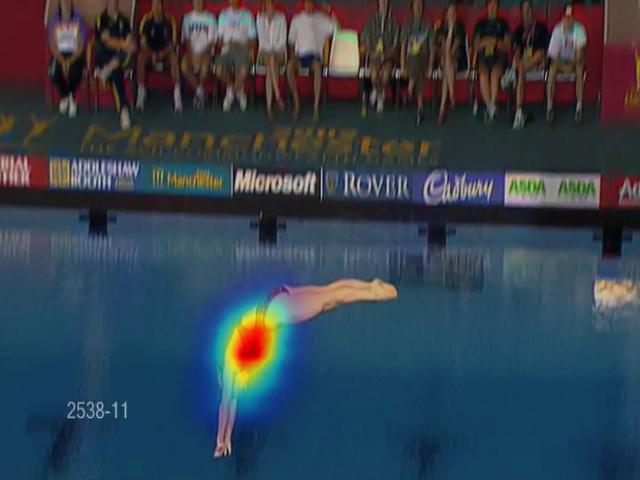} &\includegraphics[width=0.25\linewidth]{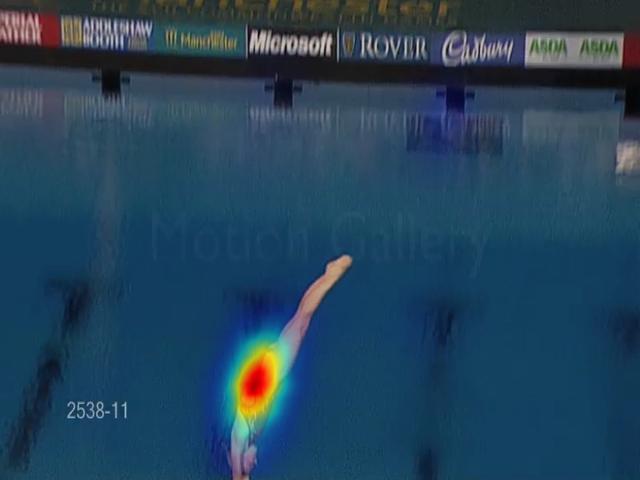}
&\includegraphics[width=0.25\linewidth]{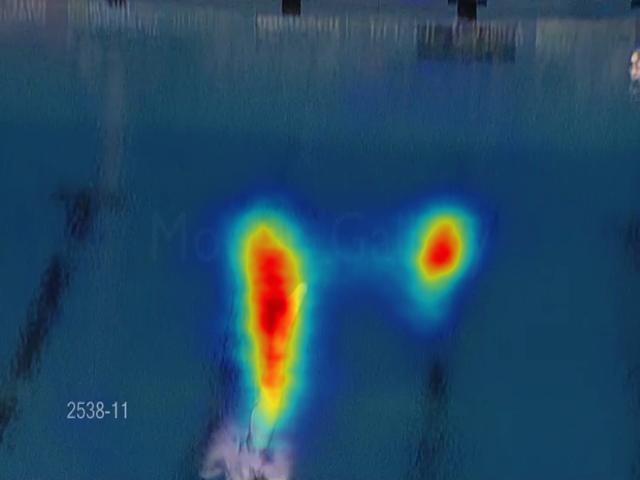}
&\includegraphics[width=0.25\linewidth]{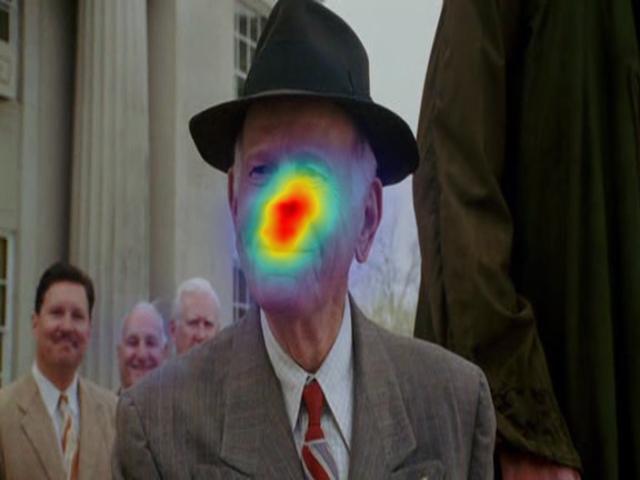} &\includegraphics[width=0.25\linewidth]{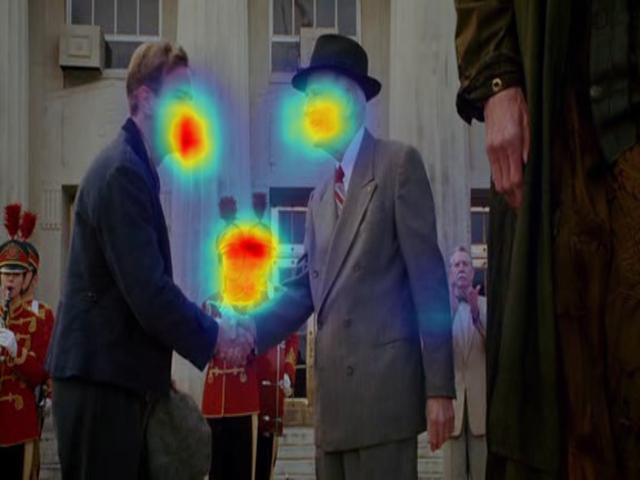} &\includegraphics[width=0.25\linewidth]{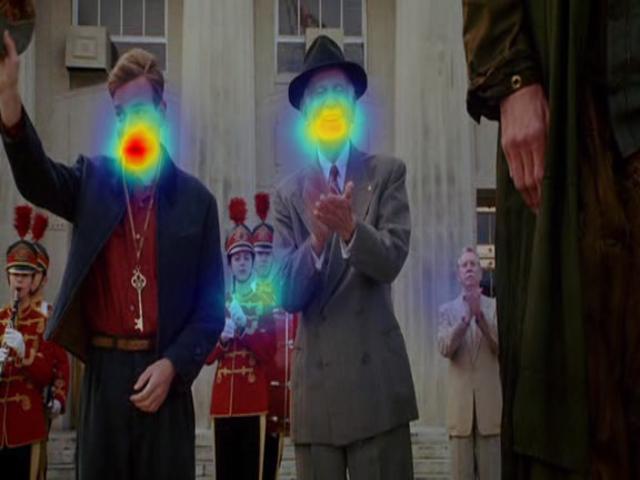} &\includegraphics[width=0.25\linewidth]{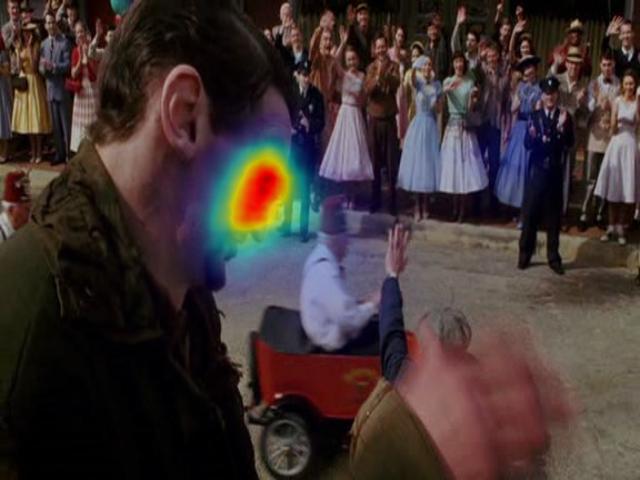}\\
\rotatebox{90}{\huge{$\quad\!\!$SalEMA}}&  \includegraphics[width=0.25\linewidth]{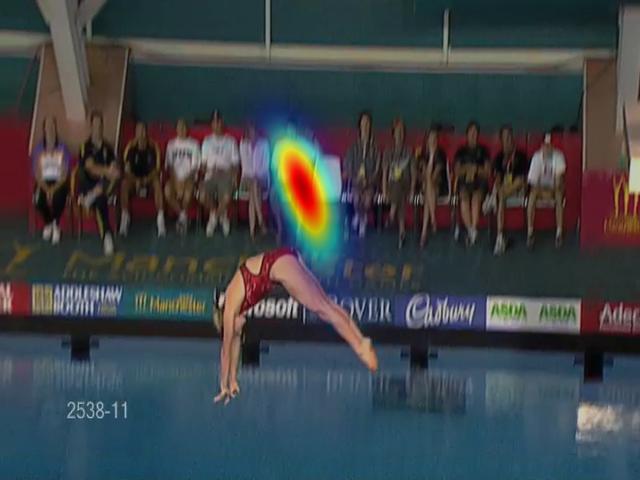} &\includegraphics[width=0.25\linewidth]{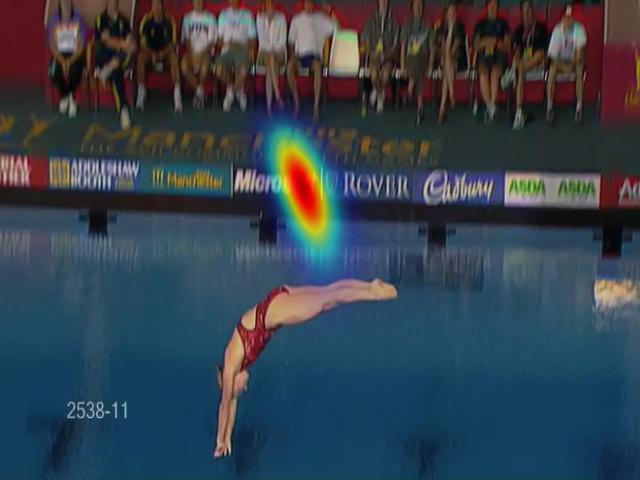} &\includegraphics[width=0.25\linewidth]{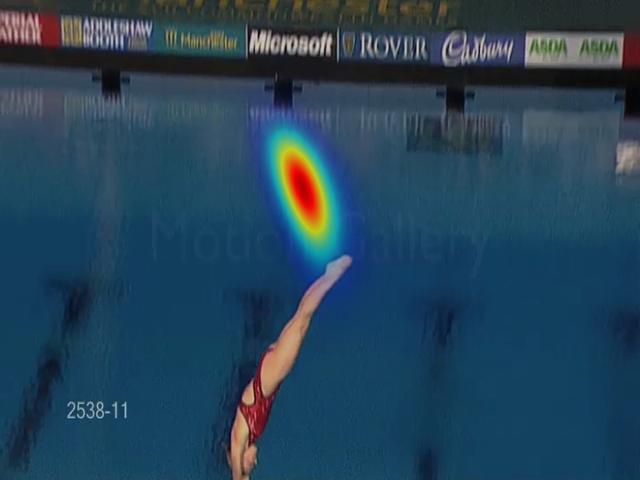}
&\includegraphics[width=0.25\linewidth]{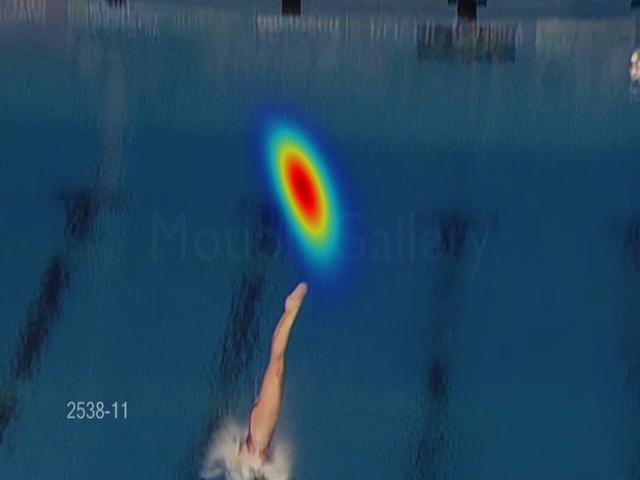}
&\includegraphics[width=0.25\linewidth]{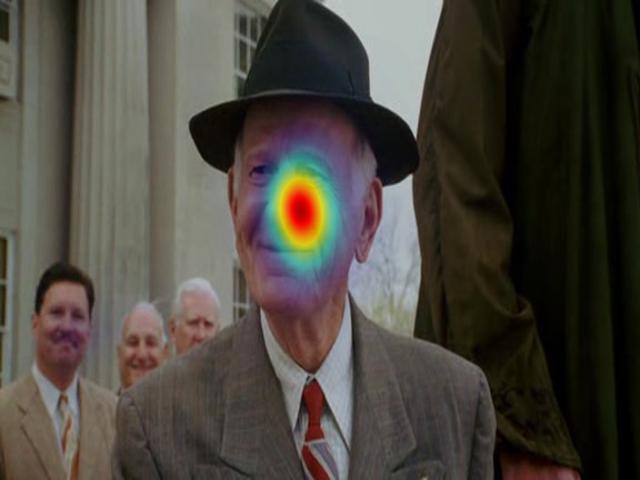} &\includegraphics[width=0.25\linewidth]{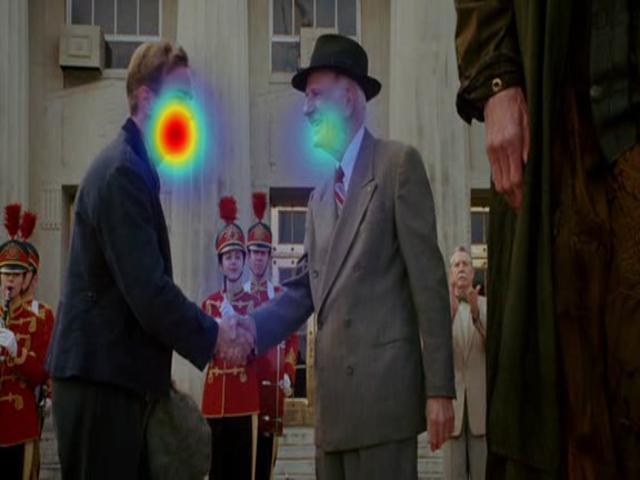} &\includegraphics[width=0.25\linewidth]{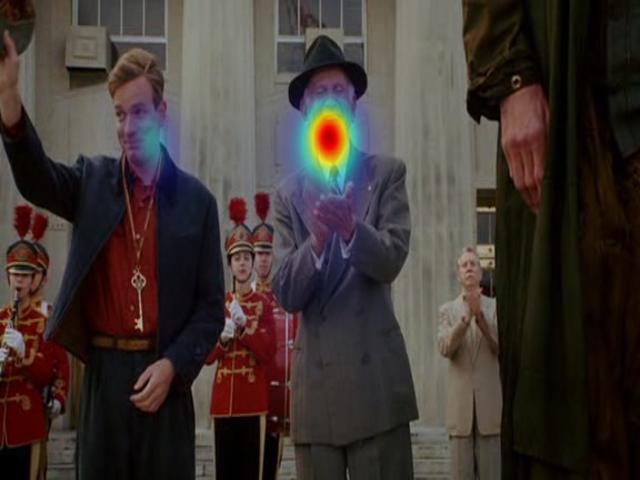} &\includegraphics[width=0.25\linewidth]{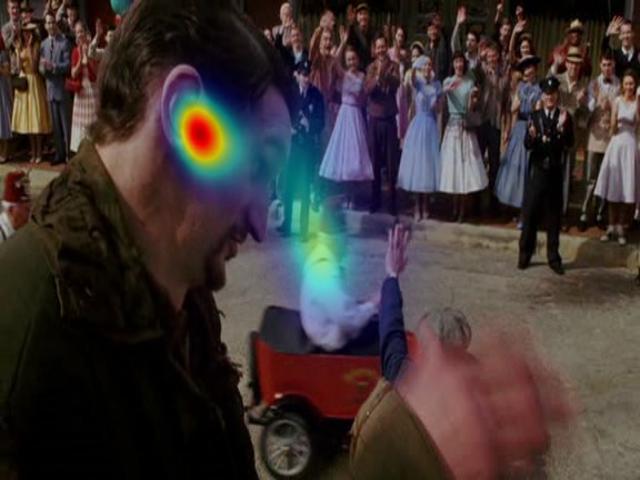}\\
\rotatebox{90}{\huge{$\quad\!\!$ACLNet}}&\includegraphics[width=0.25\linewidth]{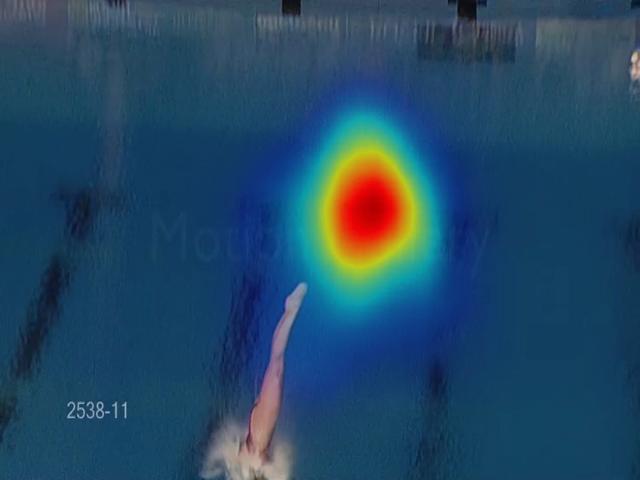} & \includegraphics[width=0.25\linewidth]{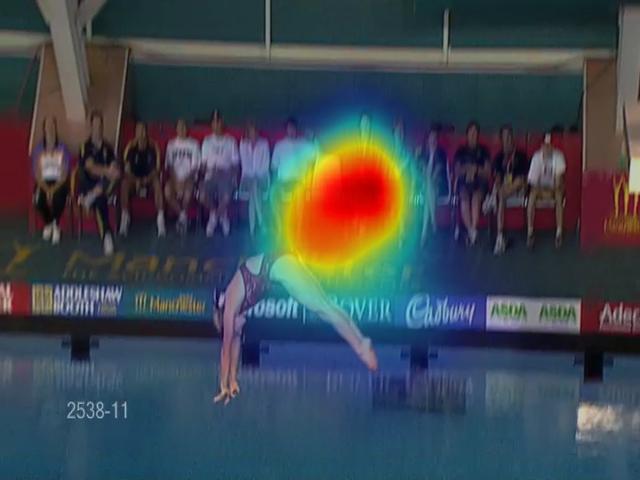} &\includegraphics[width=0.25\linewidth]{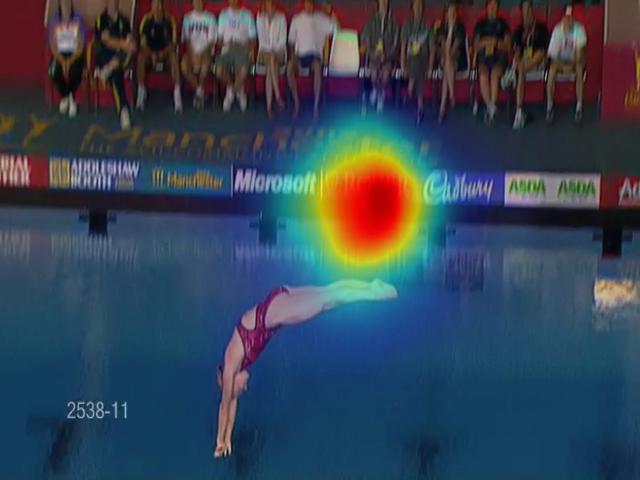} &\includegraphics[width=0.25\linewidth]{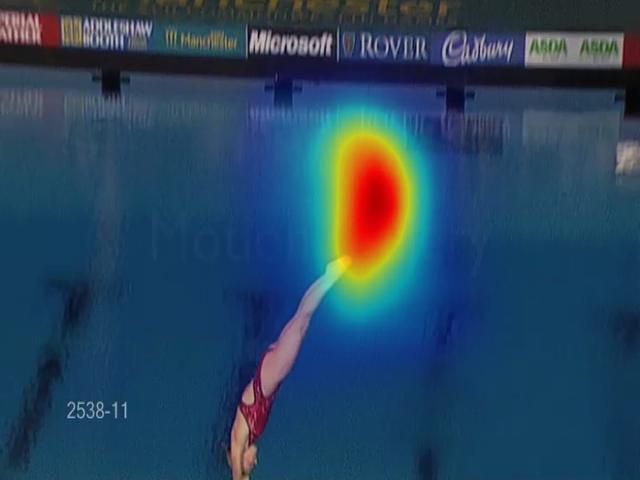} &\includegraphics[width=0.25\linewidth]{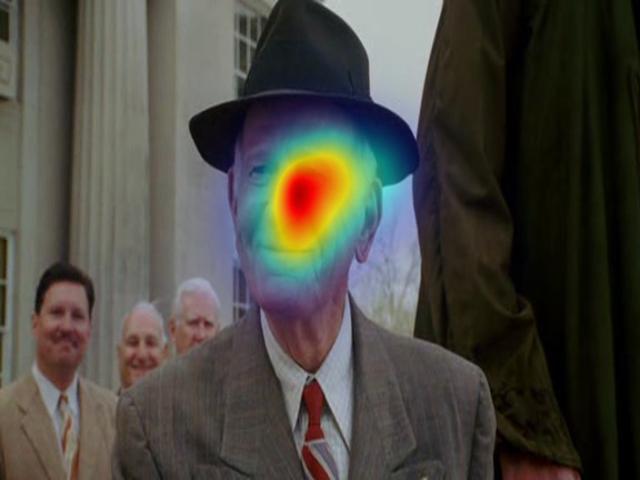} &\includegraphics[width=0.25\linewidth]{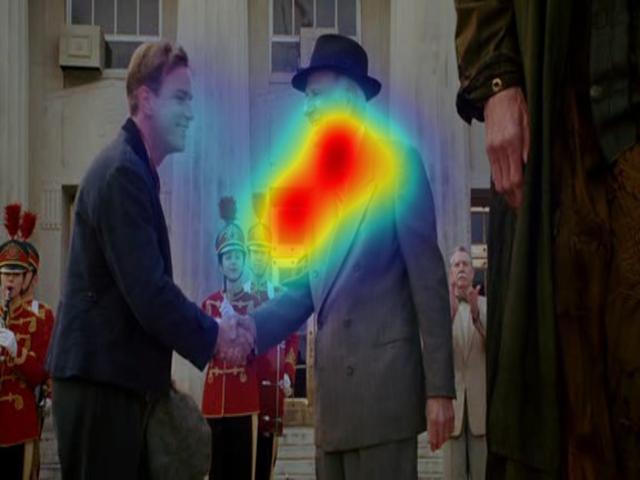} &\includegraphics[width=0.25\linewidth]{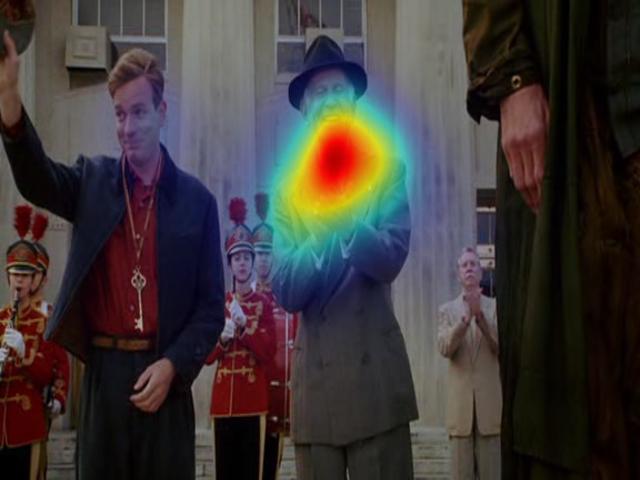} &\includegraphics[width=0.25\linewidth]{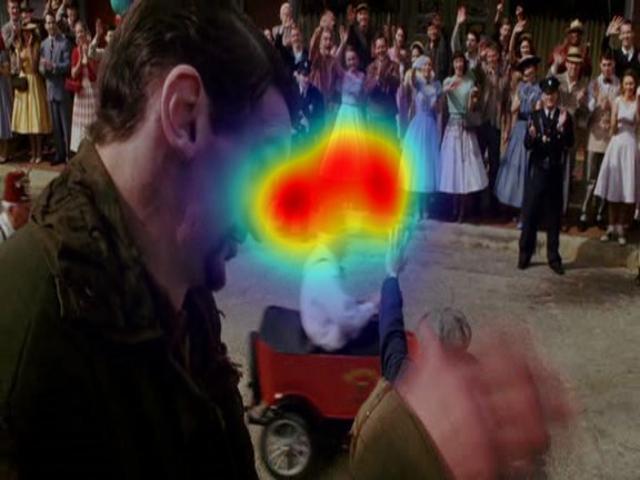}\\
\rotatebox{90}{\LARGE{$\;\;\!\!$TASED-Net}}&\includegraphics[width=0.25\linewidth]{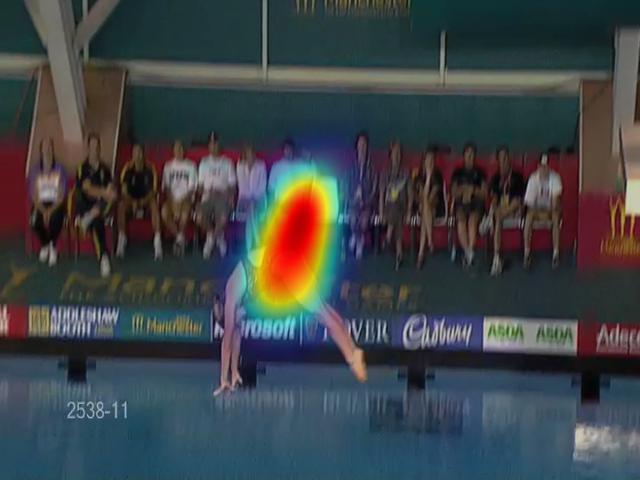} & \includegraphics[width=0.25\linewidth]{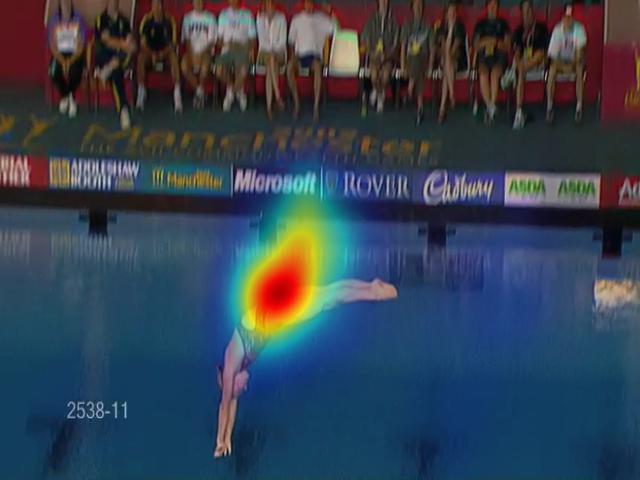} &\includegraphics[width=0.25\linewidth]{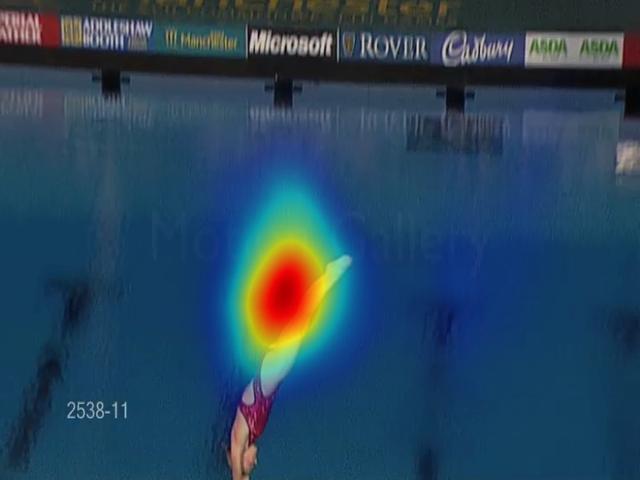} &\includegraphics[width=0.25\linewidth]{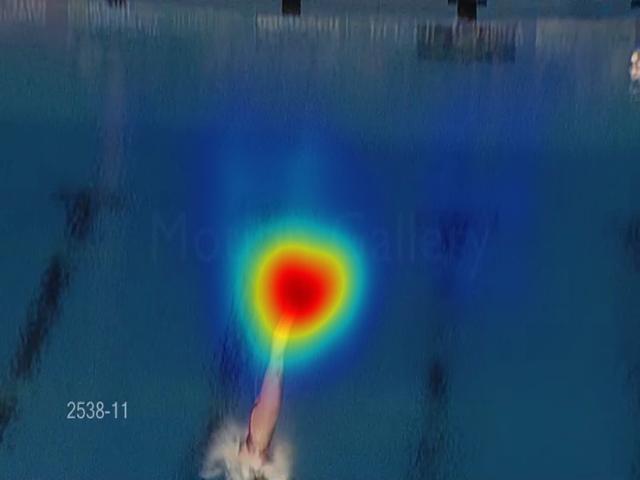} &\includegraphics[width=0.25\linewidth]{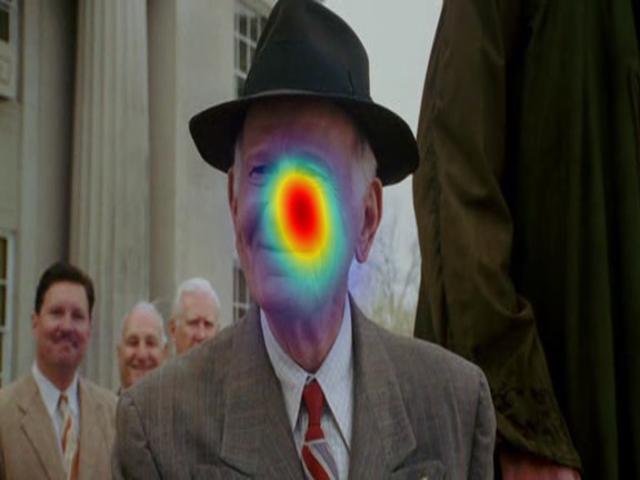} &\includegraphics[width=0.25\linewidth]{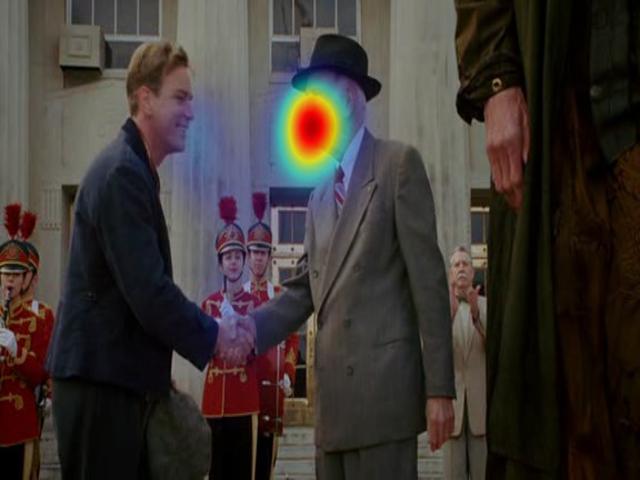} &\includegraphics[width=0.25\linewidth]{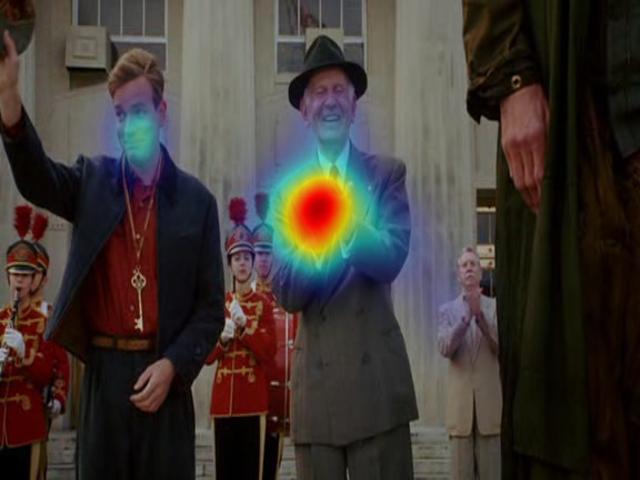} &\includegraphics[width=0.25\linewidth]{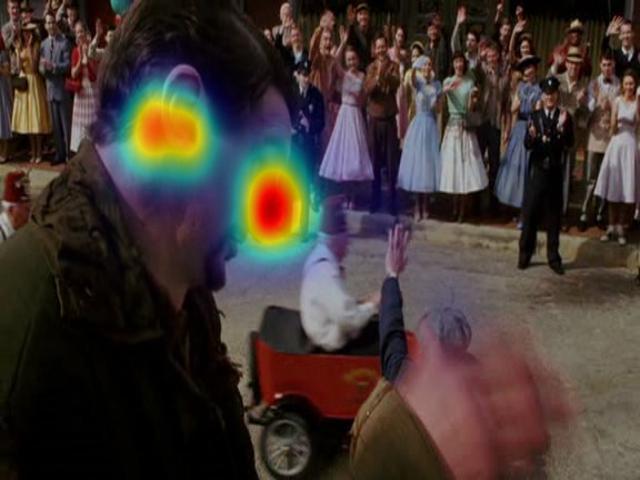}\\
\rotatebox{90}{\LARGE{$\;\;\;\,\!\!$STRA-Net}}&\includegraphics[width=0.25\linewidth]{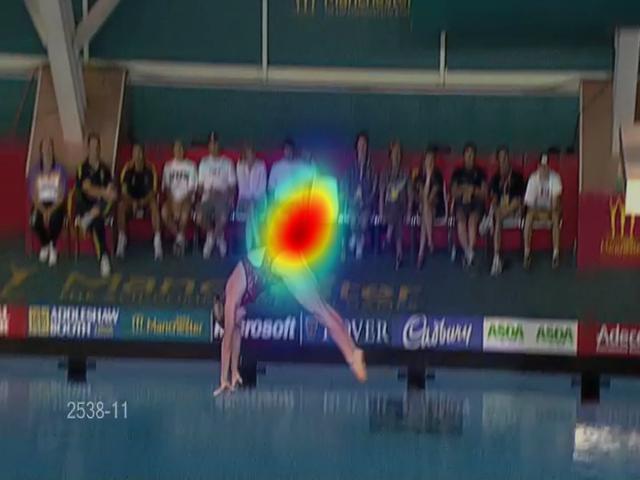} & \includegraphics[width=0.25\linewidth]{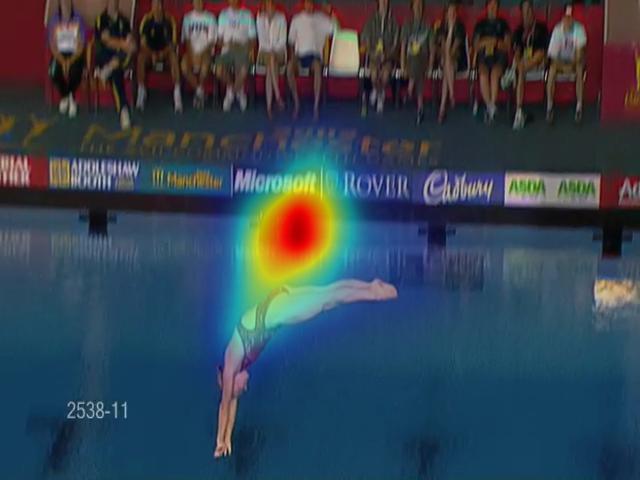} &\includegraphics[width=0.25\linewidth]{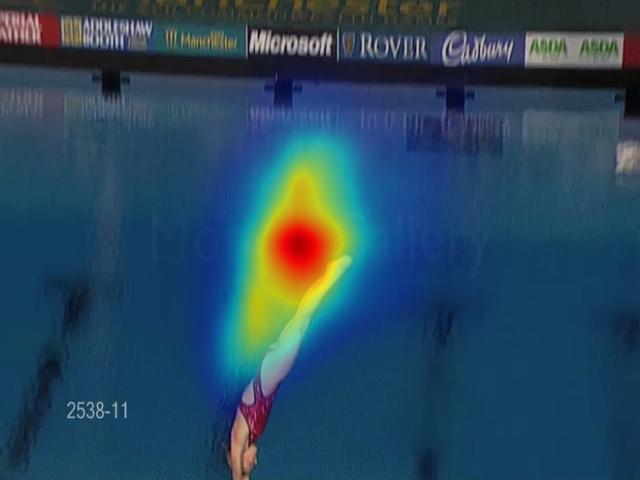} &\includegraphics[width=0.25\linewidth]{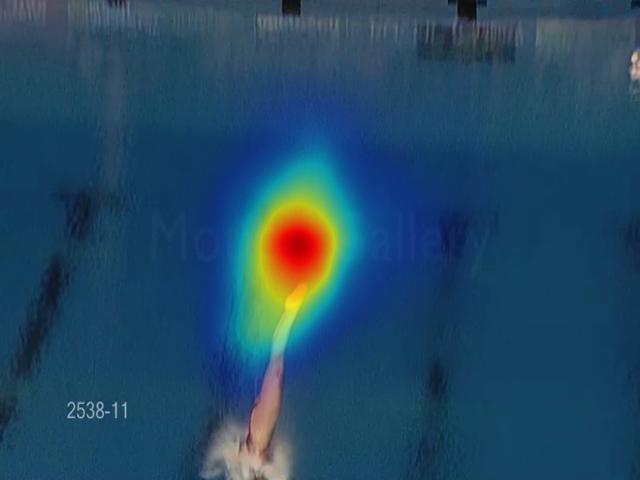} &\includegraphics[width=0.25\linewidth]{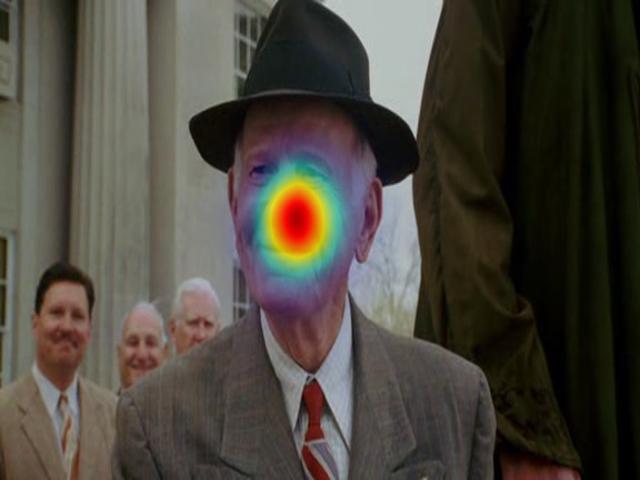} &\includegraphics[width=0.25\linewidth]{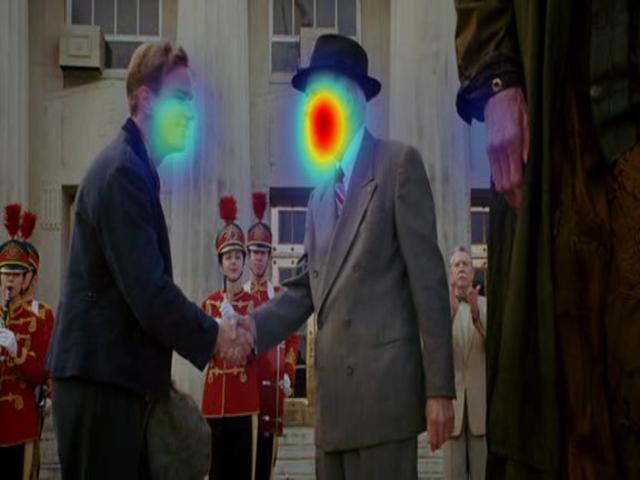} &\includegraphics[width=0.25\linewidth]{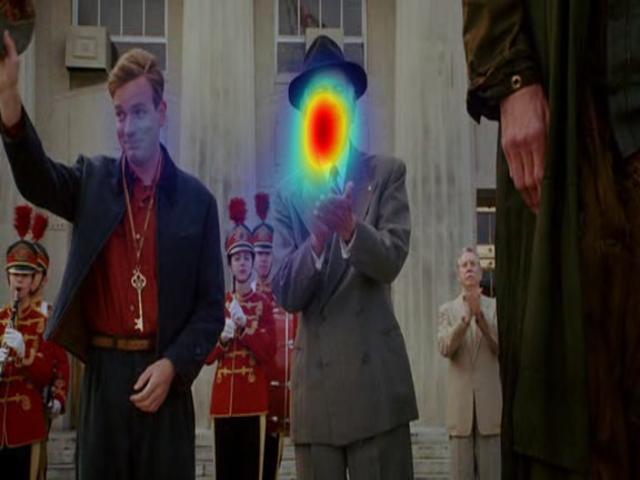} &\includegraphics[width=0.25\linewidth]{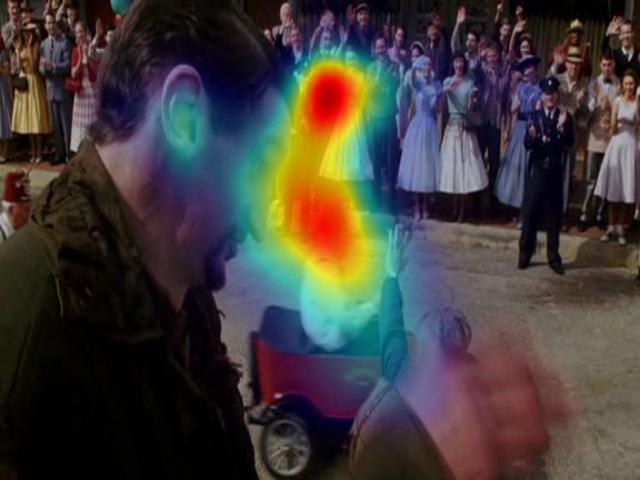}\\
\hfill
&\multicolumn{4} {c}{\huge{(a) UCF-Sports}}&\multicolumn{4} {c}{\huge{(b) Hollywood-2}}\vspace{0.3cm}\\
\rotatebox{90}{\huge{$\quad\;\;\;$GT}}
&\includegraphics[width=0.25\linewidth,height=0.1875\linewidth]{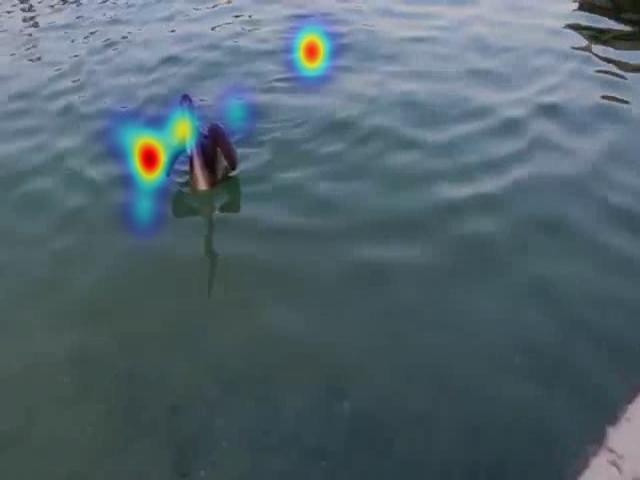} 
&\includegraphics[width=0.25\linewidth,height=0.1875\linewidth]{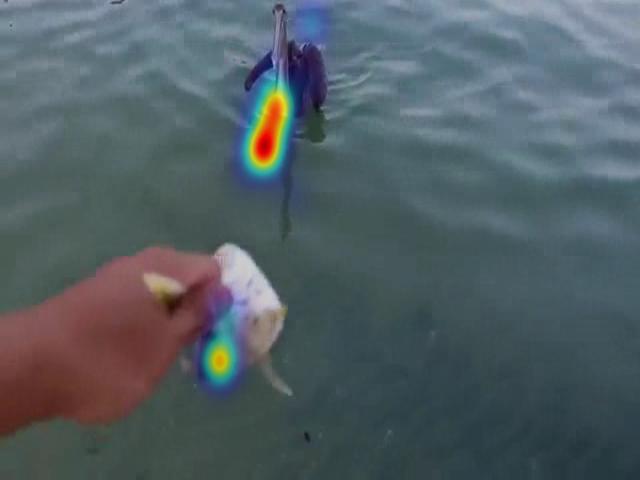} &\includegraphics[width=0.25\linewidth,height=0.1875\linewidth]{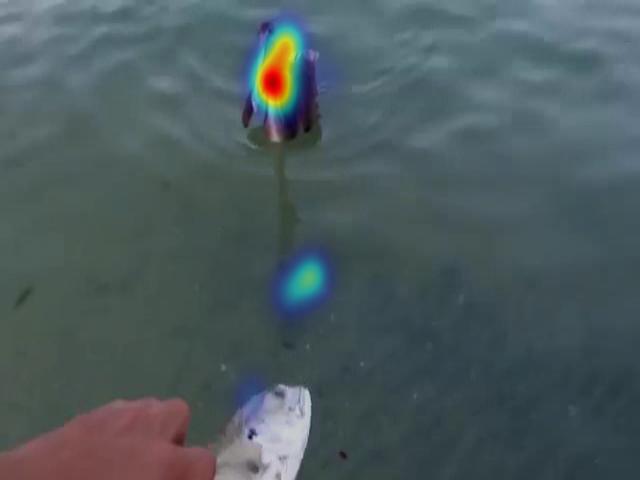} &\includegraphics[width=0.25\linewidth,height=0.1875\linewidth]{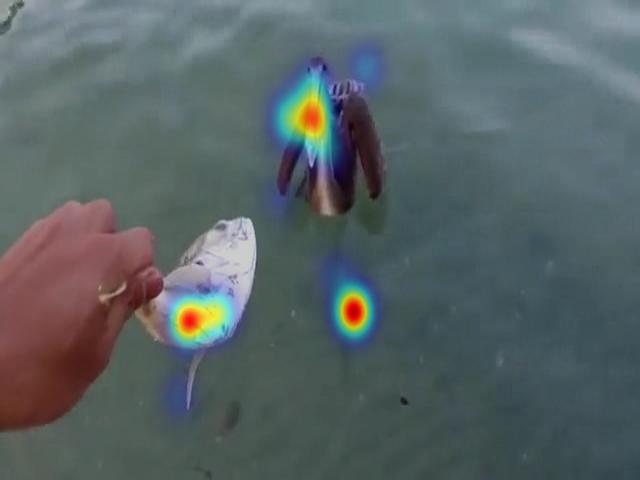} &\includegraphics[width=0.25\linewidth]{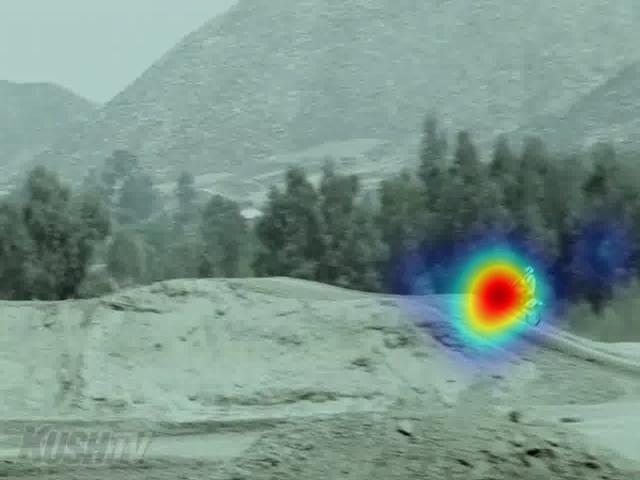} &\includegraphics[width=0.25\linewidth]{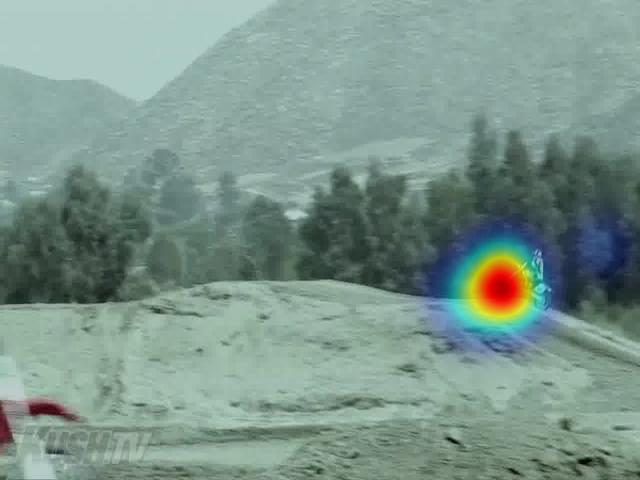} &\includegraphics[width=0.25\linewidth]{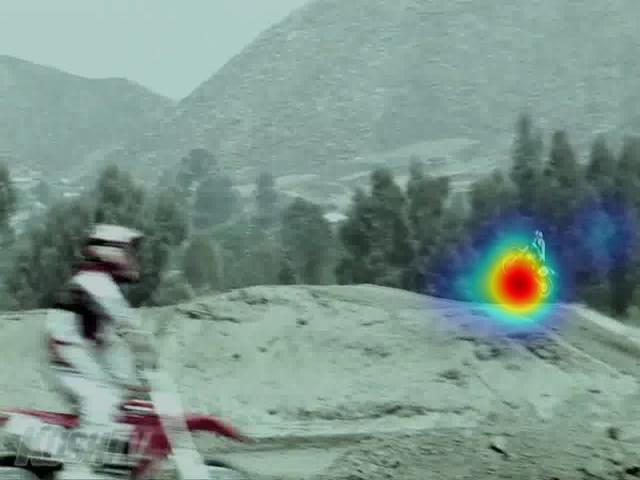} &\includegraphics[width=0.25\linewidth]{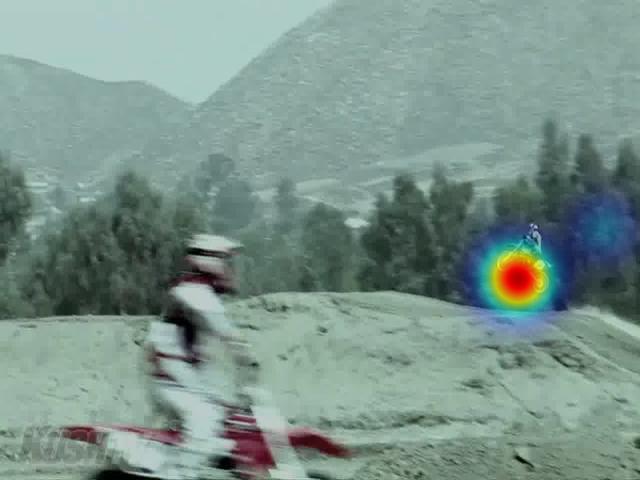}\\
\rotatebox{90}{\huge{$\quad\;$Ours}}
&\includegraphics[width=0.25\linewidth,height=0.1875\linewidth]{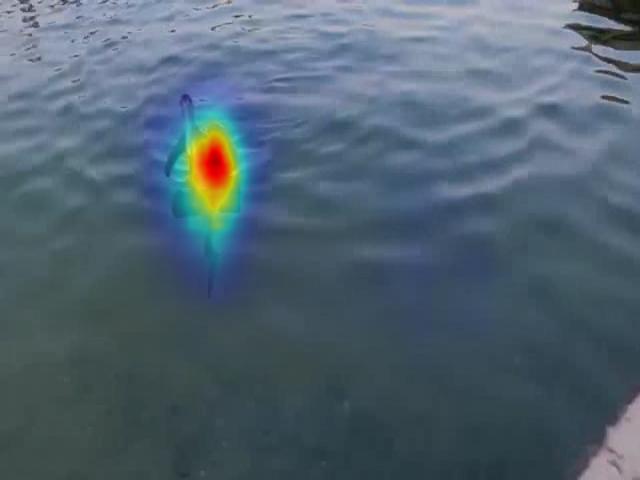} 
&\includegraphics[width=0.25\linewidth,height=0.1875\linewidth]{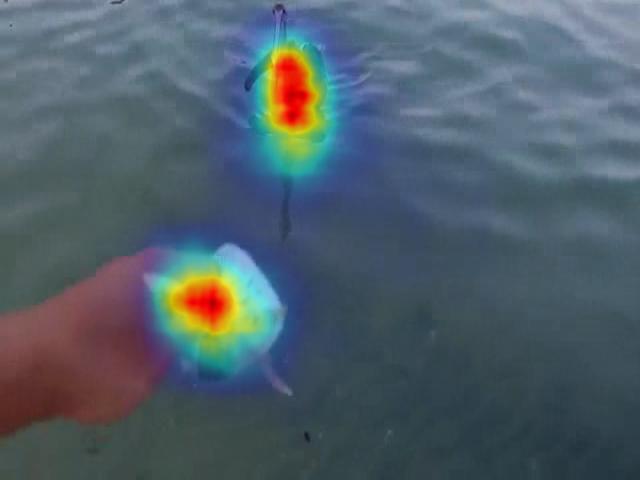} &\includegraphics[width=0.25\linewidth,height=0.1875\linewidth]{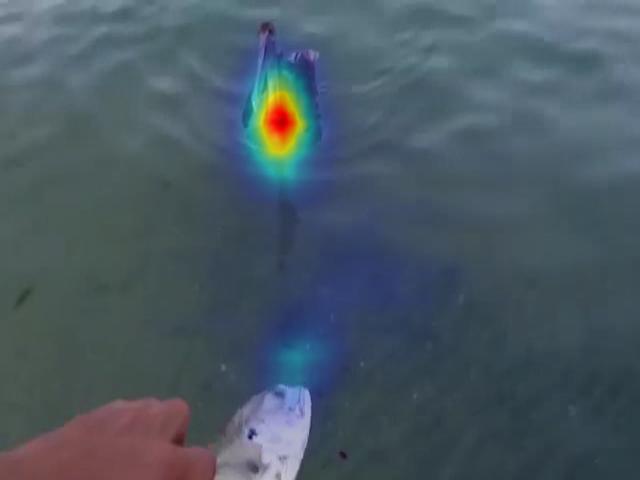} &\includegraphics[width=0.25\linewidth,height=0.1875\linewidth]{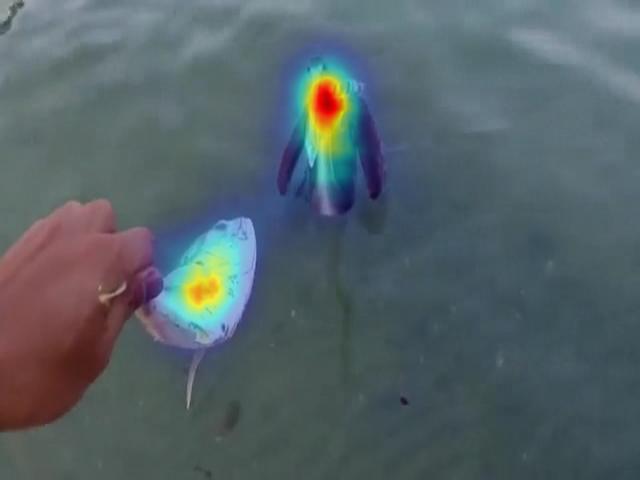} &\includegraphics[width=0.25\linewidth]{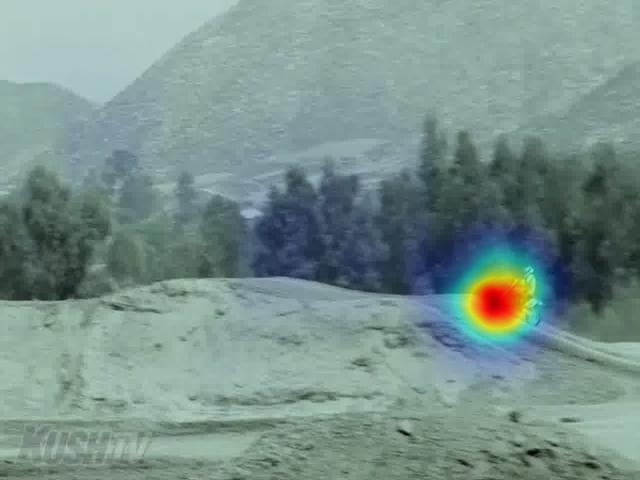} &\includegraphics[width=0.25\linewidth]{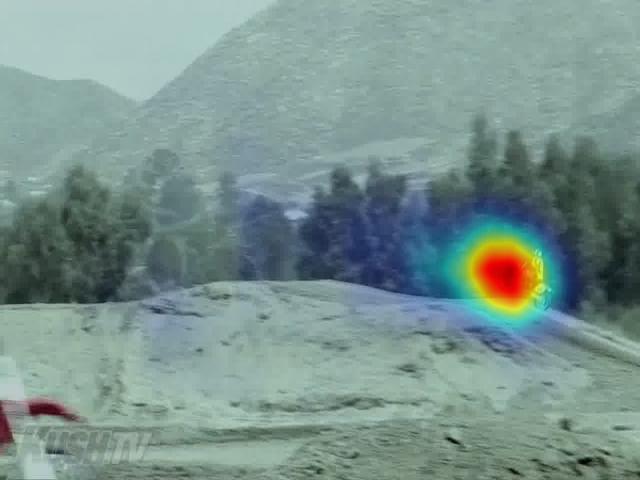} &\includegraphics[width=0.25\linewidth]{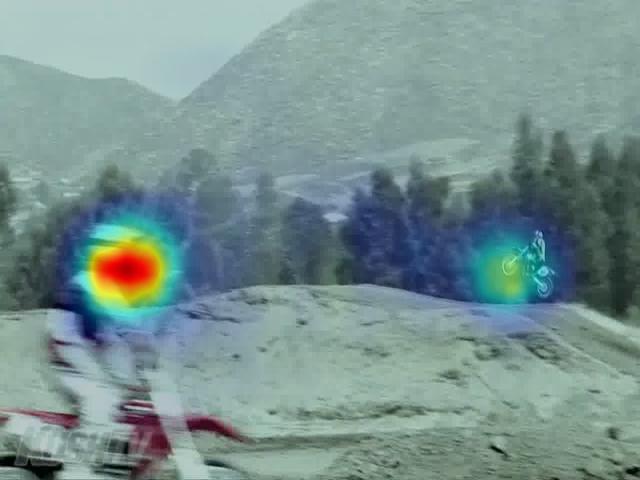} &\includegraphics[width=0.25\linewidth]{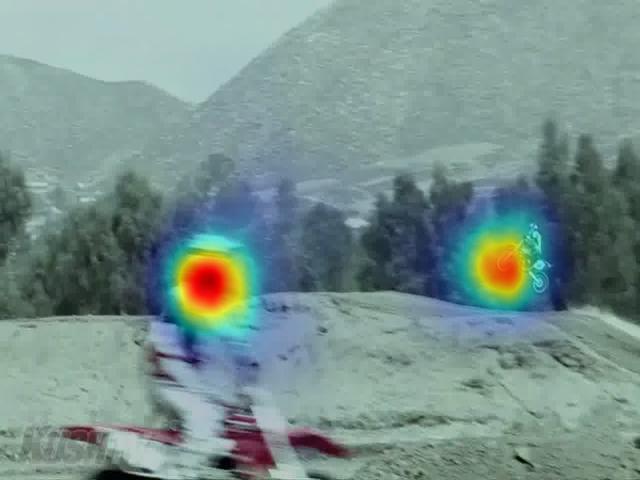}\\
\rotatebox{90}{\huge{$\quad\!\!$SalEMA}}
&\includegraphics[width=0.25\linewidth,height=0.1875\linewidth]{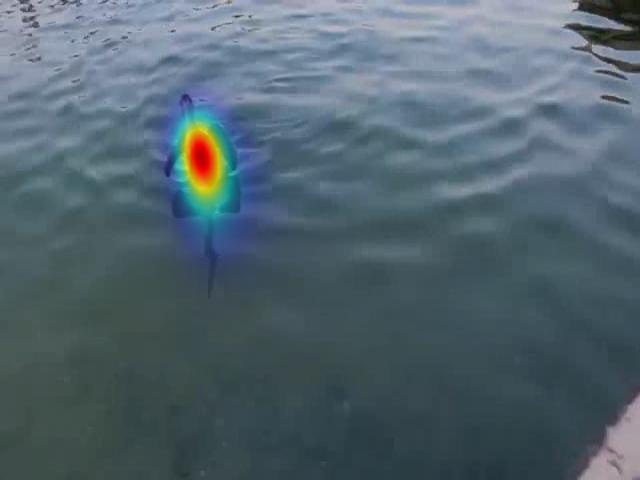} 
&\includegraphics[width=0.25\linewidth,height=0.1875\linewidth]{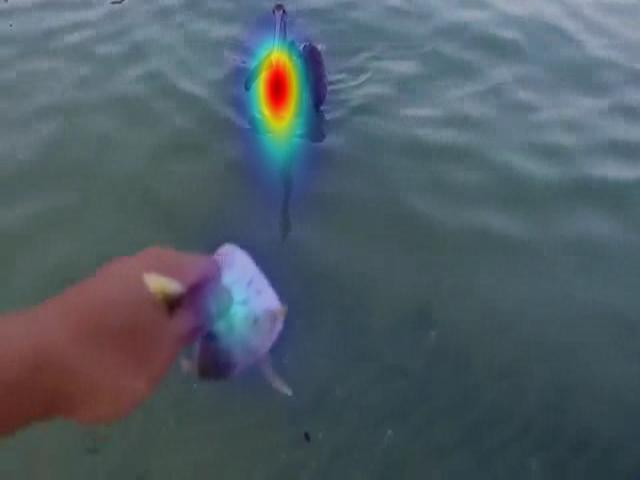} &\includegraphics[width=0.25\linewidth,height=0.1875\linewidth]{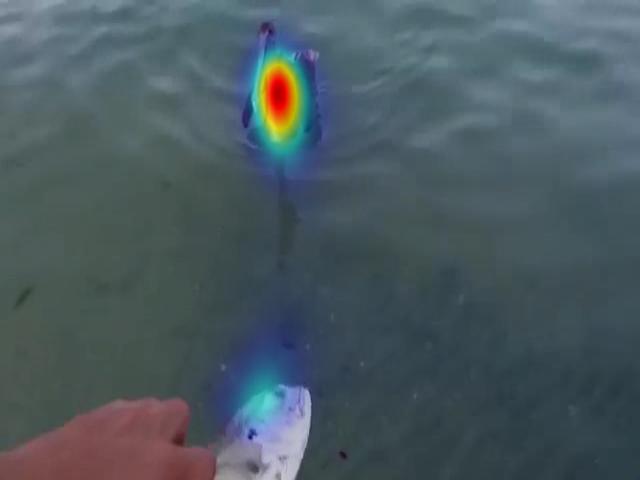} &\includegraphics[width=0.25\linewidth,height=0.1875\linewidth]{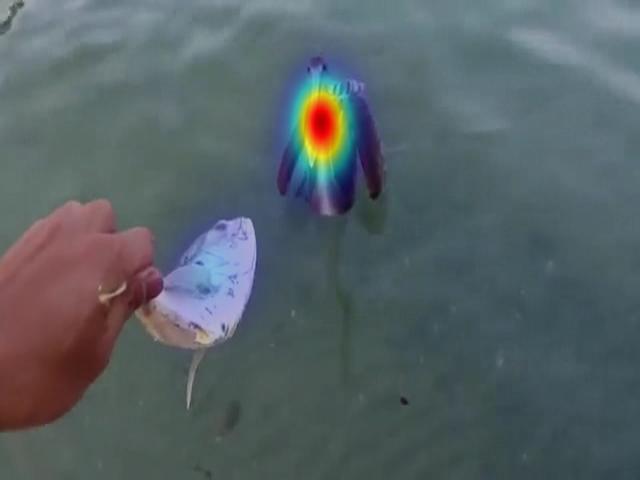} &\includegraphics[width=0.25\linewidth]{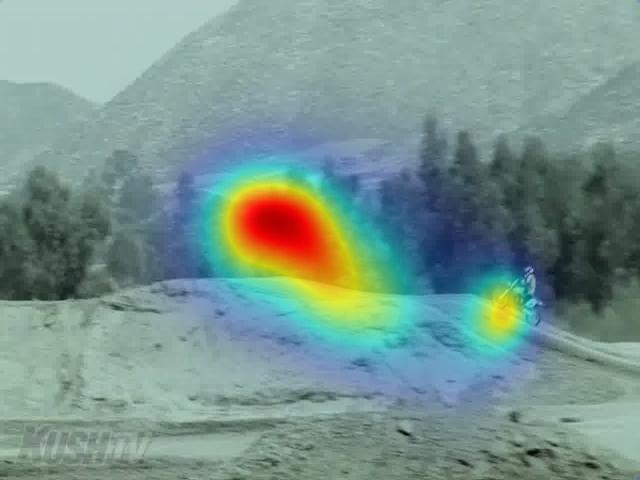} &\includegraphics[width=0.25\linewidth]{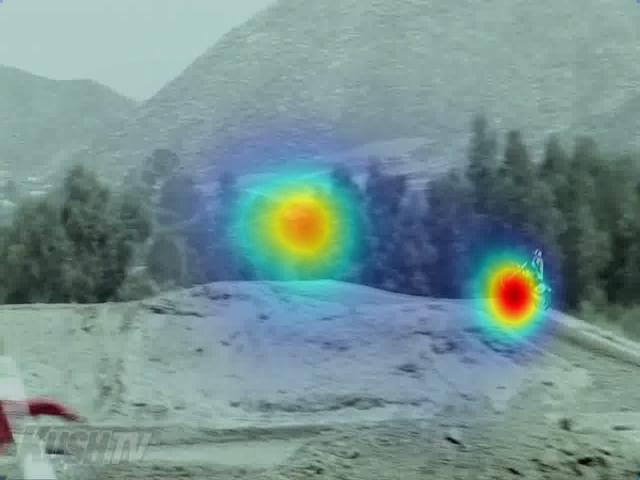} &\includegraphics[width=0.25\linewidth]{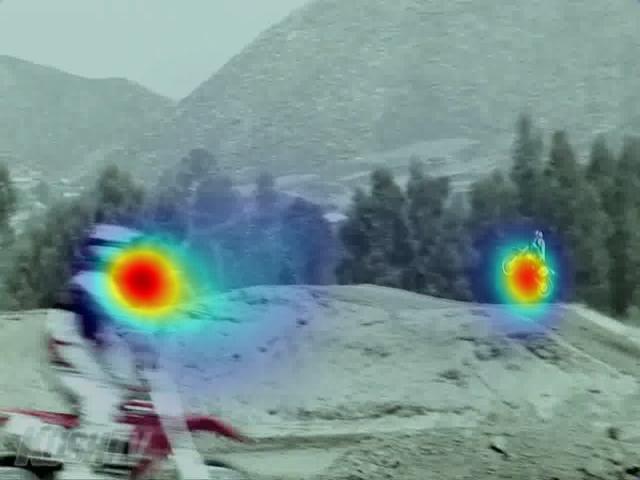} &\includegraphics[width=0.25\linewidth]{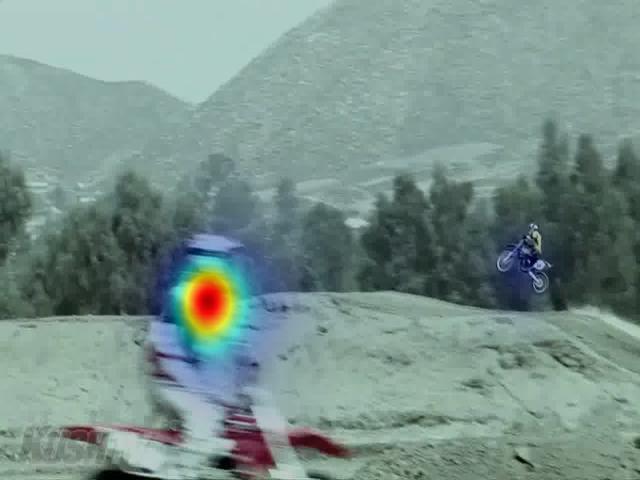}\\
\rotatebox{90}{\huge{$\quad\!\!$ACLNet}}
&\includegraphics[width=0.25\linewidth,height=0.1875\linewidth]{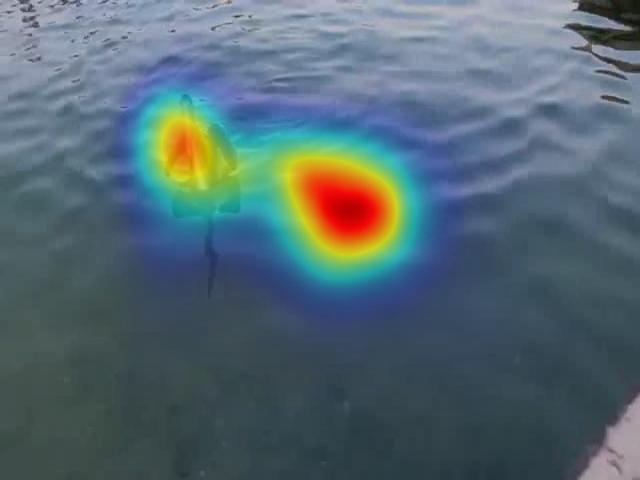} 
&\includegraphics[width=0.25\linewidth,height=0.1875\linewidth]{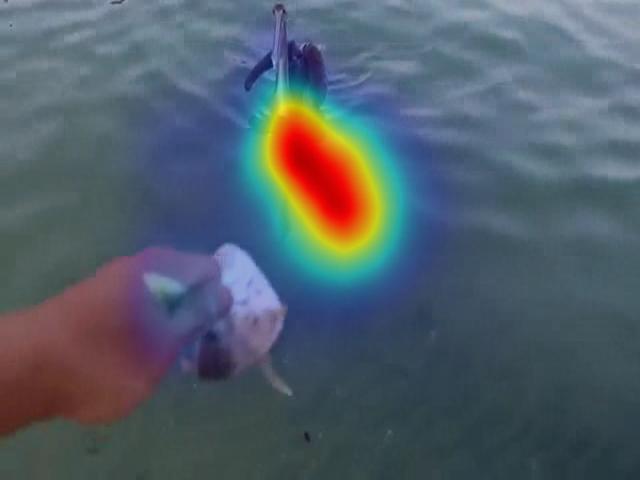} &\includegraphics[width=0.25\linewidth,height=0.1875\linewidth]{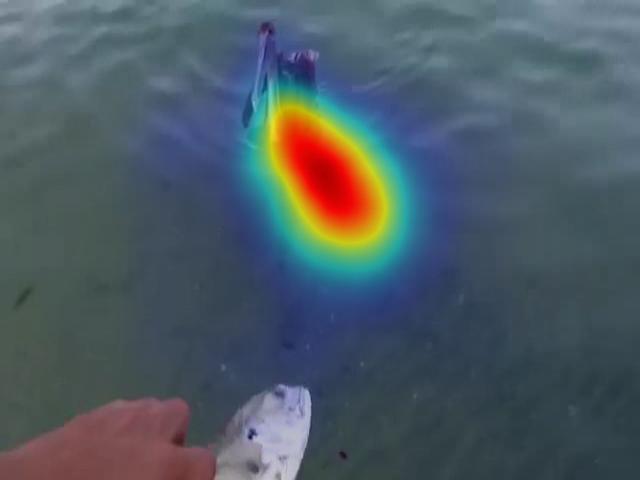} &\includegraphics[width=0.25\linewidth,height=0.1875\linewidth]{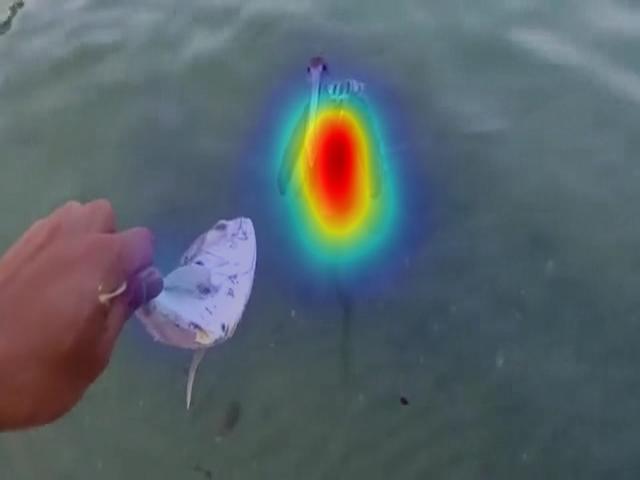} &\includegraphics[width=0.25\linewidth]{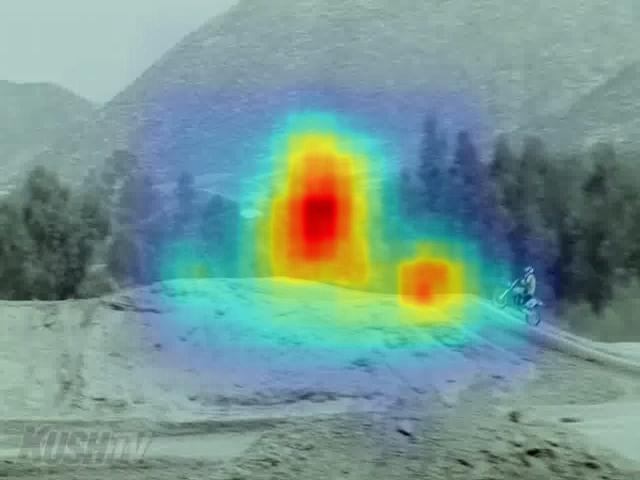} &\includegraphics[width=0.25\linewidth]{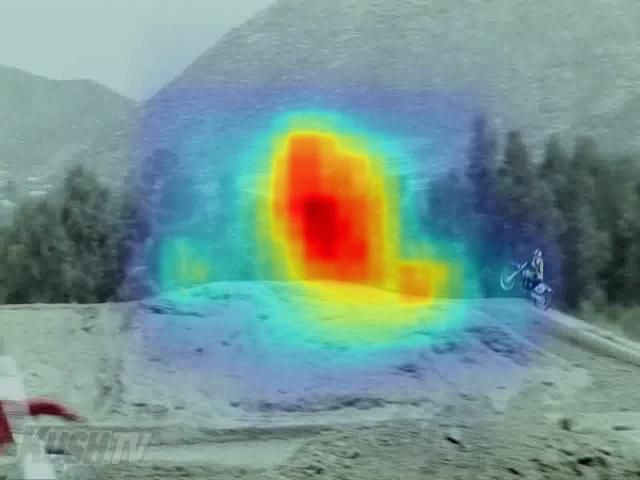} &\includegraphics[width=0.25\linewidth]{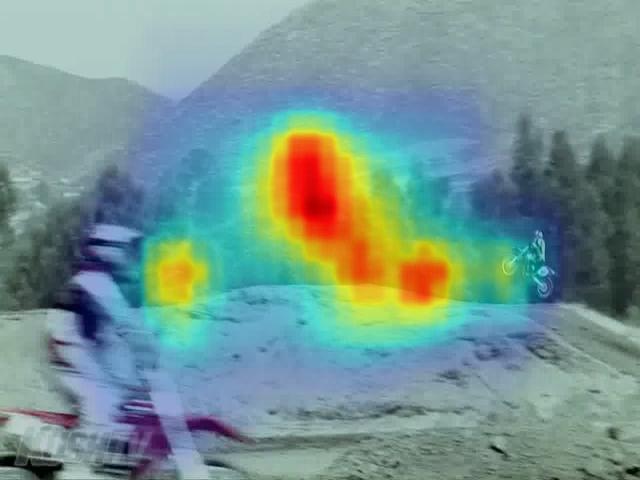} &\includegraphics[width=0.25\linewidth]{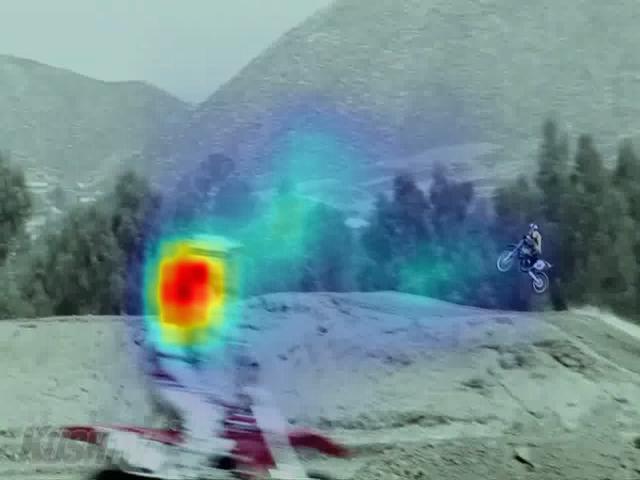}\\
\rotatebox{90}{\LARGE{$\;\;\!\!$TASED-Net}}
&\includegraphics[width=0.25\linewidth,height=0.1875\linewidth]{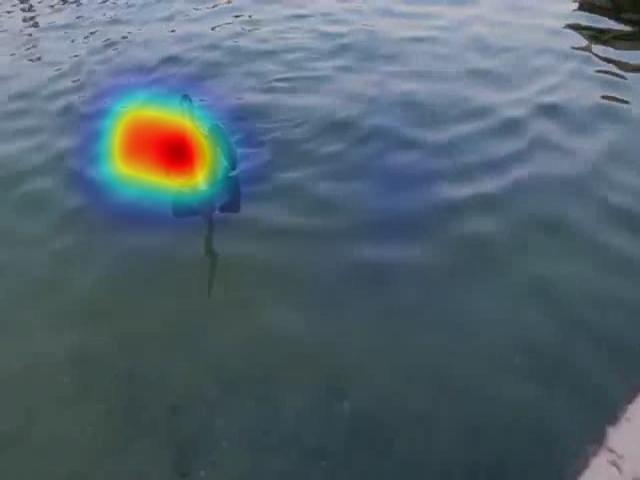} 
&\includegraphics[width=0.25\linewidth,height=0.1875\linewidth]{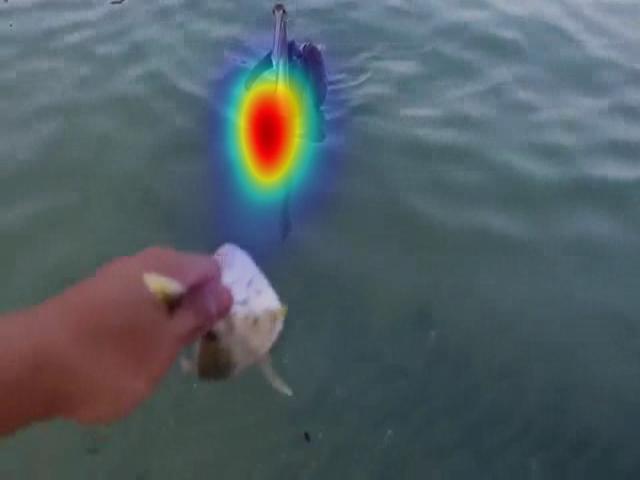} &\includegraphics[width=0.25\linewidth,height=0.1875\linewidth]{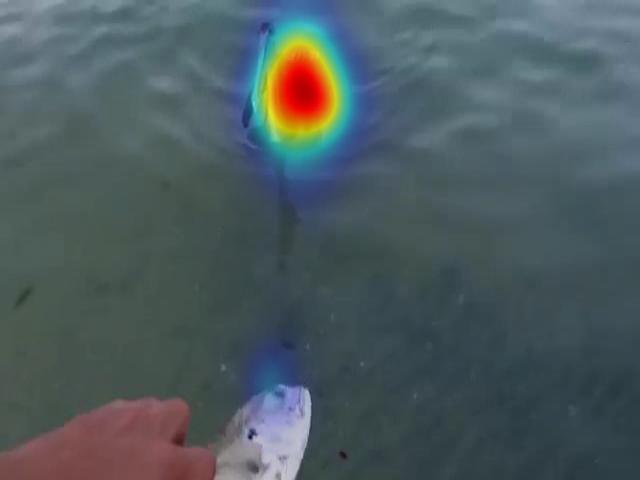} &\includegraphics[width=0.25\linewidth,height=0.1875\linewidth]{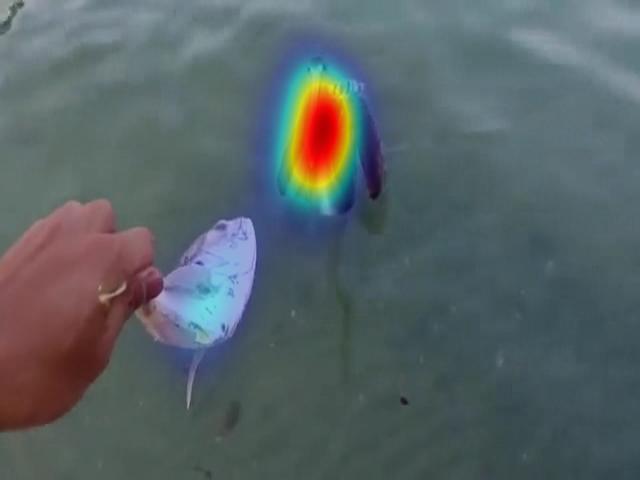} &\includegraphics[width=0.25\linewidth]{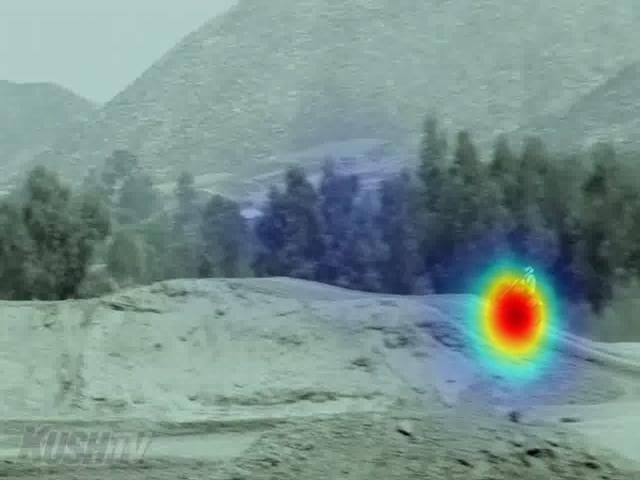} &\includegraphics[width=0.25\linewidth]{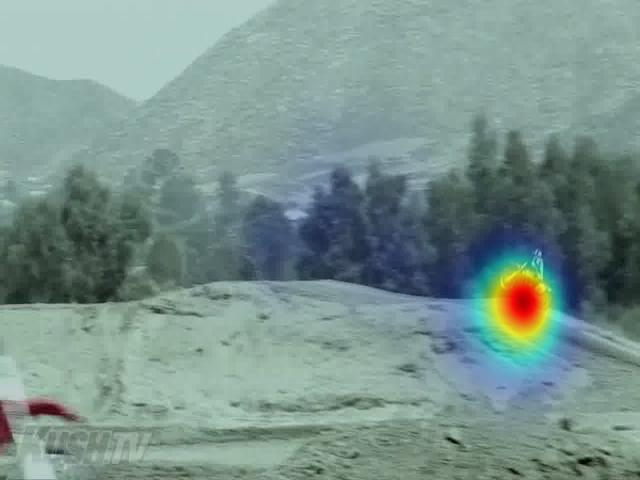} &\includegraphics[width=0.25\linewidth]{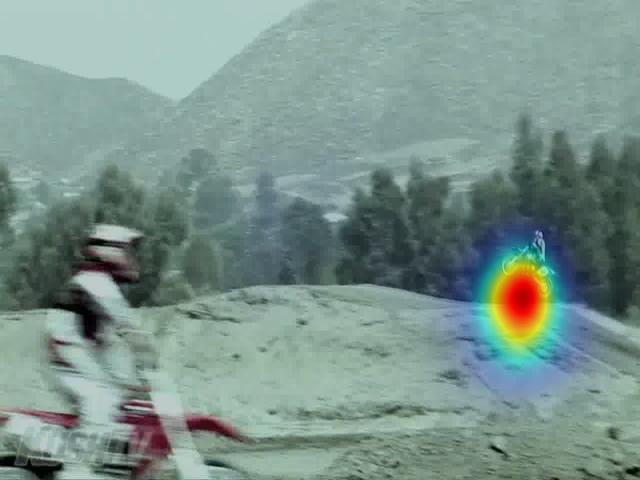} &\includegraphics[width=0.25\linewidth]{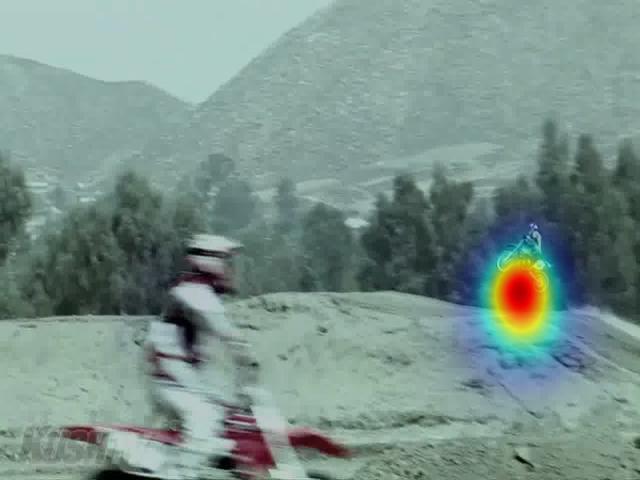}\\
\rotatebox{90}{\LARGE{$\;\;\;\,\!\!$STRA-Net}}
&\includegraphics[width=0.25\linewidth,height=0.1875\linewidth]{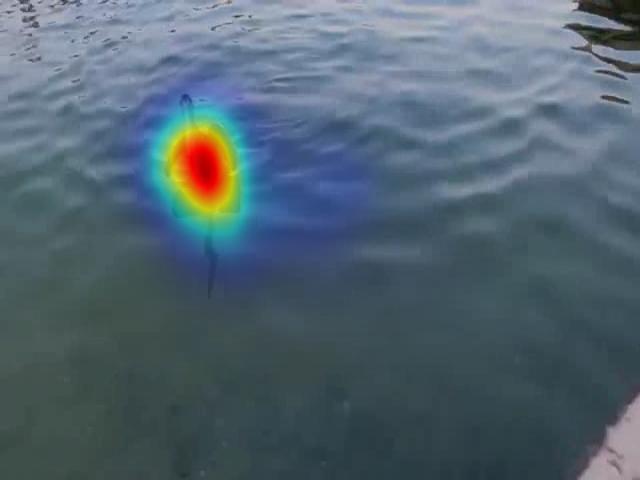} 
&\includegraphics[width=0.25\linewidth,height=0.1875\linewidth]{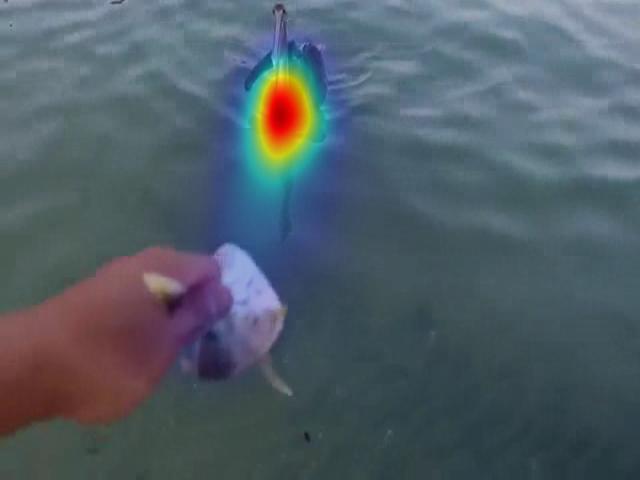} &\includegraphics[width=0.25\linewidth,height=0.1875\linewidth]{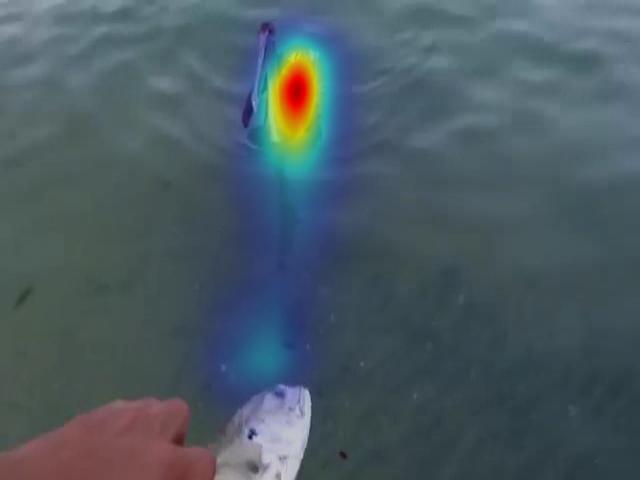} &\includegraphics[width=0.25\linewidth,height=0.1875\linewidth]{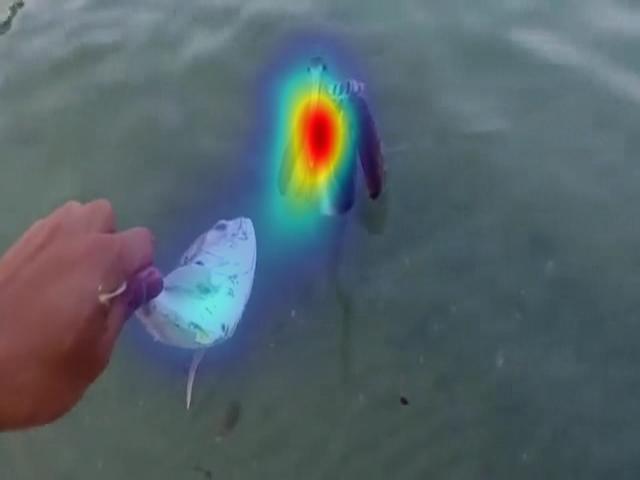} &\includegraphics[width=0.25\linewidth]{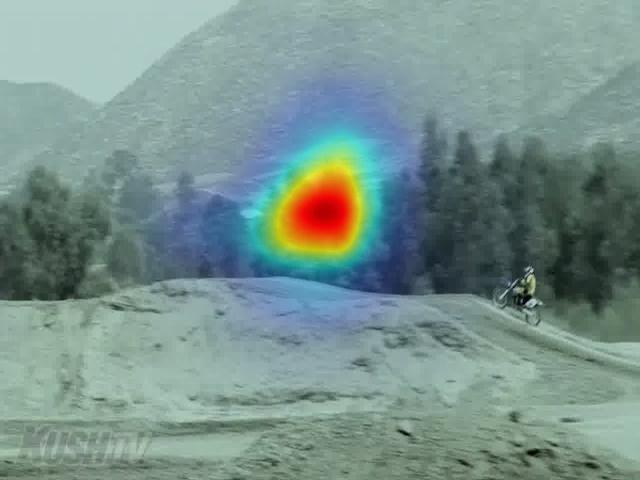} &\includegraphics[width=0.25\linewidth]{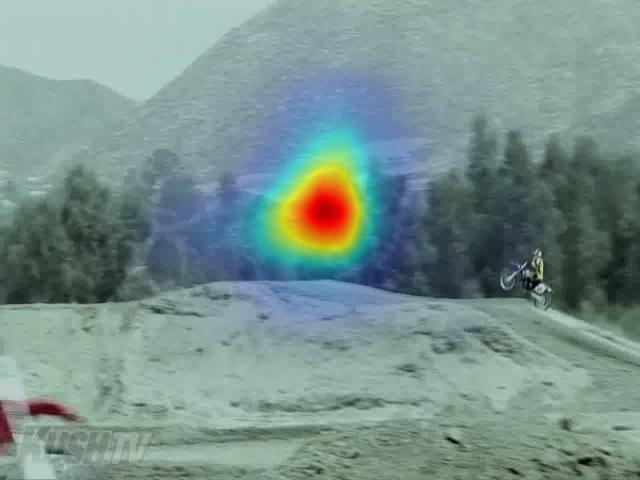} &\includegraphics[width=0.25\linewidth]{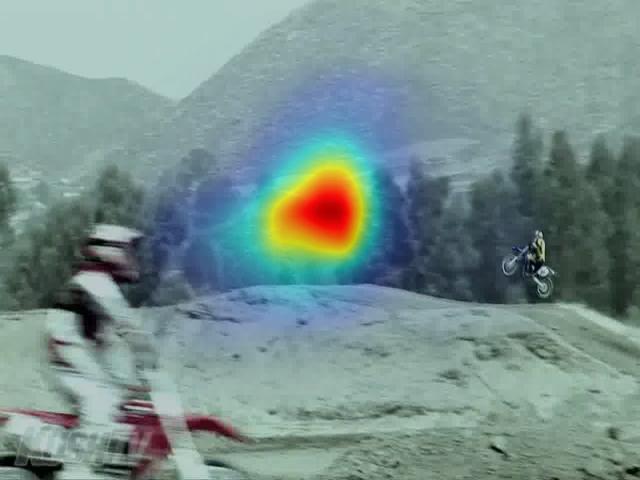} &\includegraphics[width=0.25\linewidth]{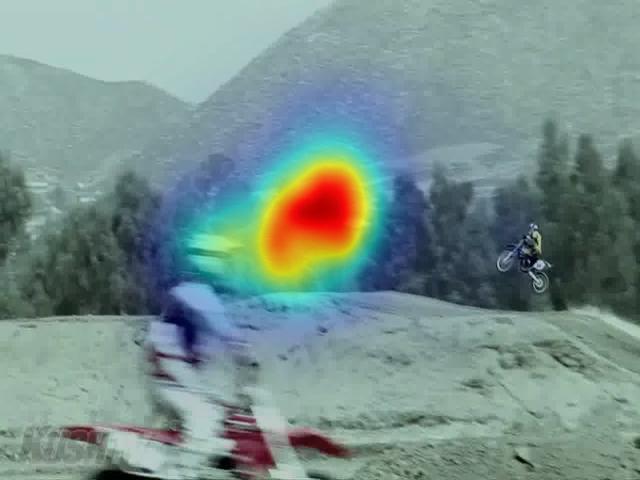}\\
& \multicolumn{4} {c}{\huge{(c) DHF1K}}&\multicolumn{4} {c}{\huge{(d) DIEM}}
\end{tabular}}
\caption{Qualitative results of our proposed framework and the deep learning based SalEMA, ACLNet and SalGAN models. Our approach, in general, produces more accurate saliency predictions than these state-of-the-art models.}
\label{fig:qualitative_results}
\end{figure*}

In Fig.~\ref{fig:qualitative_results}, we show some sample saliency maps predicted by our proposed model and three other deep saliency networks: ACLNet, SalEMA, STRA-Net, and TASED-Net models. As one can observe, our model makes generally better predictions than the competing approaches. For instance, for the sequence from UCF-Sports (Fig.~\ref{fig:qualitative_results}(a)) most the models fail to identify the salient region on the swimmer, or for the sequence from the Hollywood-2 dataset (Fig.~\ref{fig:qualitative_results}(b)) our model is the only model that correctly predicts the soldier at the center of the background as salient. Similar kind of observations are also valid for the sample sequences from DHF1K (Fig.~\ref{fig:qualitative_results}(c)) and DIEM (Fig.~\ref{fig:qualitative_results}(d)) datasets.\\

\noindent\textbf{Performance on DIEM-Meta and LEDOV-Meta.} 
As mentioned before, \cite{temporalsal} have recently showed that most of the current benchmarks for video saliency  include many sequences in which spatial attention is more dominant than temporal effects in describing saliency. DIEM-Meta and LEDOV-Meta datasets  are curated in a special way to contain video frames in which temporal signals are found to be more influential than appearance cues. Hence, they both offer a better way to test how well a dynamic saliency model utilizes temporal information. In our experimental evaluation, we compare our proposed model with the state-of-the-art deep trackers, which are all trained on the combined training set that includes frames from DIEM or LEDOV datasets. As can be seen from Table~\ref{tab:meta_diem} and Table~\ref{tab:meta_ledov}, our model outperforms all the other models in DIEM-Meta, and is the second best model in LEDOV-Meta, achieving highly competitive performances. These results demonstrate the effectiveness of the proposed gated mechanism and its ability to use temporal information to the full extent, as compared to the state-of-the-art approaches.

Overall, the results reported on all the six datasets considered in our experimental analysis suggest that our model has better capacity to mimic human attention mechanism by combining the temporal and static clues in an effective way. It has a better generalization ability that it can predict where people look at the videos from unseen domains much better. Moreover, it utilizes the temporal information more successfully with its gated fusion mechanism, which adaptively integrates spatial and temporal cues depending on video content.

\begin{table}[!t]
		\caption{Performance comparison on DIEM-Meta dataset. The best and the second best performing models are shown in bold typeface and underlined, respectively.}
		\centering
		\resizebox{\columnwidth}{!}{%
	{\begin{tabular}{p{0.33\linewidth}||ccccc}
		\hline
		\backslashbox{\small{Method}}{\small{Metric}}&\small{AUC-J$\uparrow$} & \small{CC$\uparrow$} & \small{NSS$\uparrow$} & \small{SIM$\uparrow$} &\small{KLDiv$\downarrow$}\\ 
		\hline
		\small{ACLNet} & \small{0.845} & \small{0.437} & \small{1.627} & \small{0.391}& \small{1.473}\\ 
		\small{SalEMA} & \small{0.832} & \small{0.392} & \small{1.576} & \small{0.374}& \small{{1.664}}\\ 
	    \small{STRA-Net} & \small{0.840} & \small{0.419} & \small{1.637} & \small{0.385}& \small{1.634}\\ 
	    \small{TASED-Net} & \small{\underline{0.857}} &\small{\underline{0.455}} & \small{\underline{1.810}} & \small{\textbf{0.416}}& \small{\underline{1.479}}\\
	   \small{Ours} & \small{\textbf{0.857}} & \small{\textbf{0.460}} & \small{\textbf{1.814}}& \small{\underline{0.395}}& \small{\textbf{1.305}}\\ 
		\hline 
	\end{tabular}}
	}
	\label{tab:meta_diem}
\end{table}

\begin{figure*}[!t]
\centering
\includegraphics[width=0.99\textwidth,height=0.26\textheight]{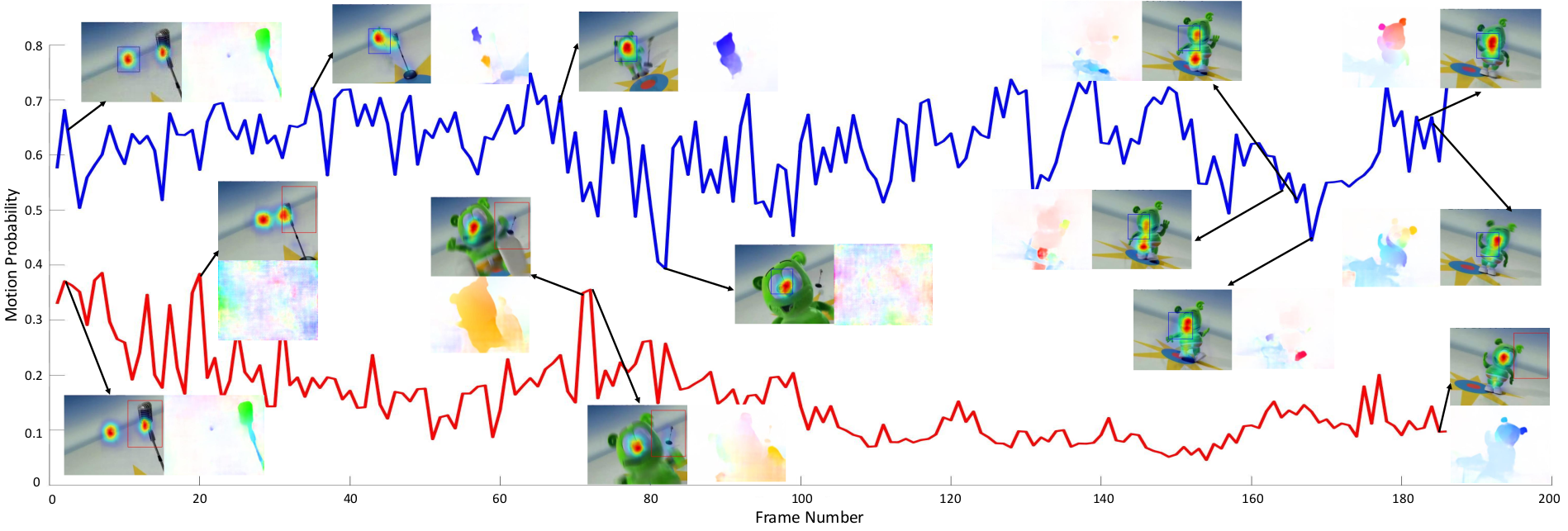}
\caption{Our model dynamically decides the contribution of motion and appearance streams via gated fusion. Here, we plot the average motion probabilities (the contribution of motion stream) for two regions having different characteristic, one  containing a moving object (the gummy bear) and the other with relatively no motion, shown with red and blue, respectively. As can be seen, our model assigns higher weights to the motion stream when motion becomes the dominant visual cue, and the weights adaptively change throughout the sequence.}
\label{fig:contrib-gated}
\end{figure*}

\begin{table}[!t]
		\caption{Performance comparison on LEDOV-Meta dataset. The best and the second best performing models are shown in bold typeface and underlined, respectively.}
		\centering
		\resizebox{\columnwidth}{!}{%
	{\begin{tabular}{p{0.33\linewidth}||cccccc}
		\hline
		\backslashbox{\small{Method}}{\small{Metric}}&\small{AUC-J$\uparrow$}  & \small{CC$\uparrow$} & \small{NSS$\uparrow$} & \small{SIM$\uparrow$}& \small{KLDiv$\downarrow$}\\ 
		\hline
		\small{ACLNet} & \small{0.879} &\small{0.384} & \small{1.750} & \small{0.342}& \small{1.837}\\ 
	    \small{SalEMA} & \small{0.863} & \small{0.380} & \small{1.815} & \small{0.353}& \small{1.850}\\ 
	    \small{STRA-Net} & \small{\textbf{0.893}} & \small{0.423} & \small{2.041} & \small{\underline{0.370}}& \small{2.304}\\
	    \small{TASED-Net} & \small{0.882} &\small{\textbf{0.489}} & \small{\textbf{2.450}} & \small{\textbf{0.403}}& \small{\underline{1.697}}\\ 
	    \small{Ours}&\small{\underline{0.892}} & \small{\underline{0.457}} & \small{\underline{2.190}} & \small{\underline{0.370}}& \small{\textbf{1.485}}\\ 
		\hline 
	\end{tabular}}
	}
	\label{tab:meta_ledov}
\end{table}

\subsection{Ablation study.}
In this section, we aim to analyze the influence of each component of our proposed deep dynamic saliency model. We perform the ablation study on UCF-Sports dataset by disabling or removing some blocks of our model and by examining how these changes affect the model performance. As we did with training our proposed model, for each version of our model under evaluation, we first train a single stream model on SALICON dataset and then use this model to finetune the actual two-stream version. Accordingly, Table~\ref{tab:results_ablation} reports the performance of different versions of our saliency model.\\

\noindent
\textbf{Effect of gated fusion.} As we emphasized before, the role of gated fusion block is to adaptively integrate spatial and temporal streams is a key component of our model. In our analysis, we replace the gated fusion block with a standard $1\times 1$ convolution layer. As can be seen from Table~\ref{tab:results_ablation}, the performance of the model decreases considerably without the gated fusion mechanism. That is, using a dynamic weighting strategy, instead of a fixed weighting scheme (learned via $1\times 1$ convolution), generates much better predictions.  Fig.~\ref{fig:contrib-gated} shows a visualization of how our proposed gated fusion operates, demonstrating the behavior of the weighting scheme for both dynamic and static parts of a given scene, In particular, we plot the motion probabilities averaged within the corresponding image regions over time, which clearly shows that the motion probability (the contribution of motion stream) for the region that contains a moving object is, in general, much higher than that of the static region. Moreover, depending on the characteristics of the regions, it shows the changes in the motion probabilities throughout the whole sequence. For example, when no motion is taking place in the region initially containing the moving object, the weight of the temporal stream starts to fall. These results supports our main claim that considering the content of the video while combining temporal and spatial cues is a more appropriate way to model saliency estimation on dynamic scenes.

\noindent
\textbf{Effect of multi-level information.} Previous studies demonstrate that low and high-level cues are equally important for saliency prediction~\cite{Bruce_2016_CVPR, bylinskii2016should}. Motivated with these, we included a multi-level information block to fuse features extracted from different levels of our deep model. For this analysis, we disable this multi-level information block and train a single-scale model instead. Compared to our full model, disabling this block reduces the performance as can be seen in Table~\ref{tab:results_ablation}. Employing a representation that contains information from low and high levels helps to improve the performance of our model. We speculate that our multi-level information block allows the network to better identify the regions semantically important for saliency.

\renewcommand*{\arraystretch}{1.15}
\begin{table}[!t]
		\caption{Ablation study on UCF-Sports dataset.}
		\centering
		\resizebox{\columnwidth}{!}{%
	{\begin{tabular}{l||cccccc}
		\hline
		\backslashbox{Method}{Metric}&AUC-J$\uparrow$  & CC$\uparrow$ & NSS$\uparrow$ & SIM$\uparrow$ & \small{KLDiv$\downarrow$} \\ 
		\hline
		w/o spatial attention&0.872&0.474&2.884&0.374&2.223\\ 
		w/o channel-wise attention&0.892&0.489&2.923&0.319&1.707\\ 
		w/o spatial \& ch.-wise attention&0.875&0.447&2.885&0.364&2.646\\
		w/o multi-level information&0.890&0.484&2.755&0.303&1.711\\
		w/o gated fusion& 0.900&0.480&2.913&0.353&1.676\\
		full model&\textbf{0.914}&\textbf{0.526}&\textbf{3.333}&\textbf{0.382}&\textbf{1.516}  \\
		\hline 
	\end{tabular}}}
	\label{tab:results_ablation}
\end{table}

\noindent
\textbf{Effect of attention blocks.} As discussed before,  { the reasons we introduce the attention blocks are to eliminate the irrelevant features via the spatial attention and to choose the most informative feature channels via the channel-wise attention when processing a video frame. In this experiment, we remove the spatial and the channel-wise attention blocks from our full model and train two different models, respectively. The results given in Table~\ref{tab:results_ablation} support our assertion that both of these attention blocks improve the model performance. Disabling them results in a much lower performance as compared to that of the full model.}

\section{Summary and Conclusion}
In this study, we proposed a new spatio-temporal saliency network for video saliency. It follows a two-stream network architecture that processes spatial and temporal information in separate streams, but it extends the standard structure in many ways. First, it includes a gated fusion block that performs integration of spatial and temporal streams in a more dynamic manner by deciding the contribution of each channel one frame at a time. Second, it utilizes a multi-level information block that allows for performing multi-scale processing of appearance and motion features. Finally, it employs spatial and channel-wise attention blocks to further increase the selectivity. Our extensive set of experiments on six different benchmark datasets shows the effectiveness of the proposed model in extracting the most salient parts of the video frames both qualitatively and quantitatively. Moreover, our ablation study demonstrates the gains achieved by each component of our model. Our analysis reveals that the proposed model deals with the videos from unseen domains much better that the existing dynamic saliency models. Additionally, it uses temporal cues more effectively via the proposed gated fusion mechanism which allows for adaptive integration of spatial and temporal streams.

We believe that our work highlights several important directions to pursue for better modeling of saliency in videos. As future work, we plan to explore more efficient ways to include the temporal information. For instance, instead of using optical flow images, one can use features extracted from early and mid layers of an optical flow network model to encode motion information. This can reduce the memory footprint of the model and decreases the running times.

\section*{Acknowledgments}
This work was supported in part by TUBA GEBIP fellowship awarded to E. Erdem.

\bibliographystyle{IEEEtran}
\bibliography{06_Bibliography_Clean}

\end{document}